%% file: paper.tex
\documentclass[]{fairmeta}
\usepackage{hyperref}
\usepackage{xcolor}
\usepackage{array}
\usepackage{graphicx}
\usepackage{tabularx}
\usepackage{colortbl}
\usepackage{amssymb}
\usepackage{verbatim}
\usepackage{amsmath}      % equation, align
\usepackage{booktabs}     % \toprule, \midrule, \bottomrule
\usepackage{multirow}     % \multirow
\usepackage{colortbl}     % \cellcolor in tabular
\usepackage{cleveref}     % \Cref{...}
\usepackage{svg}
\usepackage{arydshln}
\usepackage{rotating}   % REQUIRED: Provides the sidewaystable environment
\usepackage{arydshln}   % REQUIRED: For dashed baseline separators
\usepackage{algorithm}      % REQUIRED: pseudo-code environments in appendix
\usepackage{algpseudocode}  % REQUIRED: \State, \For, \If, ... macros
\usepackage{titletoc}           % REQUIRED: local (appendix-only) table of contents
\usepackage{caption}
\usepackage{wrapfig}
\usepackage[table]{xcolor}

\newcommand{\resroot}{./figures/results/bon}
\newcolumntype{Y}{>{\centering\arraybackslash}X}

\newlength{\demoimg}
\algnewcommand{\LComment}[1]{\State \(\triangleright\) #1}

\newcommand{\dldir}{dl3dv_006048_9f361436f272b69706cdba2747ab7b86d7227bbbd95458cc927255d11f6e9bbb_frame_00101_to_frame_00079}

\newcommand{\tntdir}{tnt_barn_src-000001_tgt-000016_idx000002}

\newcommand{\redir}{re10k_5a7df53b469c351d_src-frame_00020_tgt-frame_00070_idx047519}

\newcommand{\democellsrc}[1]{\srcframe{\includegraphics[width=\demoimg]{figures/results/#1/src.png}}}

\newcommand{\democellgt}[1]{\gtframe{\includegraphics[width=\demoimg]{figures/results/#1/gt.png}}}

\newcommand{\democellours}[2]{\eacframe{\includegraphics[width=\demoimg]{figures/results/#1/#2.png}}}

\newcommand{\democellbase}[2]{\baseframe{\includegraphics[width=\demoimg]{figures/results/#1/#2.png}}}

\newcommand{\demohd}[1]{\makebox[\demoimg][c]{\scriptsize #1}}

\newcommand{\demorow}[2]{%

  \rotatebox{90}{\makebox[\demoimg][c]{\scriptsize #2}}
  & \democellsrc{#1}
  & \democellbase{#1}{photonvs}
  & \democellbase{#1}{vivid}
  & \democellbase{#1}{ovie}
  & \democellbase{#1}{gld}
  & \democellbase{#1}{lagernvs}
  & \democellours{#1}{eac_shadow}
  & \democellours{#1}{eac_base}
  & \democellours{#1}{eac_bon}
  & \democellgt{#1} \\%
}

\newlength{\borderrule}
\newcommand{\srcframe}[1]{{\setlength{\fboxrule}{\borderrule}\setlength{\fboxsep}{-\borderrule}\fcolorbox{legendinput}{legendinput}{#1}}}
\newcommand{\gtframe}[1]{{\setlength{\fboxrule}{\borderrule}\setlength{\fboxsep}{-\borderrule}\fcolorbox{legendtarget}{legendtarget}{#1}}}
\newcommand{\eacframe}[1]{{\setlength{\fboxrule}{\borderrule}\setlength{\fboxsep}{-\borderrule}\fcolorbox{legendours}{legendours}{#1}}}
\newcommand{\baseframe}[1]{{\setlength{\fboxrule}{\borderrule}\setlength{\fboxsep}{-\borderrule}\fcolorbox{legendbaseline}{legendbaseline}{#1}}}
\newcommand{\rescellsrc}[1]{\srcframe{\includegraphics[width=\linewidth]{#1}}}
\newcommand{\rescellgt}[1]{\gtframe{\includegraphics[width=\linewidth]{#1}}}
\newcommand{\rescellours}[1]{\eacframe{\includegraphics[width=\linewidth]{#1}}}

\definecolor{redfirst}{rgb}{1.0, 0.68, 0.68}   % Strong pastel red (1st Place)
\definecolor{redsecond}{rgb}{1.0, 0.80, 0.80}  % Medium pastel red (2nd Place)
\definecolor{redthird}{rgb}{1.0, 0.91, 0.91}   % Soft pastel red (3rd Place)

\newcommand{\first}[1]{\cellcolor{redfirst}\boldmath #1}
\newcommand{\second}[1]{\cellcolor{redsecond}#1}
\newcommand{\third}[1]{\cellcolor{redthird}#1}

\newcommand{\modelsep}{\arrayrulecolor{black!25}\midrule\arrayrulecolor{black}}

\definecolor{dscomment}{HTML}{FC03EB}

\definecolor{olcomment}{HTML}{009E73}

\definecolor{eachighlight}{rgb}{0.93, 0.96, 1.0}
\definecolor{pcccolor}{RGB}{192,57,43}
\definecolor{psnrcolor}{RGB}{41,128,185}
\definecolor{ssimcolor}{RGB}{39,174,96}
\definecolor{lpipscolor}{RGB}{142,68,173}
\definecolor{anchorbg}{RGB}{255,249,230}
\definecolor{legendinput}{RGB}{37,116,205}
\definecolor{legendtarget}{RGB}{34,150,79}
\definecolor{legendours}{RGB}{210,92,38}
\definecolor{legendbaseline}{RGB}{100,100,100}

\newcommand{\modelname}{ESC}

\title{ESC: Escaping 2D Diffusion Ambiguity with Geometric Constraints for Consistent Novel View Synthesis}

\author[1,2,3,*]{Ge Ya (Olga) Luo}
\author[1]{Daniel Severo}
\author[1]{Brian Karrer}
\author[2,5,6]{Chris Pal}
\author[1,2,4,5,\dagger]{Adriana Romero-Soriano}
\author[1,\dagger]{Michal Drozdzal}

\affiliation[1]{FAIR at Meta}
\affiliation[2]{Mila -- Qu\'{e}bec AI Institute}
\affiliation[3]{Universit\'{e} de Montr\'{e}al}
\affiliation[4]{McGill University}
\affiliation[5]{Canada CIFAR AI Chair}
\affiliation[6]{Polytechnique Montr\'{e}al}

\contribution[*]{Work done at Meta}
\contribution[\dagger]{Joint last author}

\abstract{\input{main/abstract}}

\date{\today}
\correspondence{Ge Ya (Olga) Luo at \email{xugeya@mila.quebec}}

\begin{document}

\maketitle

\input{main/introduction}

\input{main/related_work}

% ESC Generative Model
\section{Method}
%Our method aims to address the critical challenges of monocular novel-view synthesis by bridging explicit geometric reasoning with generative diffusion priors. 
We argue that evaluating 3D alignment between input and generated novel views must be lifted beyond 2D image-quality and camera-pose metrics into 3D space. To operationalize this, we tackle the problem from two complementary angles. First, in \Cref{sec:pcc_analysis}, we introduce \emph{Projection Consistency Constraint (PCC)}, a novel 3D consistency objective that evaluates geometric integrity across synthesized views. Then, in \Cref{sec:esc_main}, we present \modelname, a geometry-aware diffusion framework that jointly synthesizes RGB and depth conditioned on Pl\"ucker rays and leverages PCC for inference-time Best-of-$N$ selection to enforce 3D coherence.

% PCC metric
\input{main/PCC/main}

% ESC model
\input{main/esc}

\input{main/experiments}

\input{main/ablation/main}

\section*{Acknowledgments}
We thank Xiaofeng Zhang, Nicolas Beltran-Velez, Jianhao Yuan, and Felix Friedrich for insightful discussions and valuable exchanges of ideas related to this work. We are also grateful to Professor Andrea Vedaldi for his valuable feedback.
\clearpage
\newpage
\bibliographystyle{assets/plainnat}
\bibliography{paper}

\clearpage
\newpage
\beginappendix
\input{appendix/main}

\end{document}

%% file: main/abstract.tex
Generating 3D-consistent novel views from 2D diffusion models remains a fundamental challenge in computer vision. In this work, we introduce a framework that enhances geometric alignment for monocular view synthesis under drastic camera changes by jointly generating a target view and its corresponding depth map. By conditioning the diffusion pipeline on the single source image, source depth, and target camera rays, our approach forces the model to inherently reason in 3D space even under extreme viewpoint shifts. To guarantee spatial fidelity, we propose a novel per-pixel depth consistency constraint, which verifies and enforces 3D consistency across generated views. Our framework establishes a robust pipeline for geometrically consistent view synthesis, yielding superior visual accuracy and spatial alignment compared to current state-of-the-art methods. With this work, we pave the road towards robust 3D world models capable of understanding complex environments from a single view image.\looseness-1

%% file: main/introduction.tex
\section{Introduction}
\begin{figure}[p]
\centering
\includegraphics[width=\linewidth]{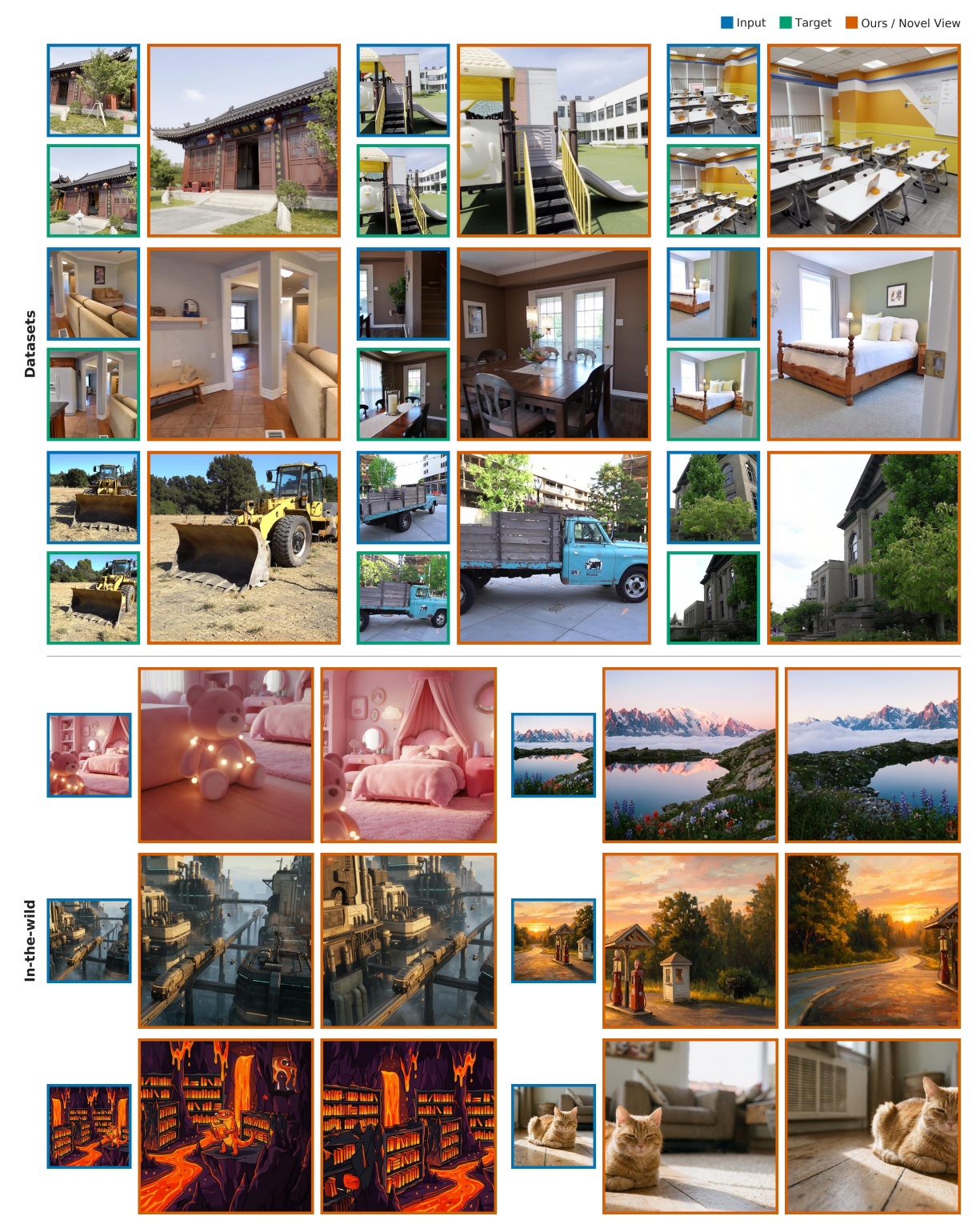}
% \caption{Given a single \textcolor{legendinput}{input} view, our method generates \textcolor{legendours}{novel views} that remain 3D-consistent with the \textcolor{legendtarget}{target} across challenging benchmarks (DL3DV, RealEstate10K, Tanks \& Temples) and selective in-the-wild scenes. \md{can we specify/clarify what is the prompt to in-the-wild setup?} \ars{Can we use colorblind-friendly colors? e.g. orange-blue-gray, maybe also consider showing results from prior art, hopefully showcasing some of their failure-modes}}
\caption{\textbf{Novel view synthesis highlights.} Given a single \textcolor{legendinput}{input} view, its corresponding depth map, and Pl\"ucker ray conditioning, our method generates \textcolor{legendours}{novel views} that remain 3D-consistent with the \textcolor{legendtarget}{target} across challenging datasets (DL3DV, RealEstate10K, T\&T) and selective in-the-wild scenes. For the datasets, we illustrate a single novel view generation alongside the ground-truth target, whereas for the in-the-wild setting, we demonstrate two distinct novel view generations.}
\label{fig:teaser}
\end{figure}

Monocular novel view synthesis (NVS) is a fundamental yet highly challenging problem in computer vision, requiring the generation of unseen views of a 3D scene from only a single 2D image. This setting is inherently ambiguous: a single observation provides limited evidence about scene depth, object geometry, and occluded regions, making accurate 3D reasoning difficult without additional geometric constraints. This difficulty increases with extreme viewpoint changes, where local pixel interpolation is insufficient. Because large portions of the target view may be completely unobserved in the source image, prediction becomes an ill-posed extrapolation problem that requires the model to synthesize both scene structure and appearance. %In such cases, large portions of the target view may be completely unobserved in the source image, requiring the model to hallucinate both structure and appearance of the scene. 
To bridge this gap, recent approaches have increasingly leveraged diffusion models~\citep{ho2020ddpm, rombach2021stablediffusion, lipman2023flowmatching}, whose strong generative priors provide plausible visual completions when direct image evidence is missing.\looseness-1

However, relying primarily on generative priors introduces a fundamental tension between photorealistic synthesis and geometric validity. A prevalent strategy for addressing this tension follows a ``geometry-warping and 2D-completion'' pipeline: pixels or features from the source view are first projected into the target view using estimated scene geometry, and a 2D diffusion model is then employed to complete the newly exposed regions~\citep{ren2025gen3c, lin2026vista4d, huang2025voyager, chen2026cognvs, rahary2026ovie, cao2024mvinpainter, yu2024viewcrafter}. % This inpainting approach is geometrically grounded, since the target view is initialized through depth-aware warping. 
Although depth-aware warping provides a geometrically grounded initialization, the central challenge of NVS remains in the regions it cannot resolve: projection transfers only visible surfaces and cannot recover content occluded in the source view. A 2D completion model must therefore infer target-view visibility, depth discontinuities, and severe distortions under large camera shifts without explicit 3D awareness. Treating these ambiguous regions as ordinary missing pixels provides no spatial constraints, often yielding locally plausible but geometrically invalid structures with broken occlusion boundaries, inconsistent depth, and severe floating artifacts.\looseness-1

%Yet the central difficulty of NVS remains precisely in the regions that warping cannot explain. The projection can only transfer surfaces that were visible in the input view; it cannot reveal content hidden behind occluders, nor can it fully determine target-view visibility, depth discontinuities, or geometry-dependent distortions under large camera motion. As a result, the most geometrically ambiguous parts of the problem are delegated to a 2D completion model.

% Reducing NVS to image-plane inpainting in this way is fundamentally limited. Because the model treats these rendering holes as ordinary missing 2D pixels rather than disoccluded 3D surfaces, it is encouraged to synthesize locally plausible textures without guaranteeing a valid 3D configuration. Consequently, the generated views often suffer from inconsistent depth, broken occlusion boundaries, duplicated surfaces, and floating structures.

Escaping these geometric ambiguities requires reasoning beyond the 2D canvas. Rather than relying on purely image-plane priors, we propose to ground synthesis in underlying scene geometry. We introduce \textbf{\modelname}\footnote{Our framework name, \textbf{ESC}ape the Allegory of the Cave (\modelname), draws an analogy to Plato's allegory: standard 2D models are trapped observing only 2D "shadows" on the image plane without understanding the 3D shapes casting them. ESC guides the model outside the cave by conditioning directly on 3D geometry.}, a geometry-aware diffusion framework that jointly predicts target-view depth and appearance conditioned on source view information and Pl\"ucker rays.
% Standard 2D diffusion models struggle with novel-view synthesis because, having been trained exclusively on image-plane databases, they lack comprehension of underlying 3D structures. For such models, a warped target image is merely another incomplete 2D projection, leaving hidden surfaces and occlusion boundaries geometrically ambiguous. Escaping this ambiguity requires reasoning beyond the rendered 2D canvas. To explicitly ground generation in scene geometry, we condition synthesis on depth and jointly predict depth alongside appearance. 
% \ds{This paragraph is nice but maybe could just be a shorter footnote on the name.}
% We note that merely exposing the diffusion model to 3D-aware depth and appearance data during training is insufficient to strictly enforce geometrically consistent generation. While joint depth-appearance prediction endows the network with structural awareness, the probabilistic nature of diffusion models means that soft conditioning alone can still yield locally plausible textures that violate global spatial relationships. 
However, simply exposing the diffusion model to depth data during training does not guarantee 3D consistency. Due to the probabilistic nature of diffusion, the model can still introduce global geometric violations. To rigorously bind the generative process to the underlying scene geometry, we introduce \textbf{P}rojection \textbf{C}onsistency \textbf{C}onstraint \textbf{(PCC)}, an explicit geometric constraint that enforces 3D consistency across synthesized views by penalizing geometric contradictions when reprojecting synthesized pixels into 3D space. By applying PCC during inference, our proposed framework ensures completed regions remain spatially consistent, enabling photorealistic novel-view synthesis that preserves coherent visibility and 3D structure.\looseness-1
% \md{Add details about the validation and comment on the results. Reference fig 1.}

We comprehensively evaluate our framework across challenging large-scale datasets, including DL3DV~\citep{ling2023dl3dv}, RealEstate10K~\citep{zhou2018re10k}, and Tanks \& Temples~\citep{knapitsch2017tnt}, as well as on diverse in-the-wild captures. As illustrated in Figure~\ref{fig:teaser}, our framework effectively mitigates geometric ambiguity to synthesize reliable novel views even for cases where we generate novel views from out-of-distribution images obtained by sampling a state-of-the-art text-to-image generative model~\citep{deepmind_imagen}.

%. We design the PCC as a novel consistency metric as well as corresponding loss function that evaluates and verifies the 3D validity of every generated pixel via its predicted depth. By explicitly projecting each synthesized pixel into 3D space and penalizing geometric contradictions, this constraint forces the model to generate novel views that are not only photorealistic, but strictly anchored to a coherent and physically valid 3D scene representation. 

In summary, our primary contributions are as follows:
\begin{itemize}
    \item 
    \textbf{Geometry-Grounded Generation Framework:} We introduce ESC, a multi-modal diffusion pipeline that conditions generation on Pl\"ucker rays via a Ray ControlNet and jointly predicts RGB and depth, grounding novel view synthesis directly in explicit 3D scene geometry.
    \item 
    \textbf{Explicit Geometric Consistency Constraint:} We propose the PCC, a novel depth-reprojection constraint that penalizes 3D contradictions to enforce spatial consistency across views. Experiments show that PCC captures true 3D spatial consistency where traditional 2D and pose-based metrics fail.
    \item 
    \textbf{State-of-the-Art Performance:} Our full framework surpasses existing baselines, delivering superior visual quality and robust geometric consistency under extreme viewpoint changes.
\end{itemize}

%% file: main/related_work.tex
\section{Related Work}

\textbf{Approaches to Novel View Synthesis.}
Recent NVS approaches can be broadly categorized into feed-forward regression models and diffusion-based generative architectures. Feed-forward regressors, such as OVIE~\citep{rahary2026ovie} and LagerNVS~\citep{szymanowicz2026lagernvs}, predict the target view in a single network pass, prioritizing inference speed at the cost of photorealism and fine-grained texture. Conversely, diffusion-based approaches achieve superior visual fidelity through iterative denoising. Notable examples include PhotoNVS~\citep{yu2023photonvs}, a pixel-space DDPM requiring extensive denoising steps, and VIVID~\citep{elata2024vivid}, a latent diffusion model that utilizes classifier-free guidance. To tackle higher resolutions and multi-view consistency, GLD~\citep{jang2026gld} further advances this paradigm with a multi-stage cascade diffusion design. Our work builds on the generative paradigm.

\textbf{Depth-Aware and Ray-Conditioned Synthesis.}
Recent advances in NVS highlight the critical role of explicit depth priors and precise camera conditioning for geometric fidelity- see \textit{e.g.,} Sharp~\citep{mescheder2026sharp}, Flash3D~\citep{szymanowicz2025flash3d} and Voyager~\citep{huang2025voyager}. Among recent 3D Gaussian Splatting frameworks~\citep{kerbl20233dgs, mescheder2026sharp, szymanowicz2025flash3d}, depth estimation is heavily leveraged to explicitly place or initialize Gaussian primitives along target camera rays, effectively anchoring the rendering process in physical space. We adopt the same geometric principle within a diffusion backbone. In particular, we conceptualize NVS as a ray-conditioned generative process anchored at specific depths, which allows us to enforce rigorous constraints such as our Projection Consistency Constraint (PCC)-ensuring that the generated textures adhere strictly to the underlying 3D structure. To this end, we build on the Qwen-Edit architecture~\citep{wu2025qwenimage}. While recent efforts~\citep{fal2025qwen, dx81522025qwen} have adapted Qwen for multi-view generation via Low-Rank Adaptation (LoRA)~\citep{hu2021lora} with natural language pose prompts, such text-based conditioning is inherently ambiguous and lacks geometric precision. We instead eschew language-based pose approximations and condition directly on Pl\"ucker ray information~\citep{Zisserman2004MVCV} alongside explicit depth priors, providing precise, pixel-aligned viewpoint control.

\textbf{Geometric Foundation Models for 3D Reconstruction.}
Recent advances in 3D computer vision have shifted from highly specialized, task-isolated pipelines~\citep{schonberger2016colmap, bhat2023zoedepth, bochkovskii2025depthpro} toward unified, feed-forward geometric foundation models. Pioneered by architectures like DUSt3R~\citep{wang2024dust3r} and MASt3R~\citep{leroy2024mast3r}, this paradigm leverages large-scale transformer backbones to directly infer 3D scene properties from arbitrary image collections without requiring prior camera calibration. Building upon this paradigm, models such as the Visual Geometry Grounded Transformer (VGGT)~\citep{wang2025vggt} and Depth Anything 3 (DA3)~\citep{lin2025da3} establish a new state-of-the-art by enabling the single-pass joint inference of dense depth and camera parameters directly from uncalibrated viewpoints. Rather than relying on classic multi-view stereo systems, we directly exploit these models. Specifically, we employ DA3 to extract dense depth maps and target camera poses, supplying the essential geometric priors required to ground our generative novel view synthesis process.

\textbf{Novel View Synthesis Evaluation.}
Standard novel-view synthesis (NVS) evaluation relies on two complementary metric families to assess generation quality and alignment. First, \emph{2D image-quality metrics}-such as PSNR, SSIM~\citep{wang2004ssim}, LPIPS~\citep{zhang2018lpips} and FID~\citep{heusel2017fid} compare predicted views against ground-truth renderings~\citep{rahary2026ovie, jang2026gld, szymanowicz2026lagernvs, ren2025gen3c, elata2024vivid, szymanowicz2025flash3d, yu2023photonvs} by quantifying pixel-level fidelity and perceptual realism. Second, \emph{camera-pose accuracy} evaluates spatial alignment by comparing Structure-from-Motion (SfM)~\citep{schonberger2016sfm} poses estimated from generated views against the ground-truth camera conditioning~\citep{he2024cameractrl, song2026worldforge, yu2024viewcrafter}. Together, these metrics form the standard NVS benchmarking pipeline, but operate in 2D image space or along camera trajectories rather than measuring intrinsic 3D coherence. As such, they have fundamental limitations for unconstrained 3D generation. In particular, 2D metrics require ground-truth views that are often unavailable in practice and cannot capture cross-view geometric consistency -individually photorealistic yet geometrically incoherent views can score highly, while perfectly 3D-consistent predictions are penalized for plausible deviations in occluded regions in the conditioning view. Furthermore, camera-pose metrics only verify alignment to the camera conditioning prompt, not the actual structural integrity of the generated scene.

%% file: main/PCC/main.tex
\subsection{Projection Consistency Constraint: A 3D Consistency Measure for Novel-view Synthesis}
\label{sec:pcc_analysis}

\input{main/PCC/introduction}

%% file: main/PCC/introduction.tex
\begin{figure}[t]
\centering
\includegraphics[width=0.9\linewidth]{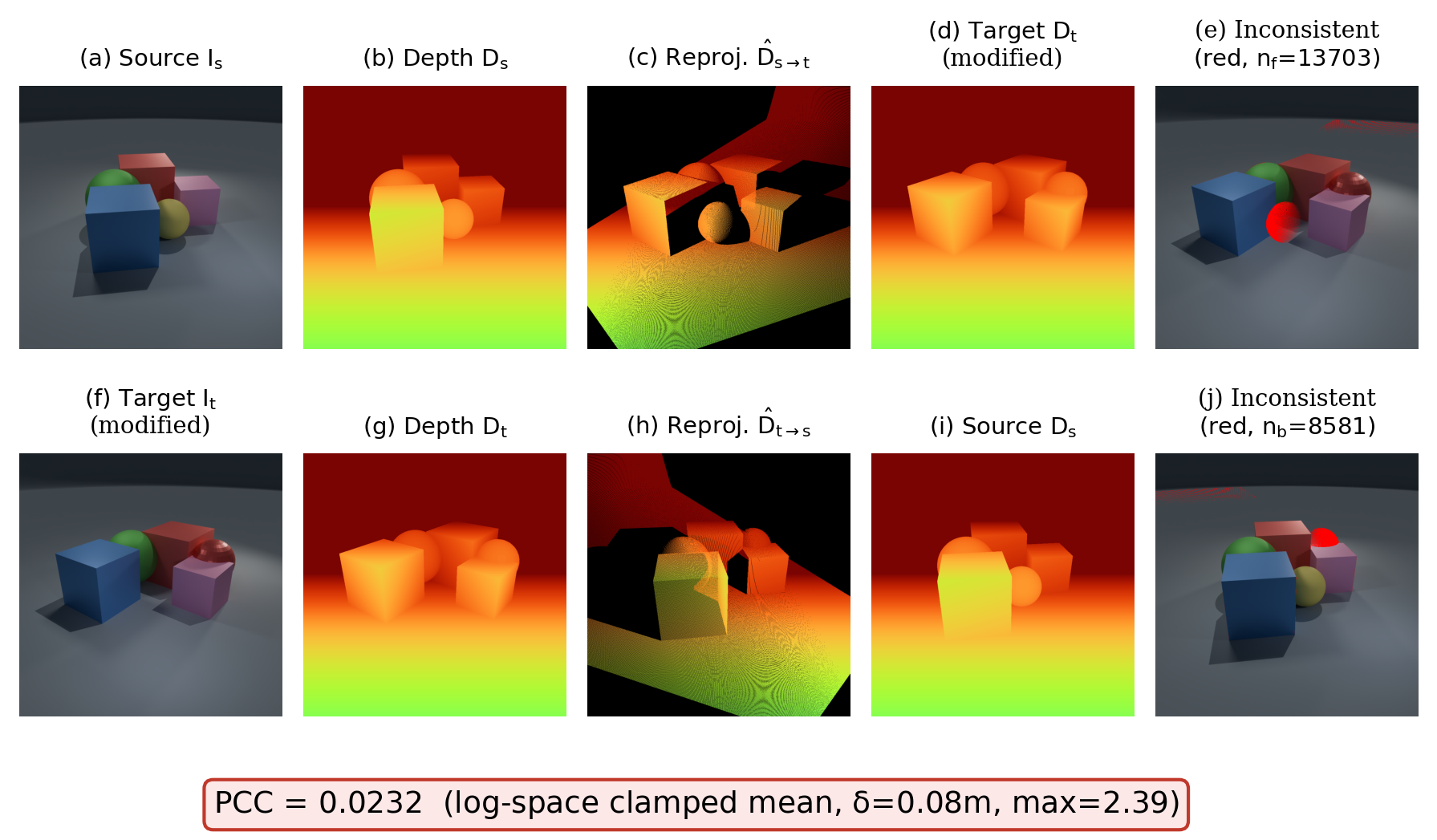}

\caption{\textbf{PCC computation pipeline.} The source view~(a) is from the original scene, while the target view~(f) has been modified: a red sphere is added and a yellow sphere is removed. \textbf{Top row ($\Delta_t$, detecting missing geometry):} the source depth~(b) is unprojected to 3D and reprojected into the target camera frame~(c); black regions indicate pixels outside the reprojection coverage. Where the yellow sphere was removed, target camera rays pass through and hit the cubes in the back instead of striking the sphere surface, producing large positive $\Delta_t$ values. These inconsistent pixels are highlighted in red~(e), with $n_f = 13703$ pixels exceeding the error threshold $\delta$. \textbf{Bottom row ($\Delta_s$, detecting hallucinated geometry):} The modified target depth is reprojected back into the source frame~(h). The added red sphere reprojects into a region with no corresponding geometry in the source perspective, yielding large positive $\Delta_s$ values and $n_b = 8581$ inconsistent pixels~(j). The final PCC score of $0.0232$ quantifies the overall geometric disagreement between the two views.\looseness-1}

\label{fig:pcc_pipeline}
\end{figure}

We propose \emph{Projection Consistency Constraint} (PCC), a bidirectional depth-reprojection metric that measures 3D geometric agreement between views (\Cref{fig:pcc_pipeline}). Given a source view with depth map $(\mathbf{I}_s, \mathbf{D}_s)$ and a target view with depth map $(\mathbf{I}_t, \mathbf{D}_t)$, with known camera intrinsics $\mathbf{K}$ and relative pose $[\mathbf{R} \mid \mathbf{t}]$—where $\mathbf{R}$ is the rotation matrix and $\mathbf{t}$ is the translation vector mapping the source camera coordinate frame to the target camera coordinate frame—PCC proceeds in three steps.

\textbf{Step 1: Unproject.} Each pixel $(u, v)$ in the source view is lifted to a 3D point in the source camera frame using the depth value and the pinhole model:
\begin{equation}
\mathbf{p}^{(s)}_{u,v} = \mathbf{D}_s(u,v) \cdot \mathbf{K}^{-1} \begin{pmatrix} u \\ v \\ 1 \end{pmatrix}.
\end{equation}
where $\mathbf{D}_s(u,v)$ represents the depth at pixel $(u,v)$. An identical unprojection procedure is applied to the target view using $\mathbf{D}_t$ to obtain the target 3D points, $\mathbf{p}^{(t)}_{u,v}$. When evaluating synthesized views where ground-truth geometry is unavailable, we estimate the depth and camera intrinsics using a pre-trained monocular depth predictor, such as Depth Anything 3 (DA3)~\citep{lin2025da3}.

\textbf{Step 2: Reproject.} The source 3D points $\mathbf{p}^{(s)}$ are transformed into the target camera frame via the relative pose and projected onto the target image plane. Z-buffer splatting~\citep{zwicker2001zbuffersplat} is used to handle occlusion: points are sorted by depth (farthest first) so that closer points overwrite farther ones, producing a reprojected depth map $\hat{\mathbf{D}}_{s \to t}$ (\Cref{fig:pcc_pipeline}(c)). The same procedure is applied symmetrically to the target 3D points $\mathbf{p}^{(t)}$, which are reprojected into the source camera frame to produce $\hat{\mathbf{D}}_{t \to s}$ (\Cref{fig:pcc_pipeline}(h)).

\textbf{Step 3: Compare.} For each pixel covered by the reprojection, we compute the depth difference. We first define a per-pixel validity mask $\mathcal{V}$ that discards three classes of unreliable correspondences: (i)~points whose reprojected depth $\hat{z}$ is non-positive, meaning they fall \emph{behind} the target camera; (ii)~points that project outside the target image domain $\Omega_t = \{(u,v) : 0 \le u < W,\; 0 \le v < H\}$; and (iii)~points originating from invalid source depths (below a small threshold $\epsilon = 0.01$\,m). The masked depth difference is:
\begin{align}
\mathcal{V}_{s \to t}(u,v) &= \mathbf{1}\!\left[\hat{z}_{s \to t}(u,v) > 0\right] \cdot \mathbf{1}\!\left[(u,v) \in \Omega_t\right] \cdot \mathbf{1}\!\left[\mathbf{D}_s > \epsilon\right], \\
\Delta_t(u,v) &= \mathcal{V}_{s \to t}(u,v) \cdot \left(\mathbf{D}_t(u,v) - \hat{\mathbf{D}}_{s \to t}(u,v)\right).
\end{align}
The backward difference $\Delta_s$ is defined symmetrically by swapping the roles of source and target. Together, these two terms capture complementary types of geometric inconsistency.

A positive $\Delta_t$ at pixel $(u,v)$ means that the reprojected source geometry lands \emph{in front of} the actual target surface at that pixel: along the target camera ray, an object from the source view occludes a point that the target view claims to see. This detects \emph{missing geometry}---content that should be present in the target view but is absent. In \Cref{fig:pcc_pipeline}, the yellow sphere was removed from the target scene; rays that should strike the yellow sphere instead pass through to the background, producing a large positive $\Delta_t$ in that region (panel~(e), red pixels). Conversely, a positive $\Delta_s$ means the reprojected target geometry lands \emph{in front of} the actual source surface: from the source camera's perspective, an object from the target view appears where no corresponding geometry exists in the source view. This detects \emph{hallucinated geometry}---content present in the target view that has no counterpart in the source. In our example, the red sphere was added to the target scene; when its depth is reprojected back to the source frame, the sphere geometry intrudes in front of the source background, producing a large positive $\Delta_s$ (panel~(j), red pixels). Together, $\Delta_t$ and $\Delta_s$ provide a complete bidirectional consistency check: $\Delta_t$ catches objects that disappeared or moved behind other surfaces from the target perspective, while $\Delta_s$ catches objects that appeared or moved in front from the source perspective.\looseness-1

Rather than simply counting inconsistent pixels, we compute a \emph{distance-based} score that reflects the magnitude of each inconsistency. Differences below a noise threshold $\delta$ are suppressed, and a log transformation with clamping is applied to compress large outliers while preserving the ordering of moderate errors. Specifically, the forward contribution is:
\begin{equation}
L_f = \sum_{(u,v) \in \mathcal{M}_{s \to t}} \min\!\Big(\log\!\big(1 + \max(\Delta(u,v) - \delta,\; 0)\big),\; \tau\Big),
\end{equation}
where $\mathcal{M}_{s \to t}$ is the set of pixels with valid reprojection coverage and $\tau$ is a clamping threshold that bounds the maximum per-pixel contribution. The process is applied symmetrically in the backward direction ($t \to s$) to obtain $L_b$. The final PCC loss is the total log-space depth error normalised by the number of valid coverage pixels across both directions:
\begin{equation}
\text{PCC} = \frac{L_f + L_b}{|\mathcal{M}_{s \to t}| + |\mathcal{M}_{t \to s}|}.
\end{equation}
Lower PCC values indicate better multi-view geometric consistency. We use $\delta = 0.08\,\text{m}$ and $\tau = 2.39$ (saturating at $\approx$10\,m raw depth error) throughout our experiments.

%% file: main/esc.tex
\subsection{ESC: Escape the Allegory of the Cave}\label{sec:esc_main}
\begin{figure}[htbp]
    \centering
    \includegraphics[width=\linewidth]{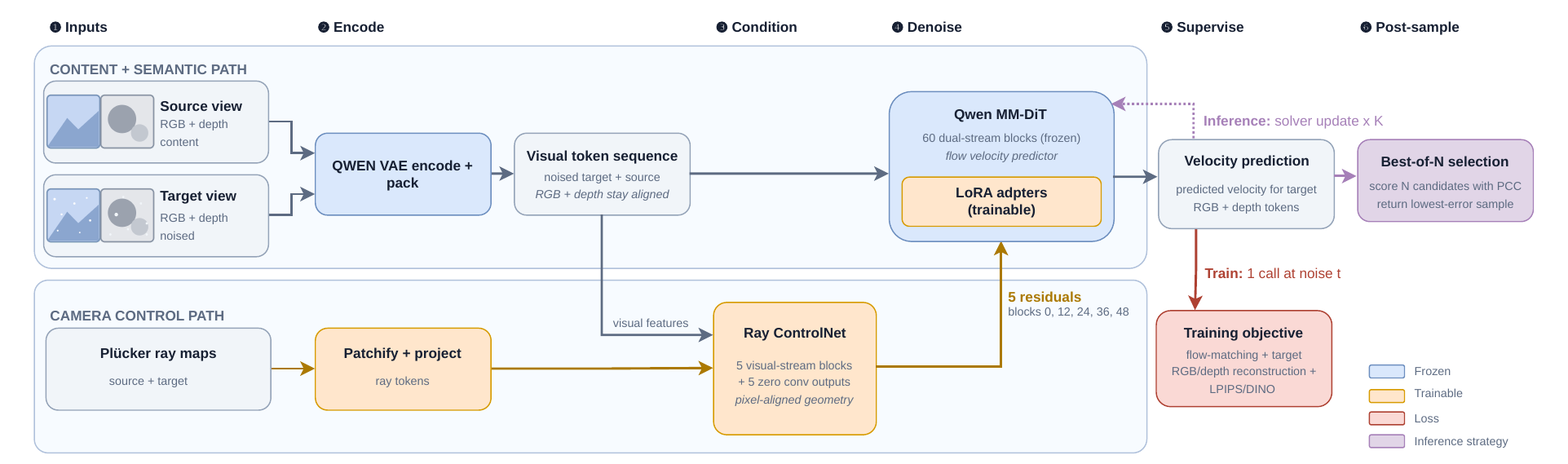}
    \caption{\textbf{\modelname~ Framework.} The framework extends Qwen MM-DiT with depth information and Pl\"ucker rays conditioning by leveraging LoRA adapters and a Ray ControlNet. At inference time, we leverage PCC-based sampling to enforce improved generated scene geometry.\looseness-1}
    \label{fig:esc_arch}
\end{figure}
We present \modelname, a geometry-aware diffusion framework for novel-view synthesis that simultaneously predicts target RGB images and depth maps (\Cref{fig:esc_arch}). We first describe our input representation, followed by our parameter-efficient adaptation strategy and the overall training and inference pipeline.

\textbf{Input Representation and Denoising Pipeline.}
\label{sub:input_representation}
% \ds{Consider talking about the data used for training before the architecture, as it isn't clear if the depth maps and camera geometry are estimated with another model or originally present in the data. You might also want to quickly mention what the camera geometry info is in a bit more detail. Not necessarily the math, just in words in a sentence, at most 2.} 
To perform geometry-aware view synthesis, our framework ingests three primary conditioning inputs: a reference source image, its corresponding rendered depth map, and the geometry of the source and target camera parameterized by Pl\"ucker ray maps. The framework also ingests a noise sample representing the target image and depth map. Note that at training time, the noise sample corresponds to a noisy version of the target groundtruth and depth.

The source image and the rendered depth map are processed and tokenized using Qwen VAE tokenizer~\citep{wu2025qwenimage, kingma2022vae}. Similarly, the noise sample corresponding to the target image and depth is fed to Qwen VAE. The resulting (source or target) depth tokens are concatenated directly to the end of the (source or target) image tokens, creating a visual token sequence. 

The Pl\"ucker rays~\citep{Zisserman2004MVCV}, which represent the camera poses of both source and target views, are tokenized and fed to a Ray ControlNet module~\citep{zhang2023controlnet} along with the visual token sequence to guide the subsequent synthesis process with explicit spatial awareness. 

The synthesis process is performed through the Qwen Multimodal-Diffusion Transformer (Qwen MM-DiT) backbone~\citep{peebles2023dit, wu2025qwenimage}, which learns to perform iterative token denoising under the combined conditioning of the visual token sequence and Ray ControlNet output. 

\textbf{Parameter-Efficient Adaptation for View Synthesis.}
\label{sec:eac:training} Optimization of the framework is performed through a Low-Rank Adaptation (LoRA)~\citep{hu2021lora} of the Qwen MM-DiT network and the Ray ControlNet network. Specifically, parameter-efficient LoRA adapters are injected into the dual-stream image branch to align the newly introduced depth tokens with native visual features, while the Ray ControlNet is utilized for precise Pl\"ucker ray conditioning. This targeted configuration provides the necessary expressive capacity to incorporate geometric and spatial controls without destabilizing the foundation model.

The training loss $\mathcal{L}_{\text{base}}$ is formulated as a weighted sum of multi-modal objectives, where pixel-space losses are gated by the flow-matching timestep $t$:
\begin{equation}
\mathcal{L}_{\text{base}} = w_{\text{fm,rgb}}\mathcal{L}_{\text{fm,rgb}} + w_{\text{fm,}d}\mathcal{L}_{\text{fm,}d} + \mathbf{1}_{\{t \le t_{\text{trunc}}\}} \big( w_{\text{recon}}\mathcal{L}_{\text{recon}} + w_{\text{lpips}}\mathcal{L}_{\text{lpips}} + w_{\text{dino}}\mathcal{L}_{\text{dino}} \big),
\end{equation}
where the $w_{(\cdot)}$ coefficients denote scalar weights controlling the relative contribution of each respective term. $\mathcal{L}_{\text{fm,rgb}}$ and $\mathcal{L}_{\text{fm,}d}$ represent the latent flow-matching objectives for the generated target RGB and target inverse depth fields, respectively, with inverse depth targets derived by inverting the pseudo-ground-truth depth estimates provided by DA3\footnote{Inverse depth $d$ is computed from the depth $z$ as $d = \frac{1}{z}$.}. To ensure high visual fidelity, the remaining terms are computed in pixel space on the decoded target views: $\mathcal{L}_{\text{recon}}$ denotes a pixel reconstruction loss, while $\mathcal{L}_{\text{lpips}}$~\citep{zhang2018lpips} and $\mathcal{L}_{\text{dino}}$~\citep{oquab2024dinov2} serve as perceptual loss terms. The indicator function $\mathbf{1}_{\{t \le t_{\text{trunc}}\}}$ ensures these three pixel-space objectives are evaluated exclusively within a truncated timestep interval, where $t_{\text{trunc}}$ is the chosen timestep threshold.

\textbf{Inference-time Best-of-$N$ (\modelname-PCC-BoN)}.
We leverage the stochastic nature of the diffusion sampling process to optimize geometric fidelity. For each source-to-target view synthesis request, we generate $N$ independent samples $\{\hat{y}_i\}_{i=1}^{N}$, by varying the initial random noise seeds. The predicted target inverse depth of each candidate sample is evaluated via the PCC metric established in \Cref{sec:pcc_analysis}, based on the depth estimates generated by the ESC. Notably, we disable scale alignment (mentioned in ~\Cref{app:scale_alignment_pcc}) for PCC during Best-of-$N$ selection. The pipeline then selects and returns the optimal sample $\hat{y}^{\star}$ that minimizes the geometric alignment error relative to the source reference view $x_{\text{src}}$:
\begin{equation}
\hat{y}^{\star} = \arg\min_{i \in \{1, \ldots, N\}} \operatorname{PCC}\left(\hat{y}_i,\, x_{\text{src}}\right).
\end{equation}

%% file: main/experiments.tex
\section{Experiments}
% \ars{Short paragraph on what is presented here: main results on real dataset and evaluation of PCC / how it complements existing metrics.}

In this section, we present a comprehensive evaluation of our proposed ESC framework and the Projection Consistency Constraint (PCC) metric. First, in \Cref{sec:main_results}, we benchmark our novel view synthesis approach on three complex, real-world datasets (DL3DV, RealEstate10K, and Tanks \& Temples), demonstrating that our geometry-aware ESC models consistently outperform established baselines in both perceptual realism and 3D consistency. Subsequently, in \Cref{sec:pcc_vs_metrics}, we conduct a systematic analysis of the PCC metric itself. Through controlled geometric perturbation and occlusion experiments, we illustrate how PCC successfully isolates multi-view geometric integrity, providing a crucial complement to standard 2D perceptual and camera-pose metrics that often fail to penalize structural hallucinations.

\subsection{Main results}\label{sec:main_results}
\begin{figure}[htbp] % full-page float; content stretched with \vfill to fill the page
\centering
\begin{minipage}[c][\textheight]{\textwidth}
\centering
\vspace*{\fill}

% ==================== FIRST ELEMENT (FIGURE) ====================
\renewcommand{\arraystretch}{0.35}
\begin{tabular}{@{}c@{}*{10}{@{\hspace{1.5pt}}c}@{}}
& \demohd{\textcolor{legendinput}{Source}} & \demohd{\textcolor{legendbaseline}{PhotoNVS}} & \demohd{\textcolor{legendbaseline}{VIVID}} & \demohd{\textcolor{legendbaseline}{OVIE}}
& \demohd{\textcolor{legendbaseline}{GLD}} & \demohd{\textcolor{legendbaseline}{LagerNVS}} & \demohd{\textcolor{legendours}{\textbf{\modelname-Shadow}}}
& \demohd{\textcolor{legendours}{\textbf{\modelname-Base}}} & \demohd{\textcolor{legendours}{\textbf{\modelname-PCC-BoN}}} & \demohd{\textcolor{legendtarget}{GT}} \\[1pt]
\demorow{\dldir}{DL3DV}
\demorow{\redir}{RE10K}
\demorow{\tntdir}{T\&T (Barn)}
\end{tabular}
\captionof{figure}{\textbf{Qualitative comparison on novel view synthesis.} For each scene we show the input source view (\textcolor{legendinput}{blue border}), the renderings produced by the baselines (\textcolor{legendbaseline}{gray border}: PhotoNVS, VIVID, OVIE, GLD, LagerNVS), our \modelname~variants (\textcolor{legendours}{orange border}: \modelname-Shadow, \modelname-Base, and the geometry-guided \modelname-PCC-BoN (Best-of-N)), and the ground-truth target view (\textcolor{legendtarget}{green border}). \textbf{Top:} an outdoor scene from DL3DV. \textbf{Middle:} an indoor scene from RealEstate10K. \textbf{Bottom:} the \emph{Barn} scene from Tanks \& Temples. Our variants reproduce scene geometry and fine detail more faithfully than the baselines, with \modelname-PCC-BoN best matching the ground-truth target.\looseness-1}
\label{fig:qualitative_nvs}

\vspace*{\fill}

% ==================== SECOND ELEMENT (FIGURE) ====================
\setlength{\tabcolsep}{1.5pt}
\renewcommand{\arraystretch}{0.5}
\begin{tabularx}{\textwidth}{Y Y | Y | Y Y Y Y Y}
\demohd{\textcolor{legendinput}{Source}} & \demohd{\textcolor{legendtarget}{GT}} & \demohd{\textcolor{legendours}{\textbf{Best}}} & \multicolumn{5}{c}{\demohd{\textcolor{legendours}{Candidates (Best-of-N)}}} \\
\hline
\rescellsrc{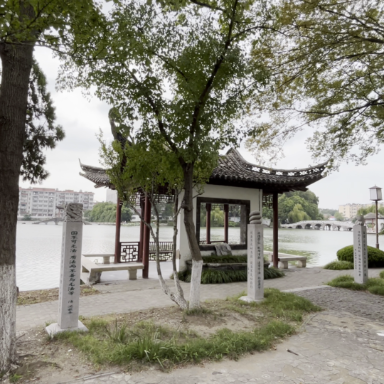} &
\rescellgt{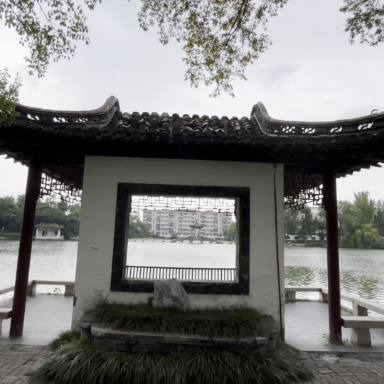} &
\rescellours{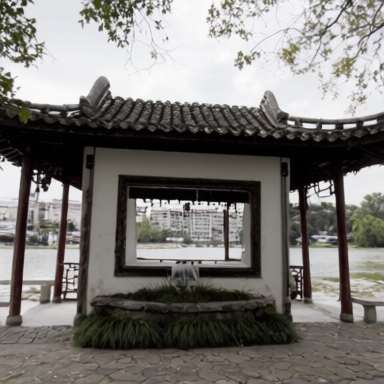} &
\rescellours{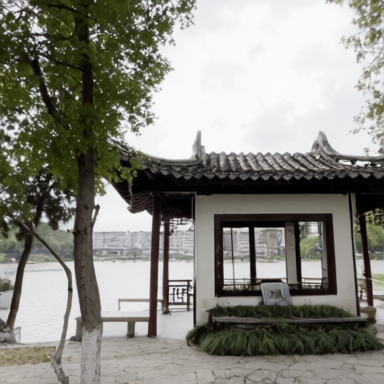} &
\rescellours{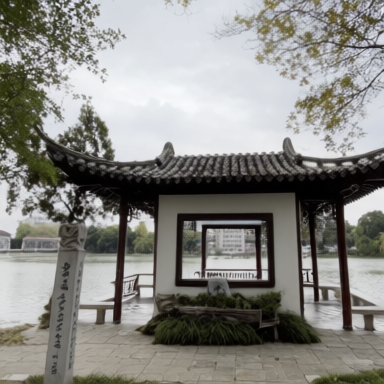} &
\rescellours{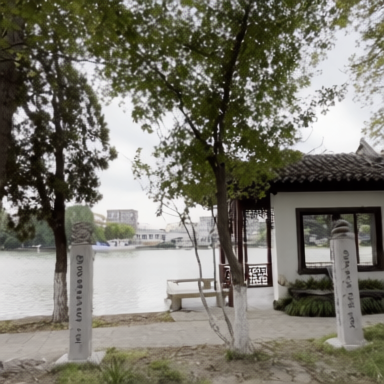} &
\rescellours{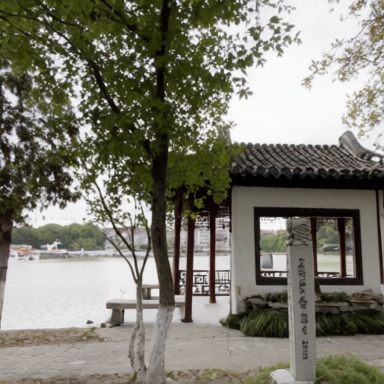} &
\rescellours{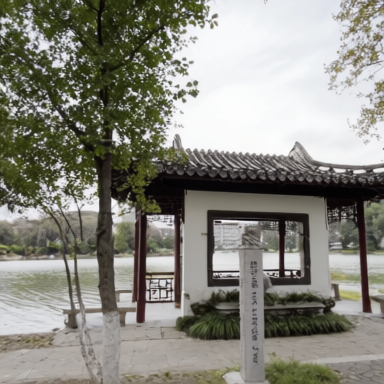} \\

\rescellsrc{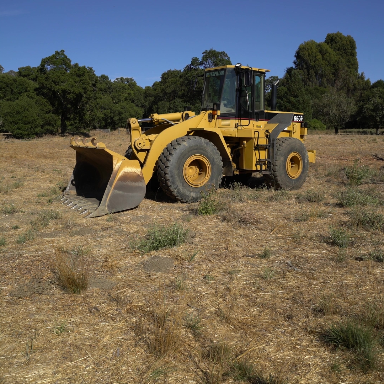} &
\rescellgt{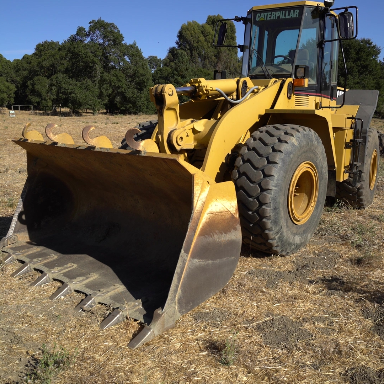} &
\rescellours{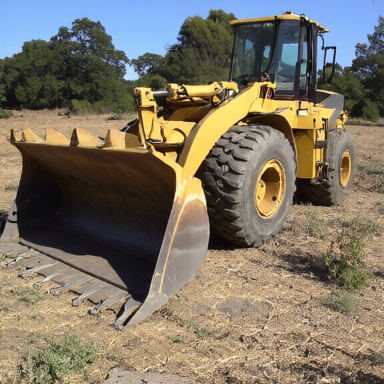} &
\rescellours{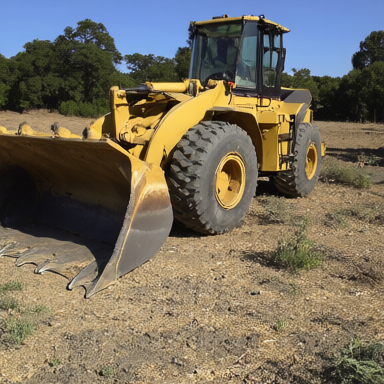} &
\rescellours{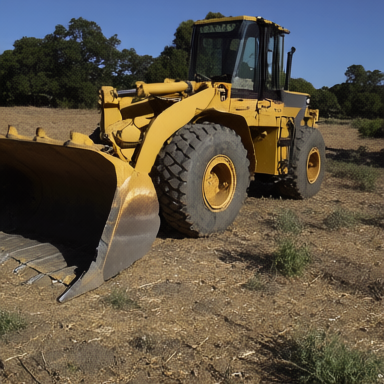} &
\rescellours{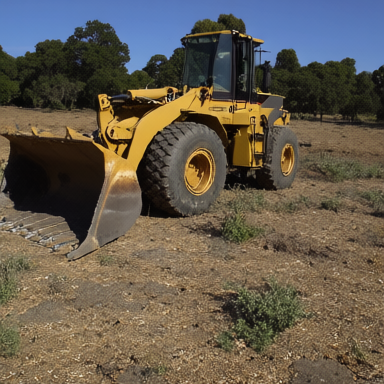} &
\rescellours{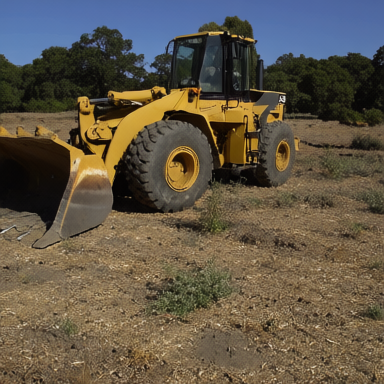} &
\rescellours{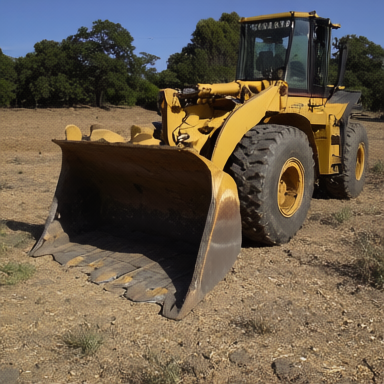} \\

\rescellsrc{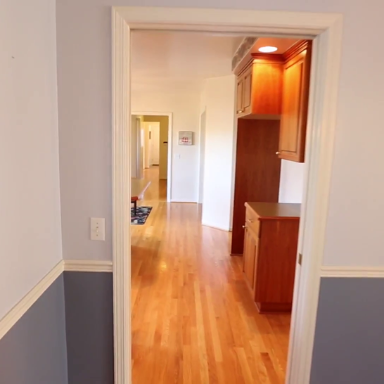} &
\rescellgt{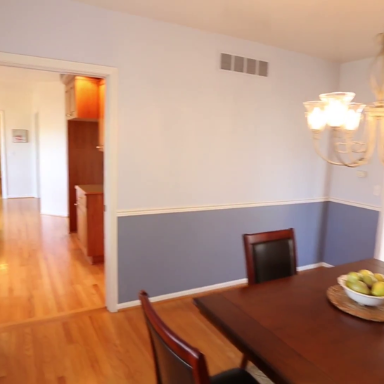} &
\rescellours{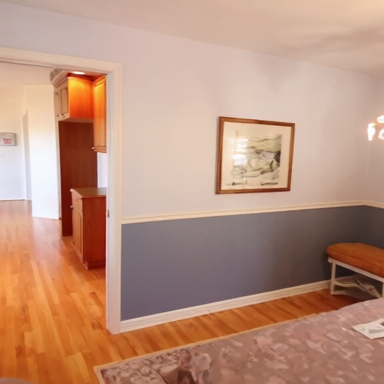} &
\rescellours{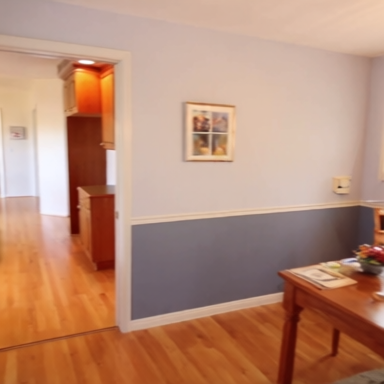} &
\rescellours{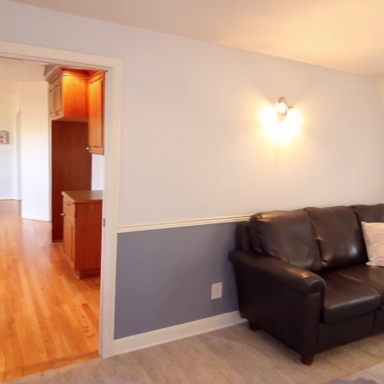} &
\rescellours{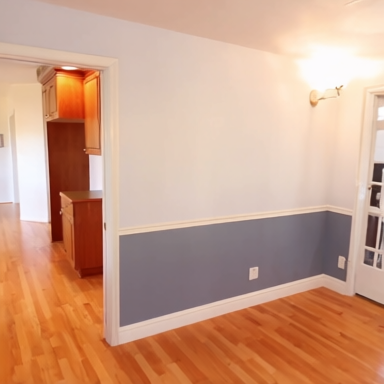} &
\rescellours{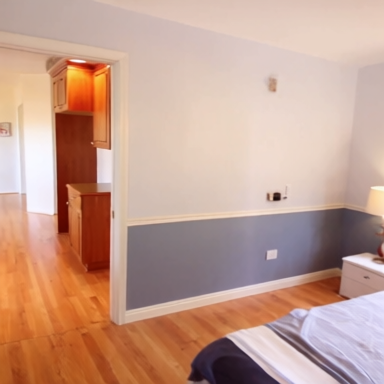} &
\rescellours{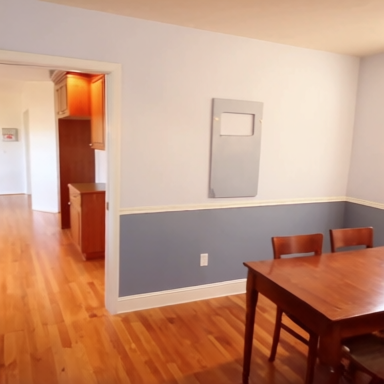} \\
\end{tabularx}
\captionof{figure}{\textbf{Best-of-N novel-view synthesis.} Each row shows the source view (\textcolor{legendinput}{blue border}), ground truth (\textcolor{legendtarget}{green border}), the PCC-selected best sample (\textcolor{legendours}{orange border}), and five additional candidates (\textcolor{legendours}{orange border}) from the same Best-of-N pool.\looseness-1}
\label{tab:best_of_n}

\vspace*{\fill}
\end{minipage}
\end{figure}

\textbf{Datasets and Preprocessing.} We evaluate our method on three datasets: DL3DV~\citep{ling2023dl3dv}, RealEstate10K (Re10K)\citep{zhou2018re10k}, and Tanks~\&~Temples (T\&T)~\citep{knapitsch2017tnt}. We leverage Depth-Anything 3 (DA3)~\citep{lin2025da3} as a preprocessing step to extract dense depth maps. A comprehensive discussion of the datasets and their specific preprocessing pipelines may be found in ~\Cref{app:sec:datasets}.\looseness-1

% We evaluate our method on three static scene video datasets: DL3DV-10k (DL3DV)~\citep{ling2023dl3dv}, RealEstate-10k (RE10k)~\citep{zhou2018re10k}, and Tank \& Temple (T\&T)~\citep{knapitsch2017tnt}. As a preprocessing step, we leverage Depth-Anything 3 (DA3)~\citep{lin2025da3} to extract dense depth maps for both the training and evaluation splits across all datasets. % Our model is trained on a joint mixture of the DL3DV and RE10k training sets, and subsequently evaluated on the test splits of each dataset individually (comprehensive dataset, split, and training details are provided in ~\Cref{app:datasets:mixed}).

\textbf{Training details.} We optimize our model parameters end-to-end using the AdamW optimizer~\citep{loshchilov2018adamw} with weight decay, coupled with a cosine learning-rate schedule and an initial warm-up phase. To stabilize training and enhance generation quality, we maintain an exponential moving average (EMA)~\citep{ho2020ddpm, polyak1992accelerate} of the trainable parameters, which is subsequently used for all evaluations. Throughout both stages, diffusion timesteps are sampled from a logit-normal noise distribution. For real image datasets, our model is trained on a joint mixture of the DL3DV and RE10k training sets and evaluated on DL3DV, Re10K and T\&T. All the training configurations are presented in \Cref{tab:mixed_training_config}.

\textbf{Metrics.} Performance is assessed across four core metric categories:
(i)~\emph{Synthesis Quality}, measured via PSNR and SSIM~\citep{wang2004ssim};
(ii)~\emph{Perceptual Quality}, evaluated using LPIPS~\citep{zhang2018lpips}, DINO-Sim~\citep{oquab2024dinov2}, and DreamSim~\citep{fu2023dreamsim};
(iii)~\emph{Pose/View Prompt Consistency}, tracking pose-AUC, RPE-R, and RPE-T~\citep{he2024cameractrl, song2026worldforge}; and
(iv)~\emph{Geometric Consistency}, assessed using our proposed PCC.

\textbf{Models.} To systematically evaluate the performance of our architecture and its ability to scale to complex, unconstrained real-life scenes, we train and evaluate two core model configurations. Specifically, \emph{\modelname-Base} represents our framework, utilizing depth conditioning to generate a \emph{single} sample per target view. Finally, \emph{\modelname-PCC-BoN} employs our complete framework with inference-time Best-of-$N$ strategy, which generates \emph{multiple} candidate samples and selects the optimal output according to our PCC metric.

% ---------------------------------------------------------------------------
% Qualitative comparison figure (two scenes x ten methods).
% Tune \demoimg if the row overflows \textwidth; labels scale with the font
% commands (\scriptsize) in the header / rotated row labels.
% ---------------------------------------------------------------------------
% \subsection{In-the-Wild Training and Evaluation Results}
% Ensure you have \usepackage{caption} in your preamble to use \captionof

\begin{table*}[t!]
\centering
\footnotesize

\begin{tabular}{l ccc c}
\toprule
& \multicolumn{3}{c}{\textbf{Perceptual Quality}} & \textbf{Geometry} \\
\cmidrule(r){2-4} \cmidrule(l){5-5}
\textbf{Model} & LPIPS $\downarrow$ & DINO-Sim $\uparrow$ & DreamSim $\downarrow$ & PCC $\downarrow$ \\
\midrule

\multicolumn{5}{l}{\textbf{DL3DV Dataset}} \\
\midrule
PhotoNVS & $0.5743 \pm 0.1229$ & $0.5647 \pm 0.0900$ & $0.3876 \pm 0.1293$ & $0.2628 \pm 0.1677$ \\
VIVID & $0.6407 \pm 0.1047$ & $0.6909 \pm 0.2080$ & $0.2953 \pm 0.1288$ & \third{$0.2475 \pm 0.1624$} \\
OVIE & $0.7099 \pm 0.1490$ & $0.2755 \pm 0.3507$ & $0.6480 \pm 0.2730$ & $0.3122 \pm 0.1598$ \\
GLD$\dagger$ & $0.5665 \pm 0.1013$ & \third{$0.7019 \pm 0.1499$} & \third{$0.2549 \pm 0.1175$} & $0.2487 \pm 0.1642$ \\
LagerNVS$\dagger$ & \third{$0.4714 \pm 0.1761$} & $0.6734 \pm 0.2334$ & $0.3231 \pm 0.2035$ & $0.2487 \pm 0.1750$ \\
\modelsep
% \modelname-shadow$\dagger$ & $0.4952 \pm 0.1386$ & \third{$0.8035 \pm 0.1540$} & \third{$0.1676 \pm 0.1304$} & $0.2533 \pm 0.1640$ \\
\modelname-Base$\dagger$ (\textit{Ours}) & \second{$0.4448 \pm 0.1569$} & \second{$0.8366 \pm 0.1389$} & \second{$0.1420 \pm 0.1252$} & \second{$0.2448 \pm 0.1646$} \\
\modelname-PCC-BoN$\dagger$ (\textit{Ours}) & \first{$0.4340 \pm 0.1556$} & \first{$0.8386 \pm 0.1388$} & \first{$0.1402 \pm 0.1232$} & \first{$0.2257 \pm 0.1556$} \\
% \modelsep
% Oracle & $\cdot$ & $\cdot$ & $\cdot$ & $0.1199 \pm 0.1539$ \\

\midrule
\multicolumn{5}{l}{\textbf{RE10K Dataset}} \\
\midrule
PhotoNVS$\dagger$ & $0.3660 \pm 0.1230$ & $0.7669 \pm 0.1027$ & $0.2176 \pm 0.0672$ & $0.0899 \pm 0.1631$ \\
VIVID$\dagger$ & $0.3741 \pm 0.1406$ & \third{$0.8841 \pm 0.0937$} & \third{$0.1088 \pm 0.0659$} & $0.0879 \pm 0.1480$ \\
OVIE & $0.5009 \pm 0.1815$ & $0.7168 \pm 0.3253$ & $0.7881 \pm 0.1383$ & $0.1344 \pm 0.1521$ \\
GLD$\dagger$ & $0.4413 \pm 0.1054$ & $0.8128 \pm 0.1406$ & $0.1394 \pm 0.1046$ & $0.0898 \pm 0.1430$ \\
LagerNVS$\dagger$ & \third{$0.3239 \pm 0.1248$} & $0.8838 \pm 0.1200$ & $0.1256 \pm 0.1030$ & \third{$0.0783 \pm 0.1490$} \\
\modelsep
% \modelname-shadow$\ddagger$ & \third{$0.3166 \pm 0.1328$} & \third{$0.8982 \pm 0.0766$} & \third{$0.0780 \pm 0.0568$} & $0.0870 \pm 0.1433$ \\
\modelname-Base$\ddagger$(\textit{Ours}) & \second{$0.2613 \pm 0.1249$} & \second{$0.9157 \pm 0.0652$} & \second{$0.0669 \pm 0.0497$} & \second{$0.0773 \pm 0.1430$} \\
\modelname-PCC-BoN$\ddagger$(\textit{Ours}) & \first{$0.2150 \pm 0.1220$} & \first{$0.9222 \pm 0.0667$} & \first{$0.0655 \pm 0.0510$} & \first{$0.0656 \pm 0.1359$} \\
% \modelsep
% Oracle & $\cdot$ & $\cdot$ & $\cdot$ & $0.0539 \pm 0.1380$ \\
\midrule
\multicolumn{5}{l}{\textbf{Tanks \& Temples Dataset}} \\
\midrule
PhotoNVS & $0.7688 \pm 0.0776$ & $0.2345 \pm 0.1670$ & $0.5902 \pm 0.1289$ & $0.4467 \pm 0.4110$ \\
VIVID & $0.7527 \pm 0.0818$ & $0.3793 \pm 0.2351$ & $0.5040 \pm 0.1699$ & $0.4416 \pm 0.4231$ \\
OVIE & $0.8219 \pm 0.0977$ & $0.0640 \pm 0.1515$ & $0.7881 \pm 0.1383$ & $0.5102 \pm 0.4792$ \\
GLD & \third{$0.6365 \pm 0.0731$} & \third{$0.5099 \pm 0.2075$} & \third{$0.4027 \pm 0.1454$} & \third{$0.4184 \pm 0.4089$} \\
LagerNVS & $0.6420 \pm 0.0994$ & $0.2738 \pm 0.2667$ & $0.6630 \pm 0.1963$ & $0.4901 \pm 0.4861$ \\
\modelsep
% \modelname-shadow & \third{$0.6336 \pm 0.0859$} & \third{$0.6549 \pm 0.2078$} & \third{$0.2928 \pm 0.1512$} & $0.4291 \pm 0.4094$ \\
\modelname-Base (\textit{Ours}) & \second{$0.5835 \pm 0.1016$} & \second{$0.7242 \pm 0.1872$} & \second{$0.2362 \pm 0.1371$} & \second{$0.3826 \pm 0.3875$} \\
\modelname-PCC-BoN (\textit{Ours}) & \first{$0.5596 \pm 0.1164$} & \first{$0.7339 \pm 0.1854$} & \first{$0.2232 \pm 0.1343$} & \first{$0.2849 \pm 0.3139$} \\
% \modelsep
% Oracle & $\cdot$ & $\cdot$ & $\cdot$ & $0.1899 \pm 0.2478$ \\
\bottomrule
\end{tabular}
\caption{\textbf{Quantitative comparison on novel view synthesis across three datasets.} We evaluate perceptual quality using LPIPS, DINO similarity, and DreamSim, and geometric consistency using PCC, on the DL3DV, RE10K, and Tanks \& Temples datasets. Arrows indicate whether higher ($\uparrow$) or lower ($\downarrow$) is better. We compare our method (\modelname) and its variants against prior baselines (PhotoNVS, VIVID, OVIE, GLD, LagerNVS), with results reported as mean $\pm$ standard deviation. The \first{best}, \second{second-best}, and \third{third-best} results in each column are highlighted. \modelname~ consistently outperforms all baselines across perceptual metrics, and the geometry-guided \modelname-PCC-BoN variant further improves geometric consistency.  Models marked with $\dagger$ were trained on the training split of the dataset.  Models marked with $\ddagger$ were trained on a restricted subset comprising less than half of the full training split.}

\label{tab:perceptual_geometry_benchmarks}
\end{table*}

\textbf{Baselines.} We benchmark our approach against five baselines: PhotoNVS~\citep{yu2023photonvs}, VIVID~\citep{elata2024vivid}, OVIE~\citep{rahary2026ovie}, GLD~\citep{jang2026gld} and LagerNVS~\citep{szymanowicz2026lagernvs}.\looseness-1

\textbf{Discussions.}
Quantitative results are compiled in \Cref{tab:perceptual_geometry_benchmarks} and \Cref{tab:full_comprehensive_landscape_benchmarks_no_std}. \modelname-Base achieves the strongest LPIPS, DINO-Sim, and DreamSim results across the three evaluated datasets. PCC-based Best-of-$N$ selection provides additional improvements in these perceptual metrics and lowers PCC. However, GLD and LagerNVS perform better on several pixel-level reconstruction and pose-consistency metrics. Taken together, these results indicate that \modelname~improves perceptual similarity and depth-reprojection consistency, while some competing approaches retain advantages under conventional reconstruction and pose metrics.

The differences across metrics reflect their complementary evaluation criteria. PSNR and SSIM emphasize pixel-aligned reconstruction, while the pose metrics evaluate agreement with the target camera; neither directly measures bidirectional depth-reprojection consistency. We therefore report these metrics alongside PCC rather than treating any single metric as definitive. The corresponding qualitative comparisons are provided in \Cref{fig:qualitative_nvs}, \Cref{fig:qualitative_nvs_dl3dv_tnt}, and \Cref{fig:qualitative_nvs_re10k}.

We attribute ESC's performance to two complementary components: \emph{geometry-aware RGB--depth generation} and \emph{PCC-guided inference-time selection}.

First, unlike methods that predict only RGB, ESC jointly predicts RGB and depth while conditioning on Pl\"ucker rays through the Ray ControlNet. This design provides explicit geometric signals during generation and encourages synthesized appearance to remain coupled to the predicted scene structure.

Second, PCC-based Best-of-$N$ selection provides an inference-time mechanism for converting additional sampling compute into improved depth-reprojection consistency. ESC-PCC-BoN generates $N$ independent diffusion samples and selects the candidate with the lowest PCC. Relative to the single-sample ESC-Base setting, this requires approximately $N$ times the denoising compute, plus the comparatively small cost of PCC evaluation. ESC-PCC-BoN should therefore be interpreted as a test-time scaling result rather than a compute-matched comparison with single-sample baselines; ESC-Base provides the corresponding
one-sample comparison.

As shown in \Cref{tab:full_comprehensive_landscape_benchmarks_no_std}, this selection improves both perceptual and geometric metrics. On DL3DV, moving from ESC-Base to ESC-PCC-BoN reduces LPIPS from $0.4448$ to $0.4340$, DreamSim from $0.1420$ to $0.1402$, and PCC from $0.2448$ to $0.2257$, while increasing DINO-Sim from $0.8366$ to $0.8386$. Thus, PCC selection provides a measurable quality--compute trade-off, although total sampling compute grows approximately linearly with $N$.

% while qualitative visualizations are presented in \Cref{fig:qualitative_nvs}, \Cref{fig:qualitative_nvs_dl3dv_tnt}, and \Cref{fig:qualitative_nvs_re10k}. 

% \textbf{Quantitative and Qualitative Analysis.} 

It is worth noting that our model excels at generating diverse, highly plausible completions for occluded regions within a scene. Rather than producing a single deterministic or blurred average, our framework can output multiple distinct, high-quality variations from identical source inputs and conditioning states (see multi-sample variations in ~\Cref{tab:best_of_n}, ~\Cref{tab:best_of_n_dl3dv}, and ~\Cref{tab:best_of_n_re10k}). As shown, our \modelname-PCC-BoN leads to improved predictions that are closer to the ground-truth view. Additional results on a custom synthetic Blender simulation dataset with  ground truth depth maps may be found in~\Cref{sub:simulation_results}.

%% file: main/ablation/main.tex
%\section{Ablations}
% In this section, we perform ablation experiments to evaluate the impact of our key framework components. Unless stated otherwise, we perform the ablations on a custom synthetic Blender simulation dataset featuring simple geometric primitives~\citep{blender, greff2021kubric}..

\subsection{PCC vs Standard Evaluation Metrics} \label{sec:pcc_vs_metrics}

A reliable 3D metric must correctly penalize geometric distortions while remaining invariant to purely visual changes in occluded regions. To prove PCC meets these criteria, we systematically evaluate its performance against established 2D perceptual and camera-pose metrics. Using controlled synthetic scenes, we demonstrate that PCC captures the true spatial consistency that standard evaluation protocols fundamentally miss. Unless stated otherwise, we perform the evaluations on a custom synthetic Blender simulation dataset featuring simple geometric primitives~\citep{blender, greff2021kubric}.

\subsubsection{Evaluating PCC Against 2D Perception Metrics}\label{sec:blender_experiments}

\input{main/PCC/pcc_vs_2D_metrics}

\subsubsection{Evaluating PCC Against Camera Pose Metrics}

\input{main/PCC/pcc_vs_camera_pose_metrics}

\input{main/ablation/frame_gap_analysis}
\input{main/ablation/eac_shadow}

\section{Conclusion}
In this paper, we presented ESC, a geometry-aware diffusion framework designed to overcome the inherent 3D ambiguities of monocular novel view synthesis. By moving beyond pure 2D image-plane inpainting, ESC explicitly grounds the generative process in 3D space through joint RGB and depth prediction, guided by Plücker ray conditioning. To strictly enforce geometric fidelity, we introduced the PCC, a novel bidirectional depth-reprojection mechanism that evaluates the 3D spatial validity of every generated pixel to guarantee structural agreement between the source and target views. Our extensive evaluations demonstrate that PCC serves as both a highly robust metric—capturing true multi-view 3D consistency where standard 2D perceptual and camera-pose metrics fail—and an effective objective for inference-time Best-of-$N$ selection. Ultimately, by coupling the strong visual priors of modern diffusion models with explicit geometric verification, our framework successfully escapes the constraints of 2D ambiguity, yielding novel views that are both photorealistic and rigorously structurally coherent.\looseness-1

%% file: main/PCC/pcc_vs_2D_metrics.tex
\begin{figure}[t]
\centering
\includegraphics[width=0.9\linewidth]{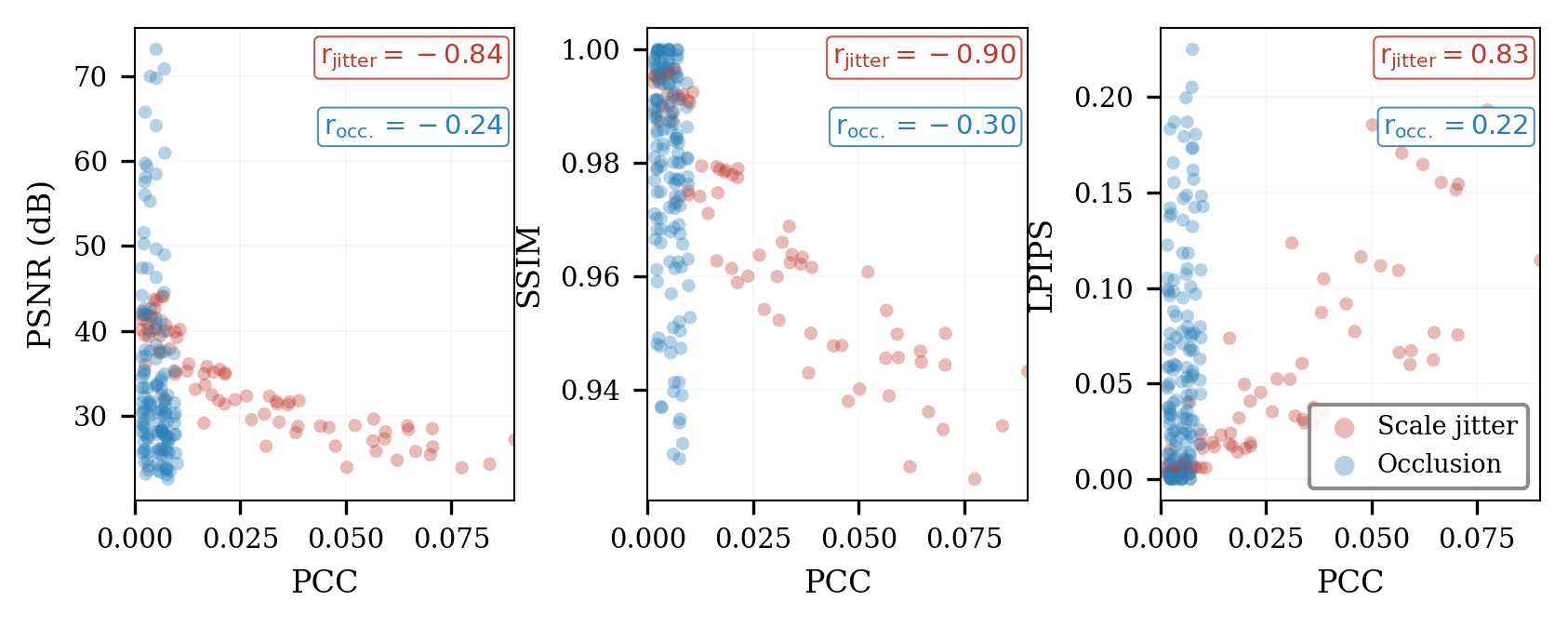}
\caption{\textbf{Metric correlation analyses.} Per-view correlation between PCC and 2D metrics across both experiments. Each point represents a single (variant, view) pair. \textcolor{pcccolor}{Red}: scale jitter experiment—PCC correlates strongly with all 2D metrics ($r{=}{-}0.84$ for PSNR, ${-}0.90$ for SSIM, ${+}0.83$ for LPIPS), since geometric distortion simultaneously changes appearance. \textcolor{psnrcolor}{Blue}: occlusion experiment—correlations weaken substantially ($r{=}{-}0.24$ for PSNR, ${-}0.30$ for SSIM, ${+}0.22$ for LPIPS), as views with large visual differences (high LPIPS, low PSNR) maintain low PCC because each scene is internally 3D-consistent. The strong-to-weak correlation shift between experiments demonstrates that PCC and 2D metrics measure fundamentally orthogonal properties.\looseness-1}
\label{fig:pcc_correlation}
\end{figure}

We design two controlled sets of experiment using synthetic scenes rendered with 15 camera views distributed on a hemisphere around a scene of objects.

\textbf{Experiment~1: Scale jitter.} We render a baseline scene containing five geometric primitives and create six \emph{jittered} variants by applying per-object random scale jitter of $\pm j\%$, with $j \in \{1, 2, 5, 10, 20, 40\}$. Each object is independently scaled by a factor drawn uniformly from $[1{-}j/100,\; 1{+}j/100]$, so some objects enlarge while others shrink within the same scene. Since objects change size, both the rendered appearance and the underlying 3D geometry are altered relative to the baseline.  We compute two types of metrics comparing the jittered and baseline scenes: (1) 2D perceptual metrics (PSNR, SSIM, LPIPS). Note that for the unmodified baseline scene, these metrics are trivially perfect ($\text{PSNR}=\infty$, $\text{SSIM}=1$, $\text{LPIPS}=0$). (2) Our PCC metric, computed by reprojecting the baseline anchor view's depth (view~0) into each jittered variant's novel view $i$: $\text{PCC}(\mathbf{D}^{\text{base}}_0, \mathbf{D}^{\text{jitter}_j}_i, \mathbf{T}_{0 \to i})$, where $\mathbf{T}_{0 \to i}\!\in\!\mathrm{SE}(3)$ denotes the relative camera transformation matrix mapping from the coordinate frame of view~0 to view~$i$. This measures whether the baseline geometry and the jittered variant geometry agree when viewed from the same camera.

\textbf{Experiment~2: Occlusion-invariant.} We design a scenario to expose the limitations of 2D perceptual metrics in evaluating 3D consistency. From a baseline scene~A, we create 10 modified variants (scene~B$_j$, $j=1,\ldots,10$) by introducing or removing objects within regions unobserved by the anchor camera. By design, the rendered images from the anchor viewpoint remain nearly indistinguishable across all variants (achieving a PSNR $>50$ dB compared to scene~A).~\footnote{The slight variance in PSNR results solely from subtle changes in global lighting reflections (see leftmost column in~\Cref{fig:occlusion_comparison}).} The 2D perceptual metrics compare view~$i$ of scene~A directly against view~$i$ of scene~B$j$. Because the actual scene content differs, these metrics should heavily penalize the novel viewpoints where the new objects are revealed. %; %specifically, PSNR drops to $24$--$30$,dB, SSIM decreases to $0.93$--$0.98$, and LPIPS rises to $0.01$--$0.22$. 
Conversely, PCC evaluates geometric consistency by bidirectionally reprojecting depth between the anchor view and the novel views. Because the added or removed objects in scene~B$j$ are strictly occluded from the anchor perspective, they introduce no depth conflicts or geometric contradictions between the two views. As a result, PCC should correctly recognize that the multi-view geometry remains perfectly aligned, yielding values nearly identical to the baseline.

% Because the actual scene content differs, these metrics heavily penalize the novel viewpoints where the new objects are revealed; specifically, PSNR drops to $24$--$30$,dB, SSIM decreases to $0.93$--$0.98$, and LPIPS rises to $0.01$--$0.22$. Conversely, PCC evaluates geometric consistency by bidirectionally reprojecting depth between the anchor view and the novel views. Because the added or removed objects in scene~B$j$ are strictly occluded from the anchor perspective, they introduce no depth conflicts or geometric contradictions between the two views. As a result, PCC correctly recognizes that the multi-view geometry remains perfectly aligned, yielding values nearly identical to the baseline ($\text{PCC}{\text{baseline}} \approx 0.005$ vs. $\text{PCC}{\text{occlusion}} \approx 0.005$, as shown in \Cref{tab:pcc_summary}). For details, please refer to \Cref{app:occlusion_test}.

\textbf{Analysis (\Cref{fig:pcc_correlation}).} The per-view scatter plots of PCC against each 2D perceptual metric reveal the critical distinction between the two experiments. Under scale jitter (red points), PCC and the 2D perceptual metrics are strongly correlated: as geometric distortion increases, both PCC and LPIPS rise together ($r = +0.83$), while PCC and PSNR exhibit negative correlation since PSNR decreases with distortion ($r = -0.84$), and similarly for SSIM ($r = -0.90$). This is expected as scale jitter simultaneously changes geometry and appearance, making all metrics respond in agreement. For additional details, please refer to  \Cref{tab:pcc_summary} in \Cref{app:jitter_test}. Under occlusion (blue points), the correlation structure fundamentally changes. The blue points cluster at low PCC values ($< 0.03$) but spread widely along the 2D metric axes, reflecting the large visual differences caused by added objects in views that see them. The Pearson correlations weaken or invert: PCC vs.\ PSNR becomes $r = +0.64$ (positive, opposite to jitter), PCC vs.\ SSIM becomes $r = +0.48$, and PCC vs.\ LPIPS becomes $r = -0.45$ (negative, opposite to jitter). Views with high LPIPS (large visual change due to added objects) show \emph{low} PCC (the scene is geometrically consistent). This decorrelation demonstrates that PCC and 2D metrics measure fundamentally orthogonal properties: 2D metrics capture \emph{whether images look different}, while PCC captures \emph{whether the underlying 3D geometry is consistent across views}. For details, please refer to \Cref{tab:pcc_summary} in \Cref{app:occlusion_test}. These experiments establish that PCC captures multi-view 3D consistency—a property that 2D metrics cannot measure. A novel-view synthesis model that hallucinates plausible but geometrically incoherent content across views may achieve high PSNR and SSIM per-view while exhibiting elevated PCC. Conversely, a model that produces a 3D-consistent scene differing from the ground truth in object composition will show degraded 2D metrics but low PCC. We therefore advocate reporting PCC alongside standard 2D perceptual metrics to provide a more complete evaluation of novel-view synthesis quality.\looseness-1

%% file: main/PCC/pcc_vs_camera_pose_metrics.tex
Evaluating the 3D consistency of generated views using PCC relies heavily on accurate depth estimation, which can be natively predicted by the generation model or extracted via an external 3D reconstruction tool. In practice, these estimations are rarely flawless. Prone to scale drift, depth boundary errors, and pose inaccuracies, such underlying artifacts inevitably compromise the reliability of the PCC measurement. Given that camera-pose accuracy metrics similarly depend on the geometric estimations of SfM or 3D reconstruction models, this section evaluates how depth inaccuracies impact PCC and compares its robustness against standard camera-pose metrics.

To extract scene geometry, we process the source image $I_\mathrm{s}$ and the generated target image $\hat{I}_\mathrm{t}$ through a multi-view geometry estimator $\mathcal{E}$, specifically utilizing Depth-Anything3-LARGE-1.1~\citep{lin2025da3}. This estimator jointly predicts the metric depth and camera parameters for both views:$$\hat{\mathbf{D}}_s,\; \hat{\mathbf{D}}_t,\; \hat{\mathbf{T}}_s,\; \hat{\mathbf{T}}_t,\; \hat{\mathbf{K}}_s,\; \hat{\mathbf{K}}_t \;=\; \mathcal{E}\!\bigl(\mathbf{I}_s,\;\hat{\mathbf{I}}_t\bigr), \label{eq:da3}$$where $\hat{\mathbf{D}}\!\in\!\mathbb{R}^{H\times W}$ are the scale-aligned depth maps (details in ~\Cref{app:scale_alignment_pcc}), $\hat{\mathbf{T}}\!\in\!\mathrm{SE}(3)$ are the camera-to-world extrinsics, and $\hat{\mathbf{K}}\!\in\!\mathbb{R}^{3\times 3}$ represent the intrinsic matrices. We use the estimated depth maps for computing PCC and the estimated camera poses for computing relative pose errors (RPEs; see ~\Cref{app:pose_auc}).

We follow the experimental protocol \Cref{sec:blender_experiments} and evaluate the PCC scores against camera-pose accuracies for the scale jitter variants. While the experiments in \Cref{sec:blender_experiments} utilize ground-truth depth generated by Blender, this section adopts a more practical setup using the DA3-estimated geometry of the synthetic scenes to compare PCC against standard camera-pose metrics. 
\begin{figure}[t]
  \centering
  \includegraphics[width=\textwidth]{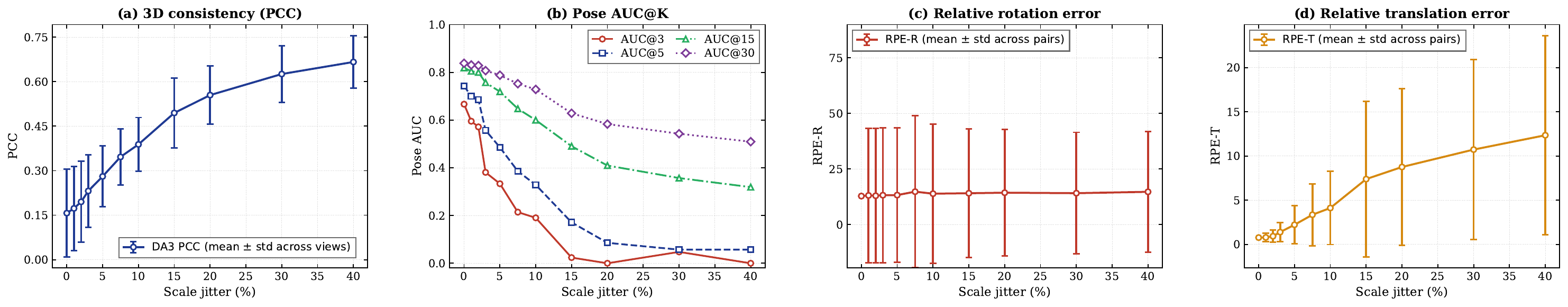}
  \caption{\textbf{Sensitivity of the DA3 PCC metric to controlled scale perturbations.} Starting from a clean reference scene, we re-render each object with a random per-object scale jitter of 1–40\% and measure (a) PCC, (b) pose AUC@{3, 5, 15, 30}, (c) relative rotation error (RPE-R), and (d) relative translation error (RPE-T), with depth and cameras predicted by DA3 and FG/BG-split affine alignment (no top-\% pruning). With one exception, all evaluated metrics track geometric error monotonically alongside increasing jitter magnitude. Specifically, PCC rises from 0.07 at 0\% jitter to 0.62 at 40\%, AUC@30 falls from 0.84 to 0.51 (with AUC@3 collapsing from 0.67 to 0 by the 20\% mark), and RPE-T increases from 0.77 to 12.35. RPE-R is the only metric that fails to reliably reflect the introduced noise, remaining largely stagnant and breaking monotonicity at 2\%, 7\%, and 30\%.}
  \label{fig:pcc_da3_jitter}
\end{figure}
% \textbf{Experiment 3: Scale Jitter 2.0.} 
% \ds{I was expecting this section to compare these two quantities: PCC(ground-truth depth and camera geometry), and PCC(estimated depth and camera geometry from DA3) as a function of jitter. Maybe add the former to the plots? This gives an idea of how the DA3 estimation influences PCC relative to gt.} 
% In this experiment, we re-evaluate the PCC scores and camera-pose accuracies for the scale jitter variants. However, rather than relying on ground-truth depth from the simulator, we compute these metrics using the scale-aligned depth estimates produced by DA3.
As shown in \Cref{fig:pcc_da3_jitter}, PCC, pose-AUC, and relative translation error (RPE-T) all worsen monotonically with increasing jitter. On the other hand, RPE-R remains essentially flat -- an expected outcome since the actual camera poses remained static while only the object scales changed. Importantly, RPE-T alone is insufficient to capture overall camera-pose accuracy; a complete assessment requires evaluating both translation and rotation. Consequently, the RPE metric suite as a whole fails to serve as a reliable 3D consistency metric under the jitter test. Furthermore, although pose-AUC appears to pass the test by degrading monotonically, it achieves this by aggregating pairwise RPEs into a single score, which masks the underlying failure of the rotation estimates. However, in general practice, any flaw in either RPE-R or RPE-T might affect the overall reliability of the AUC score since the metric relies on both components.

\subsubsection{Evaluation on a Real-World Scene}

\begin{figure}[t]
  \centering
  \includegraphics[width=0.9\linewidth]{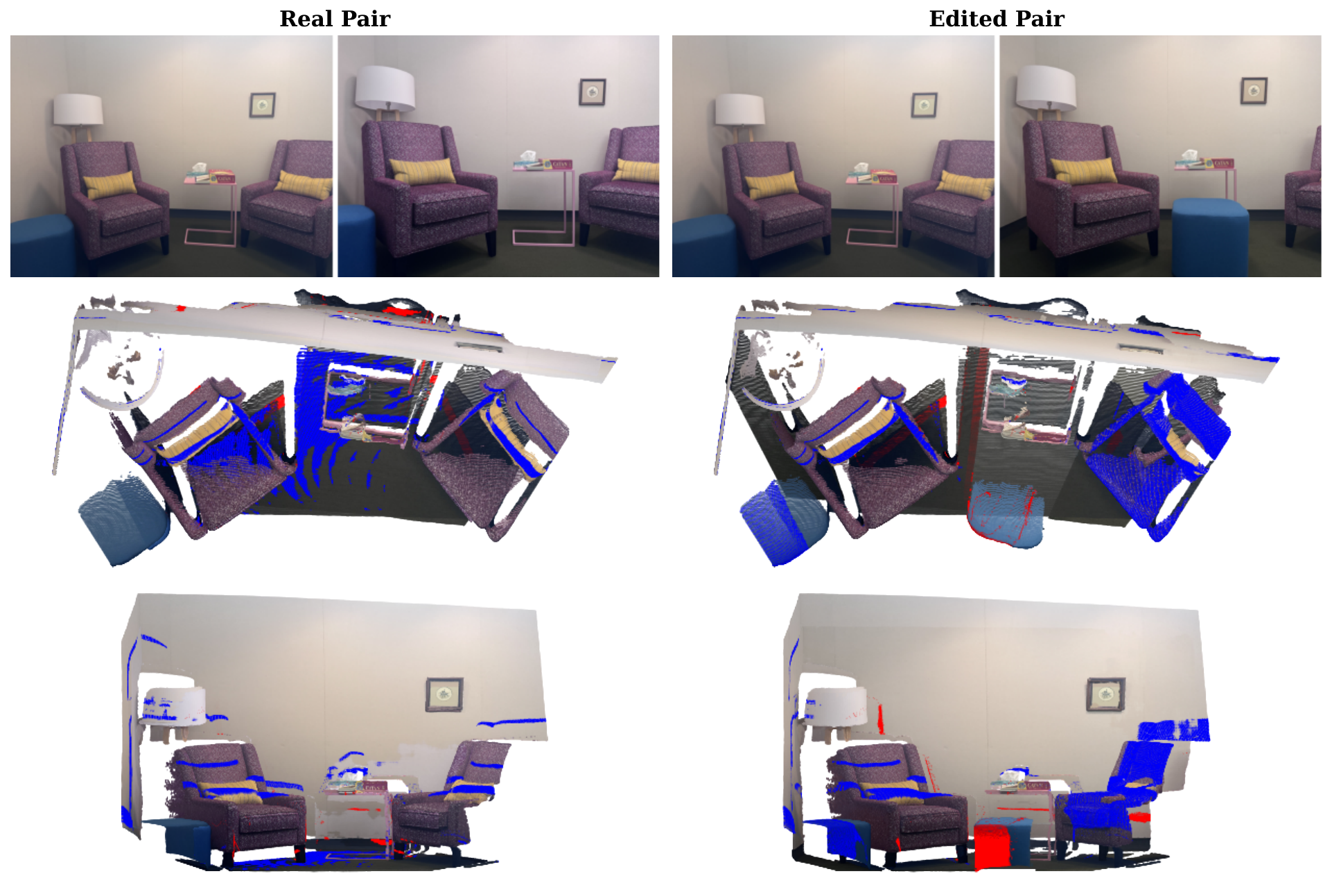}
  \caption{\textbf{Geometrically inconsistent point detections in a counterfactual scene.} The \textit{real pair} depicts a 3D-consistent scene in which all objects remain stationary as the camera moves between the two viewpoints. The \textit{edited pair} shares the same source image (same left image), and its target image (right image) is rendered from the same camera viewpoint as in the consistent pair. However, one or more objects are displaced in the second's target, producing a counterfactual scene that is incompatible with the 3D structure observed in source. By projecting the point cloud derived from each image into the other view, the PCC-based procedure identifies points whose projected geometry is inconsistent with the modified scene. These inconsistencies are highlighted as red and blue pixels, corresponding to inconsistencies detected by the forward and backward projections, respectively.}
  \label{fig:counterfactual_example}
\end{figure}

We conduct an experiment in a real-world scene to evaluate whether DA3 depth estimates can be used to detect 3D inconsistencies under a counterfactual setting. We define a \textit{Real Pair}, composed of a source and a target view, depicting a 3D-consistent scene in which all objects remain stationary while the camera moves between two viewpoints. We also define an \textit{Edited Pair} sharing the same source image as the Real Pair and an edited target image, retaining the same camera viewpoint as the one of the Real Pair target. The edited target displaces objects in the scene, creating a counterfactual scene whose object configuration is inconsistent with the 3D structure observed in the source image. An example is shown in~\Cref{fig:counterfactual_example}. 
%As shown in ~\Cref{fig:counterfactual_example}, the \textit{Real Pair} depicts a 3D-consistent scene in which all objects remain stationary while the camera moves between two viewpoints. The \textit{Edited Pair} shares the same source image—the left image—and its target image—the right image—retains the same camera viewpoint as the target image in the \textit{Real Pair}. However, the blue stool and red chair are displaced in the edited target image, creating a counterfactual scene whose object configuration is inconsistent with the 3D structure observed in the source image. 
As shown in the figure, applying our PCC evaluation using these DA3 depth estimates demonstrates that the metric can effectively isolate regions of structural inconsistency. Specifically, the forward projection from the source to the target view detects the absence of objects at their original spatial coordinates—such as the initial position of the stool, as well as the chair which was moved almost entirely out of the frame—flagging these vacated regions with projection errors (visualized as blue pixels). Conversely, the backward projection from the target back to the source identifies the conflicting geometries of the displaced objects at their newly introduced locations, such as the moved stool (visualized as red pixels). This bidirectional validation confirms that, even without ground-truth geometry, DA3-guided PCC can reliably pinpoint pixel-level hallucinations and physical displacements, effectively distinguishing geometrically coherent view synthesis from counterfactual 2D manipulations. It is worth noting, however, that off-the-shelf DA3 depth estimates are still subject to inherent noise. As observed in the Real Pair, minor misdetections emerge as sparse error pixels even when the underlying scene exhibits perfect multi-view consistency. % We hypothesize that fine-tuning foundation depth models directly using the PCC objective would mitigate these artifacts, simultaneously enhancing the accuracy of the depth priors and the precision of the PCC metric itself.

%% file: main/ablation/frame_gap_analysis.tex
\section{Sensitivity Analyses}
\subsection{How Do Frame Gap and Camera Displacement Affect Performance?}\label{sec:ablation_frame_gap}

\begin{figure}[t]
\centering
\includegraphics[width=0.75\linewidth]{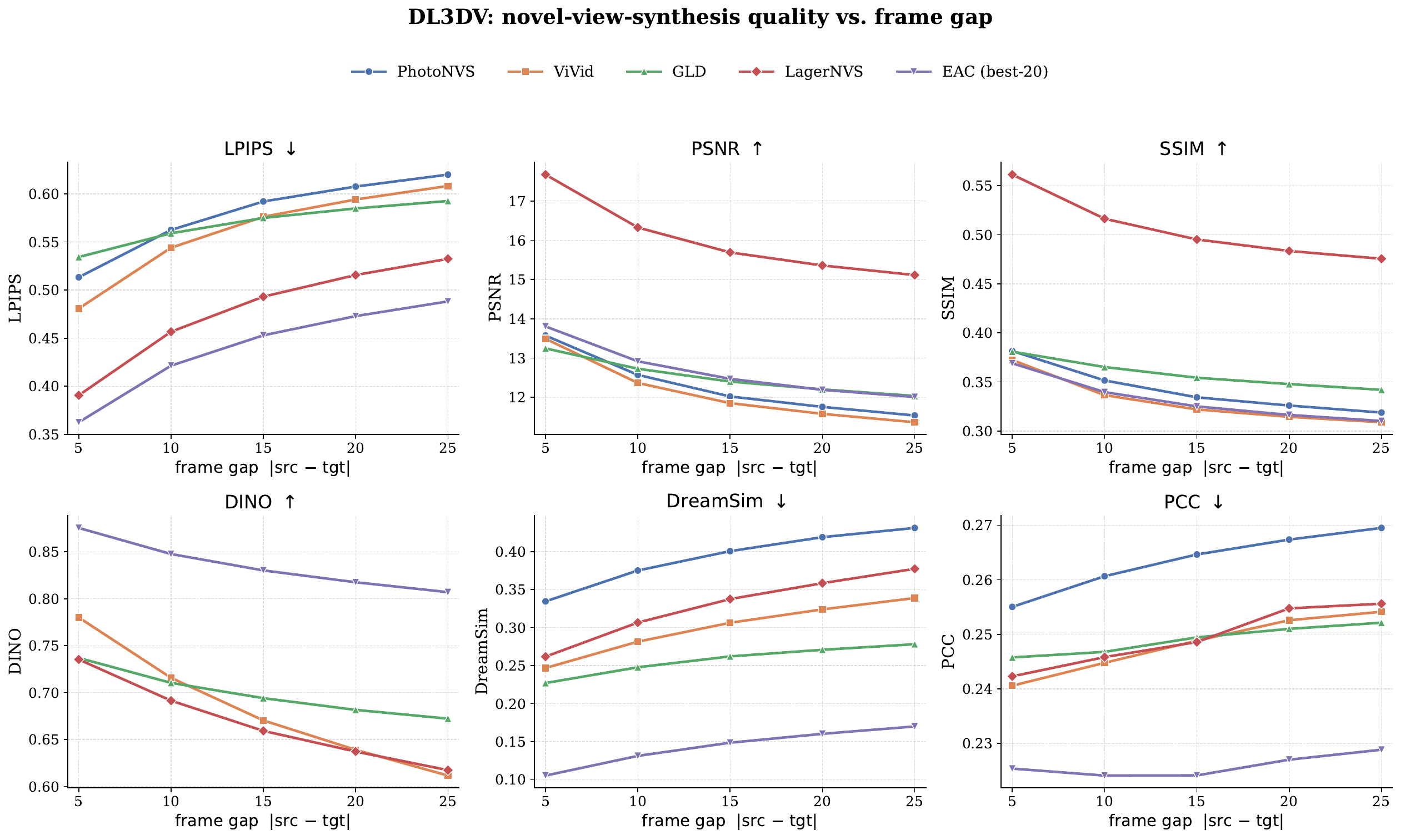}
\caption{\textbf{Effect of frame gap on model performance.} Performance consistently degrades across all models as the frame gap increases. Nevertheless, our method outperforms all baselines at every frame-gap setting on all metrics except PSNR and SSIM, for which LagerNVS achieves the best results.\looseness-1}\label{fig:dl3dv_frame_gap_ablation}
\end{figure}

Because the DL3DV dataset consists of continuous videos of static scenes, evaluating novel view synthesis requires sampling discrete source and target frame pairs from a sequence (see ~\Cref{app:datasets:eval} for exact sampling configurations). Assuming a relatively uniform camera velocity, the frame gap between the selected views serves as a proxy for the magnitude of the camera pose baseline. A larger frame gap translates to a more pronounced geometric shift, significantly increasing the difficulty of the synthesis task. For this ablation, we evaluate model resilience to baseline changes by varying the frame gap from 5 to 25 frames using a stride of 5.

As plotted in ~\Cref{fig:dl3dv_frame_gap_ablation}, we evaluate the impact of this frame gap across several core metrics: PSNR, SSIM, LPIPS, DINO-Sim, DreamSim and PCC. We observe a universal degradation in performance across all models as the frame gap widens. Notably, this performance drop is non-linear; the decline is steepest at narrower gaps (e.g., moving from 5 to 10 frames) and gradually smooths out as the baseline expands further. This trajectory and rate of degradation remain remarkably consistent across all evaluated architectures and metrics, indicating a uniform response to increasing geometric difficulty. Our model outperforms all baselines across every frame gap setting on all metrics except PSNR and SSIM, where LagerNVS scores highest. Rather than reflecting superior geometric synthesis, this discrepancy stems from two key factors: (1) \emph{Generative Blurring:} As shown qualitatively in \Cref{fig:qualitative_nvs}, \Cref{fig:qualitative_nvs_dl3dv_tnt}, and \Cref{fig:qualitative_nvs_re10k}, LagerNVS systematically produces smoother, blurred outputs. Because PSNR and SSIM heavily penalize high-frequency artifacts, they favor smooth regressions over sharp but slightly misaligned generations. (2) \emph{Rigid Pixel Comparisons:} In novel view synthesis, models must hallucinate occluded regions, which have infinite valid configurations. Because PSNR and SSIM require exact pixel-to-pixel alignment, they severely penalize high-quality, plausible generations if they deviate from the specific ground-truth pixels.\looseness-1

% See Appendix X for an exhaustive breakdown of these metric discrepancies and expanded qualitative analyses.

%% file: main/ablation/eac_shadow.tex
\subsection{Escaping the 2D Cave: Does Depth Conditioning Truly Benefit Novel View Synthesis?}\label{sec:ablation_eac_shadow}

\begin{table*}[t!]
\centering
\footnotesize

\begin{tabular}{l ccc c}
\toprule
& \multicolumn{3}{c}{\textbf{Perceptual Quality}} & \textbf{Geometry} \\
\cmidrule(r){2-4} \cmidrule(l){5-5}
\textbf{Model} & LPIPS $\downarrow$ & DINO-Sim $\uparrow$ & DreamSim $\downarrow$ & PCC $\downarrow$ \\
\midrule

\multicolumn{5}{l}{\textbf{DL3DV Dataset}} \\
\midrule
\modelname-Shadow & \third{$0.4952 \pm 0.1386$} & \third{$0.8035 \pm 0.1540$} & \third{$0.1676 \pm 0.1304$} & \third{
$0.2533 \pm 0.1640$} \\
\modelname-Base  & \second{$0.4448 \pm 0.1569$} & \second{$0.8366 \pm 0.1389$} & \second{$0.1420 \pm 0.1252$} & \second{$0.2448 \pm 0.1646$} \\
\modelname-PCC-BoN  & \first{$0.4340 \pm 0.1556$} & \first{$0.8386 \pm 0.1388$} & \first{$0.1402 \pm 0.1232$} & \first{$0.2257 \pm 0.1556$} \\
% \modelsep
% Oracle & $\cdot$ & $\cdot$ & $\cdot$ & $0.1199 \pm 0.1539$ \\

\midrule
\multicolumn{5}{l}{\textbf{RE10K Dataset}} \\
\midrule
\modelname-Shadow & \third{$0.3166 \pm 0.1328$} & \third{$0.8982 \pm 0.0766$} & \third{$0.0780 \pm 0.0568$} & \third{$0.0870 \pm 0.1433$} \\
\modelname-Base & \second{$0.2613 \pm 0.1249$} & \second{$0.9157 \pm 0.0652$} & \second{$0.0669 \pm 0.0497$} & \second{$0.0773 \pm 0.1430$} \\
\modelname-PCC-BoN & \first{$0.2150 \pm 0.1220$} & \first{$0.9222 \pm 0.0667$} & \first{$0.0655 \pm 0.0510$} & \first{$0.0656 \pm 0.1359$} \\
% \modelsep
% Oracle & $\cdot$ & $\cdot$ & $\cdot$ & $0.0539 \pm 0.1380$ \\
\midrule
\multicolumn{5}{l}{\textbf{Tanks \& Temples Dataset}} \\
\midrule
\modelname-Shadow & \third{$0.6336 \pm 0.0859$} & \third{$0.6549 \pm 0.2078$} & \third{$0.2928 \pm 0.1512$} & \third{$0.4291 \pm 0.4094$} \\
\modelname-Base  & \second{$0.5835 \pm 0.1016$} & \second{$0.7242 \pm 0.1872$} & \second{$0.2362 \pm 0.1371$} & \second{$0.3826 \pm 0.3875$} \\
\modelname-PCC-BoN & \first{$0.5596 \pm 0.1164$} & \first{$0.7339 \pm 0.1854$} & \first{$0.2232 \pm 0.1343$} & \first{$0.2849 \pm 0.3139$} \\
% \modelsep
% Oracle & $\cdot$ & $\cdot$ & $\cdot$ & $0.1899 \pm 0.2478$ \\
\bottomrule
\end{tabular}
\caption{\textbf{Ablation study of the ESC framework.} Quantitative evaluation comparing \modelname-shadow, \modelname-base, and the full \modelname-PCC-BoN pipeline across the DL3DV, RE10K, and Tanks \& Temples datasets.}

\label{tab:esc_shadow}
\end{table*}
We first evaluate the necessity of depth tracking by contrasting \modelname-shadow against our depth-aware variants. \modelname-shadow retains the exact same model architecture and adapter footprint (sharing identical LoRA and ControlNet structures) as the other variants, but lacks the input and output channels for depth tokens.\footnote{We denote the depth-free baseline as \textbf{\modelname-shadow}, as it is forced to infer geometry solely from 2D pixel projections (the ``shadows cast on the wall'').} 

As demonstrated by the quantitative ablation results in \Cref{tab:esc_shadow}, incorporating explicit depth estimation yields substantial performance gains. Specifically, our depth-aware models (\modelname-base and \modelname-PCC-BoN) consistently outperform \modelname-shadow across both perceptual metrics (LPIPS, DINO-Sim, DreamSim) and geometric consistency (PCC). 

\begin{wrapfigure}{r}{0.4\textwidth}
  \centering
  \includegraphics[width=\linewidth]{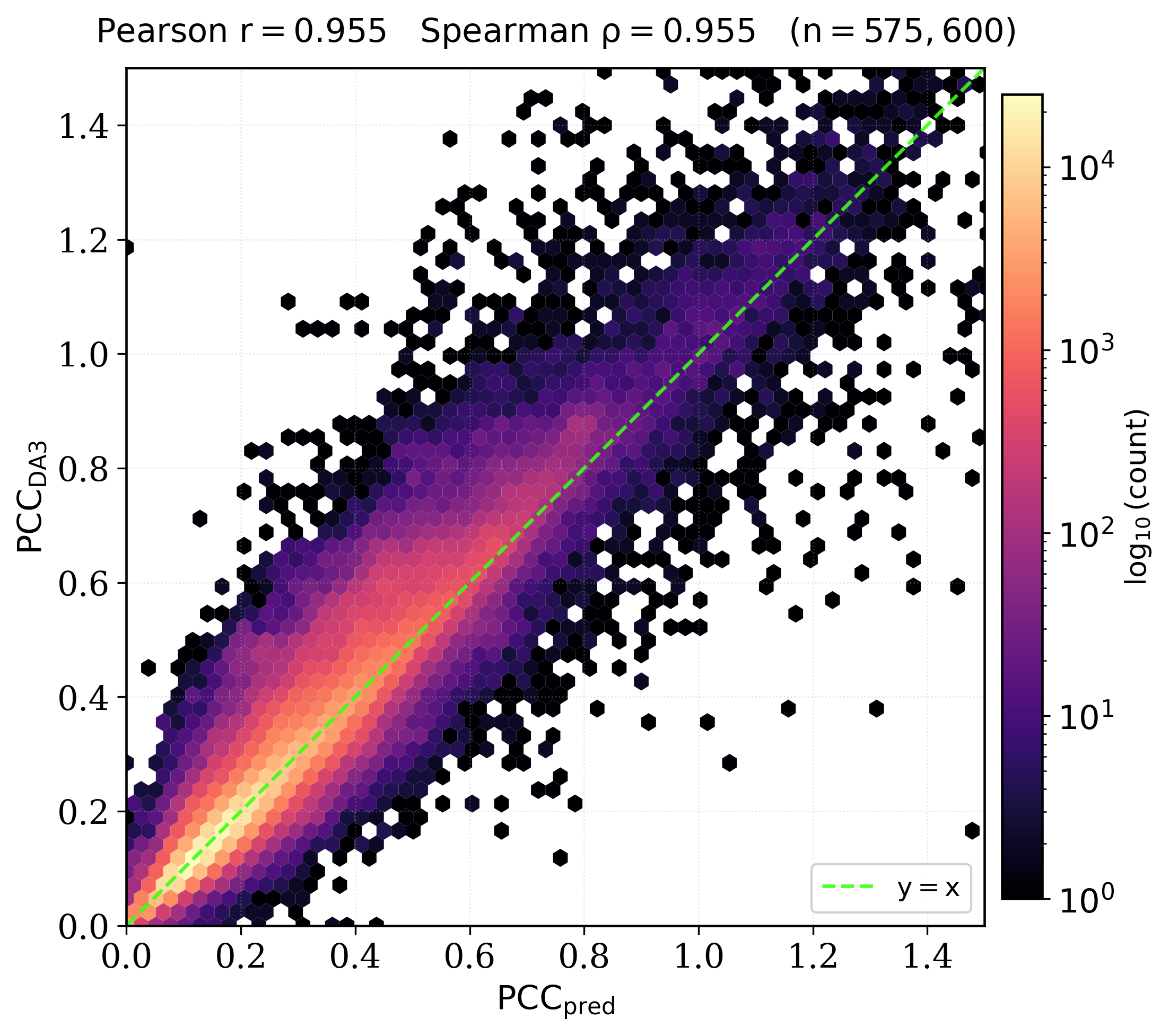}
  \caption{\textbf{Correlation between DA3-derived and predicted PCC metrics.} }
  \label{fig:pcc_da3_vs_pred}
\end{wrapfigure}

On the other hand, in our Best-of-N (ESC-PCC-BoN) inference strategy, the PCC metric serves as a geometry-guided scoring function to select the most spatially consistent sample from a candidate pool of generated views. Computing this PCC score inherently requires depth maps for both the anchor and the generated target views. Rather than relying on an external depth estimator to process the generated RGB images—which would introduce significant computational overhead during inference—we directly utilize the depth maps natively predicted by our ESC framework alongside the RGB generations. 

To validate the reliability of our internally predicted depth for driving this 3D consistency evaluation, we analyze its alignment with established, state-of-the-art geometry estimators. As illustrated in \Cref{fig:pcc_da3_vs_pred}, we compare the PCC scores calculated using our model's predicted depth against those computed using depth estimates extracted via Depth-Anything 3 (DA3). We observe a remarkably strong positive correlation (Pearson $r = 0.955$, Spearman $\rho = 0.955$) between the two metrics across the evaluated samples. This strong alignment confirms that our model's native depth predictions accurately capture the underlying multi-view geometry, proving them to be highly effective and sufficient for the Best-of-N selection process without the need for auxiliary depth estimation tools.
% \textbf{Enforcing Multi-View Consistency via PCC.} While \modelname-base introduces per-view depth conditioning and prediction, generating a globally 3D-consistent scene requires enforcing explicit geometric constraints across different viewpoints. To bridge this gap, we integrate our depth-reprojection consistency metric, PCC, using two distinct paradigms:
% \begin{itemize}
%     \item 
%     \textbf{Test-Time Selection (\modelname-PCC-BoN):} We leverage PCC as a post-hoc, geometry-guided scoring function to select the most spatially consistent sample from a candidate pool of $N$ generated views. 
%     \item 
%     \textbf{Training-Time Regularization (\modelname-FT):} We incorporate the PCC loss directly into the optimization pipeline as an explicit fine-tuning constraint, forcing the model to inherently generate view-consistent geometries. \ars{I think we do not have FT version here.}
% \end{itemize}
% Comparing \modelname-base to \modelname-PCC-BoN and \modelname-FT in \Cref{tab:blender_qualitative} and \Cref{tab:perceptual_geometry_benchmarks} highlights the critical role of cross-view consistency. While \modelname-base establishes a strong baseline, both PCC-driven variants significantly reduce geometric distortion and artifacts. Specifically, the fine-tuned variant (\modelname-FT) delivers the most robust and authentic novel view synthesis on simulated scenes, whereas the best-of-N strategy (\modelname-PCC-BoN) yields the superior performance on in-the-wild imagery.

%% file: appendix/main.tex
% ---------------------------------------------------------------------------
% Appendix-only local table of contents (uses `etoc`, no global TOC).
% Renders on its own page before any appendix section.
% ---------------------------------------------------------------------------
% {
%   \etocsetnexttocdepth{subsection}
%   \etocsettocstyle
%     {\noindent\textbf{\Large Appendix Contents}\par\vspace{0.6em}\hrule\vspace{0.6em}}
%     {\vspace{0.6em}\hrule\vspace{0.5em}}
%   \localtableofcontents
% }
% \clearpage
\crefalias{section}{appendix} % <-- Tells cleveref to treat sections as appendices from this point on
\crefalias{subsection}{appendix} % (Optional: if you also want appendix subsections labeled as Appendix A.1)

% Start a completely separate contents tracking system
\startcontents[appendices]
% Print the local contents (the '1' means it will list down to sections)
\printcontents[appendices]{}{1}{}

\newpage

\section{Datasets}\label{app:sec:datasets}
\input{appendix/datasets}
\section{Training Configurations}
\input{appendix/training_config}

\clearpage
\section{PCC: Supplementary Document}
\input{appendix/PCC_materials/main}

\section{Baseline Models}
\input{appendix/baseline_models}

\section{Additional Results}
\input{appendix/results}

% \section{Why LagerNVS Leads on PSNR/SSIM: A Diagnostic Analysis}
% \input{appendix/analysis}

\input{main/ablation/simulator_results}

%% file: appendix/datasets.tex
% ==========================================================================
%  Datasets appendix
%  - Overview of all datasets used
%  - Procedural Blender + Kubric synthetic dataset
%  - Mixed real-world training datasets (DL3DV + RE10K) and pair sampling
%  - Evaluation datasets and pair selection
% ==========================================================================

\subsection{Overview}
\label{app:datasets:overview}

We train and evaluate \textsc{ESC} and baseline models on a mix of one synthetic and three real-world novel-view synthesis benchmarks.

\begin{itemize}
\item \textbf{Blender}~\citep{blender}\textbf{ + Kubric}~\citep{greff2021kubric} \textbf{(ours, synthetic).} A procedurally generated multi-view dataset of $\sim$2{,}000 training scenes and 100 validation scenes. Each scene contains a small object-centered scene of solid-coloured \texttt{Cube}/\texttt{Sphere} primitives spawned with Kubric and rendered with Blender from a fixed 15-view yaw$\times$pitch grid.
\item \textbf{DL3DV-10K}~\citep{ling2023dl3dv}. A large in-the-wild capture collection of $\sim$10K trajectories covering indoor and outdoor scenes, with COLMAP-derived intrinsics and extrinsics. We use 8,559 scenes of the training split for finetuning and the dedicated evaluation split for testing.
\item \textbf{RealEstate10K (RE10K)}~\citep{zhou2018re10k}. A real-world collection of $\sim$72K indoor real-estate walk-through clips with per-frame camera poses. We use a subset of the official train split for finetuning and the held-out test videos for evaluation.
\item \textbf{Tanks \& Temples (T\&T)}~\citep{knapitsch2017tnt}. An evaluation benchmark of high-resolution outdoor and indoor captures with COLMAP-aligned poses; we never train on T\&T and only use it to probe out-of-distribution generalization.
\end{itemize}

The synthetic Blender + Kubric dataset is used for controlled pre-training and ablation studies; the DL3DV/RE10K mix is used for the main finetuning runs; and all three real-world benchmarks (DL3DV, RE10K, T\&T) are used for quantitative evaluation.

% ==========================================================================
\subsection{The Blender + Kubric Synthetic Dataset}
\label{app:datasets:blender}

\paragraph{Engine stack.} Each scene is procedurally composed by Kubric~\citep{greff2021kubric} --- which spawns the primitive objects, assigns their materials, and orchestrates the multi-view camera grid and then handed off to Blender~\citep{blender} as the underlying renderer to produce RGB, depth and camera poses.

\paragraph{Assets.} We deliberately keep the scene content minimal. Each scene contains a small set of solid-coloured \textbf{Kubric primitives only}: \texttt{kb.Cube} and \texttt{kb.Sphere} instances spawned on a flat ground plane. The number of objects per scene is small and varies slightly: empirically every scene contains $5$ or $6$ primitives (mean $\bar N_{\text{obj}}{\approx}5.2$ across a random sample of $50$ scenes). Each primitive is assigned:
\begin{itemize}
\item a class drawn uniformly at random from \{\texttt{Cube}, \texttt{Sphere}\};
\item a scale $s \sim \mathcal{U}(0.8, 1.6)$;
\item a colour cycled through a fixed 8-colour palette (red, green, blue, yellow, pink, cyan, orange, purple);
\item a Principled BSDF material with metallic $\sim\mathcal{U}(0,1)$ and roughness $\sim\mathcal{U}(0.2,0.8)$.
\end{itemize}
By restricting the visible content to low frequencies and providing \emph{robust geometric cues}, this primitive representation is ideally suited for diffusion-based NVS training, where every pixel can be directly mapped to a well-defined analytic surface for depth supervision.

\paragraph{Placement.} Objects are arranged in a ring around the scene origin at azimuth $\theta_i = 2\pi i / N_{\text{obj}}$ with radius $r_i \sim \mathcal{U}(1.0, 2.5)\,\text{m}$, and sit directly on the ground plane at $z = s_i$ (so the object base touches $z = 0$).

\paragraph{Lighting.} Because the dataset uses no HDRI environment, all illumination comes from a single deterministic light rig: the CLEVR~\citep{johnson2017clevr} three-point setup, instantiated via \texttt{kb.assets.utils.get\_clevr\_lights(light\_jitter=0.5)}, plus a uniform ambient term $\mathbf{c}_{\text{amb}} = 0.05$ on every channel.

\paragraph{Camera grid.} Each scene renders $N_{\text{view}}{=}15$ views arranged on a fixed (yaw, pitch) grid centred on the scene origin and \emph{shared across the whole dataset}. The grid uses $5$ yaw offsets $\in\{-90^\circ, -45^\circ, 0^\circ, +45^\circ,+90^\circ\}$ crossed with $3$ pitch offsets $\in\{-20^\circ, 0^\circ, +20^\circ\}$ about a reference elevation of $30^\circ$, with focal length $35\,\text{mm}$ and a radius of $\sim$$12\,\text{m}$ from the scene origin.

\begin{figure}[t]
\centering
\setlength{\tabcolsep}{1.5pt}
\renewcommand{\arraystretch}{0.7}
\begin{tabular}{@{}c c c c c@{}}
\includegraphics[width=0.19\linewidth]{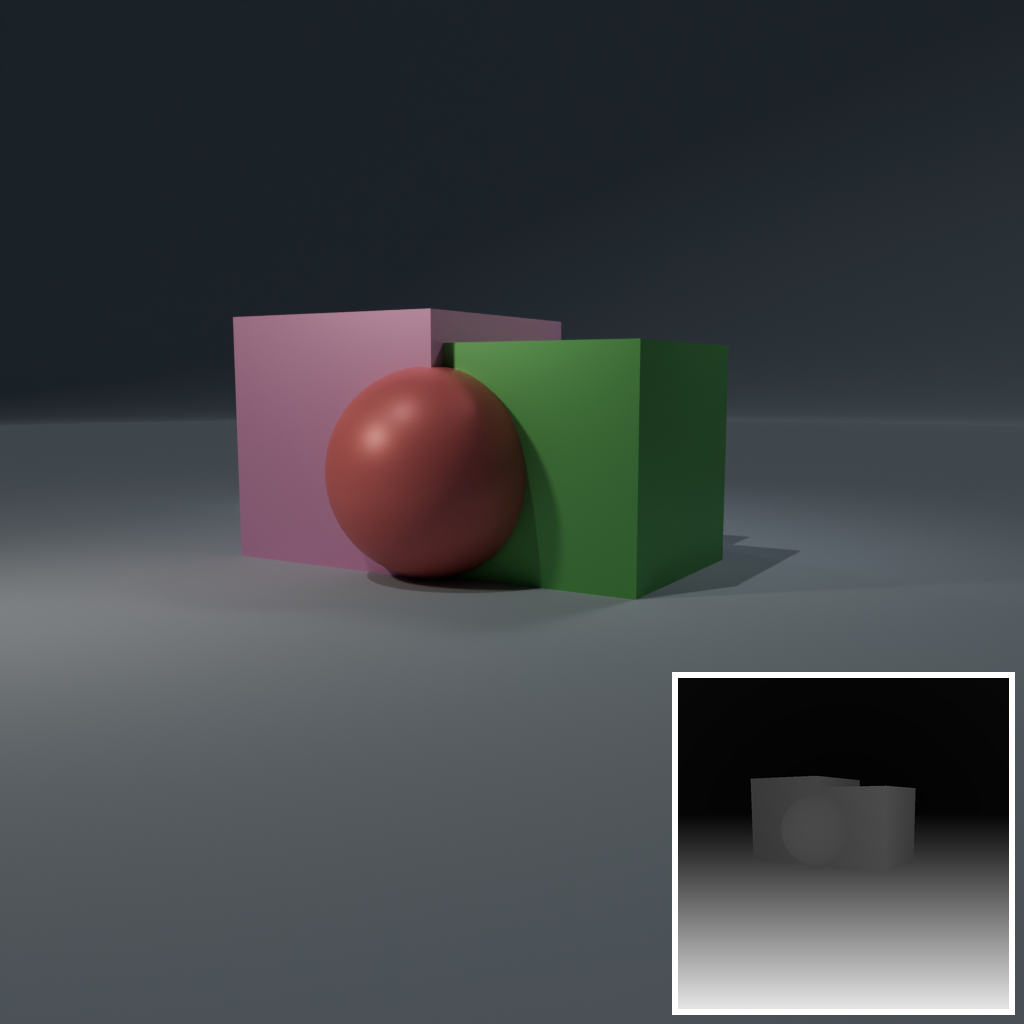} &
\includegraphics[width=0.19\linewidth]{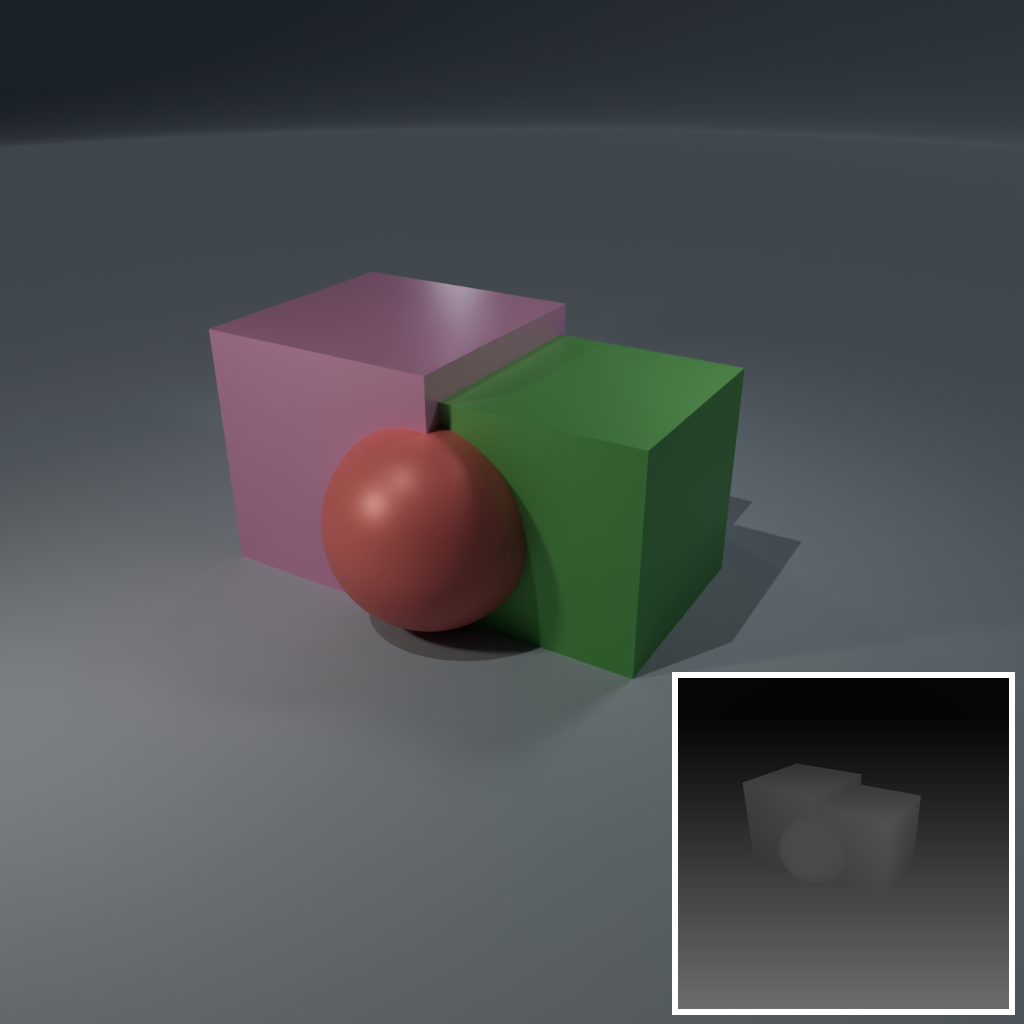} &
\includegraphics[width=0.19\linewidth]{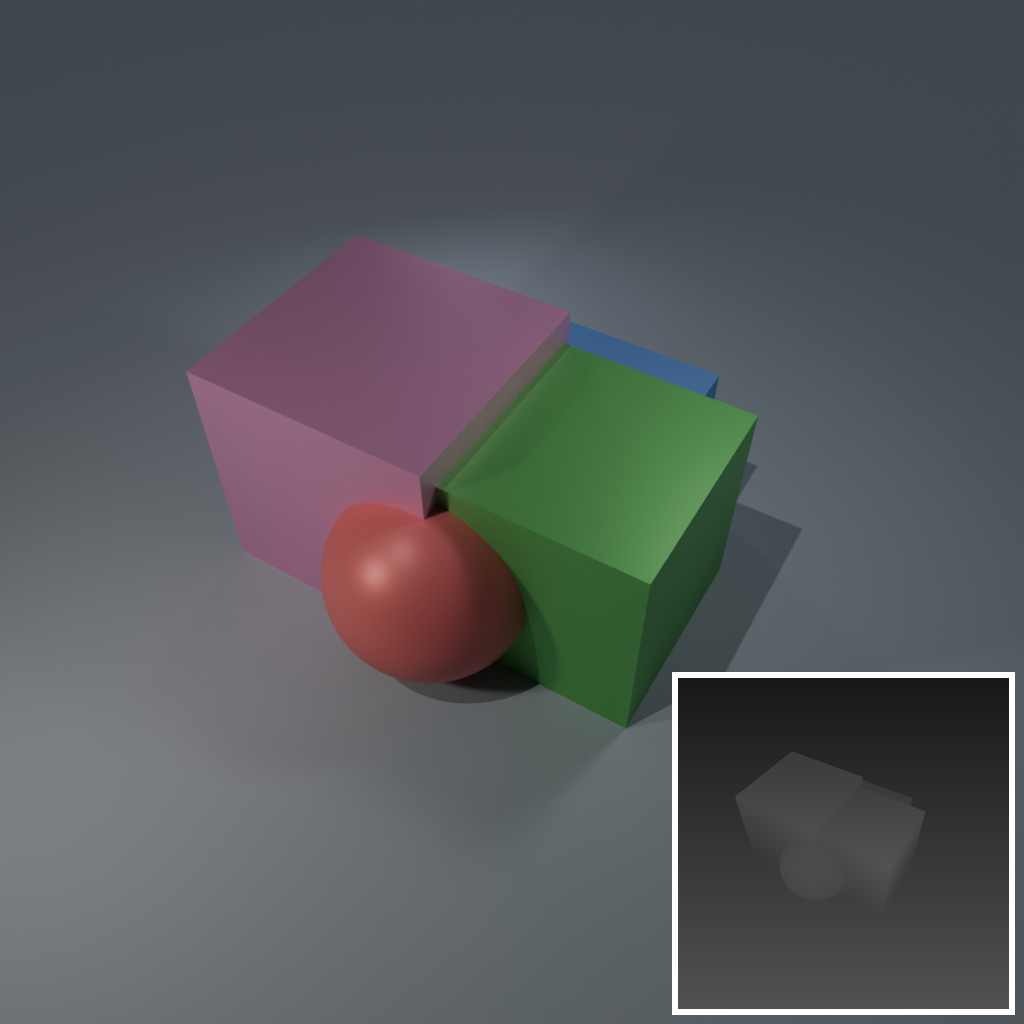} &
\includegraphics[width=0.19\linewidth]{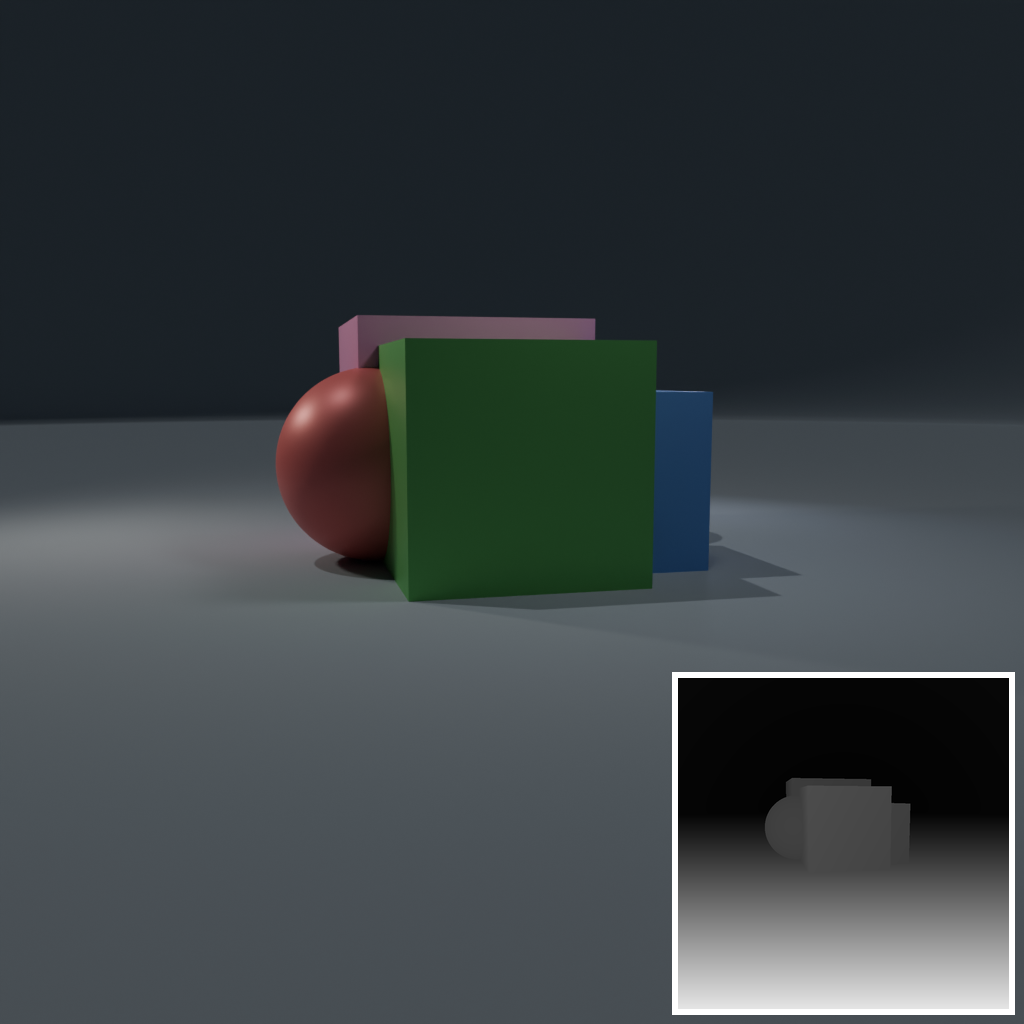} &
\includegraphics[width=0.19\linewidth]{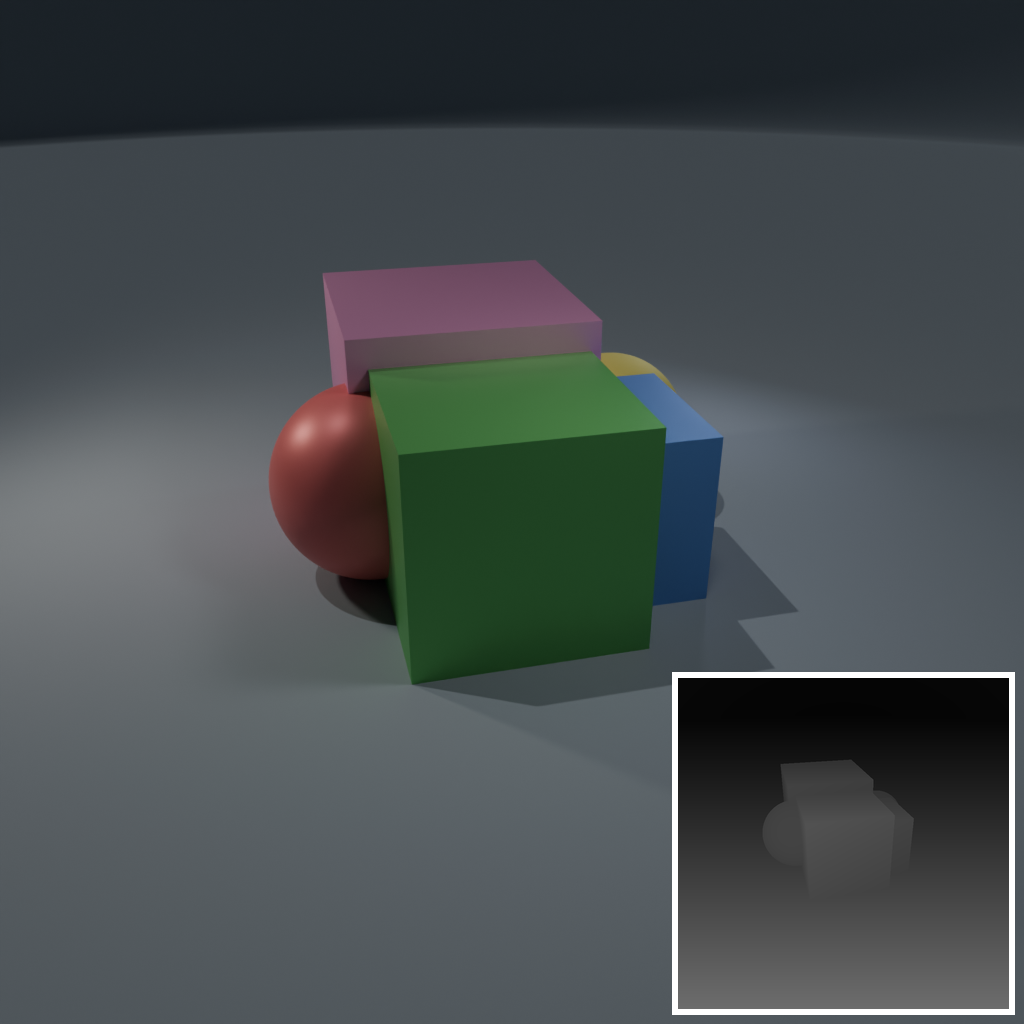} \\
\includegraphics[width=0.19\linewidth]{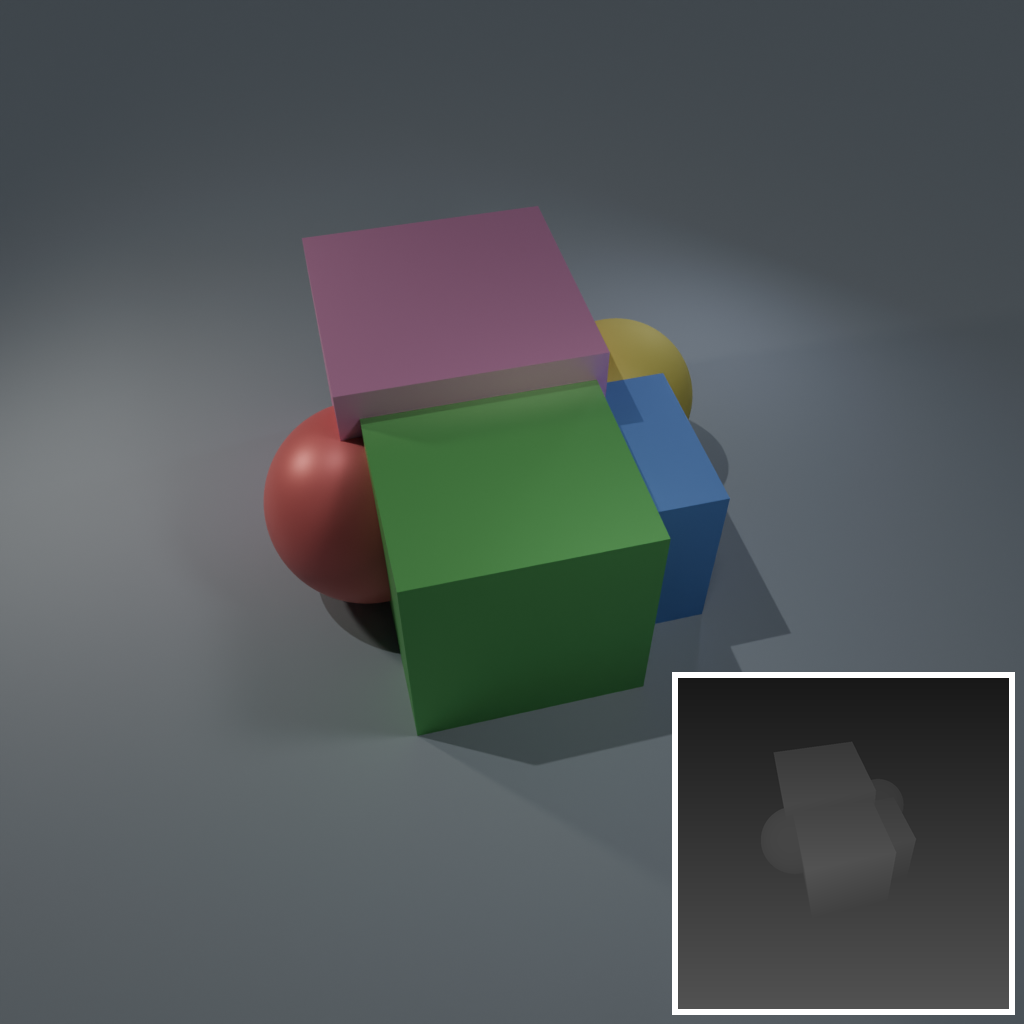} &
\includegraphics[width=0.19\linewidth]{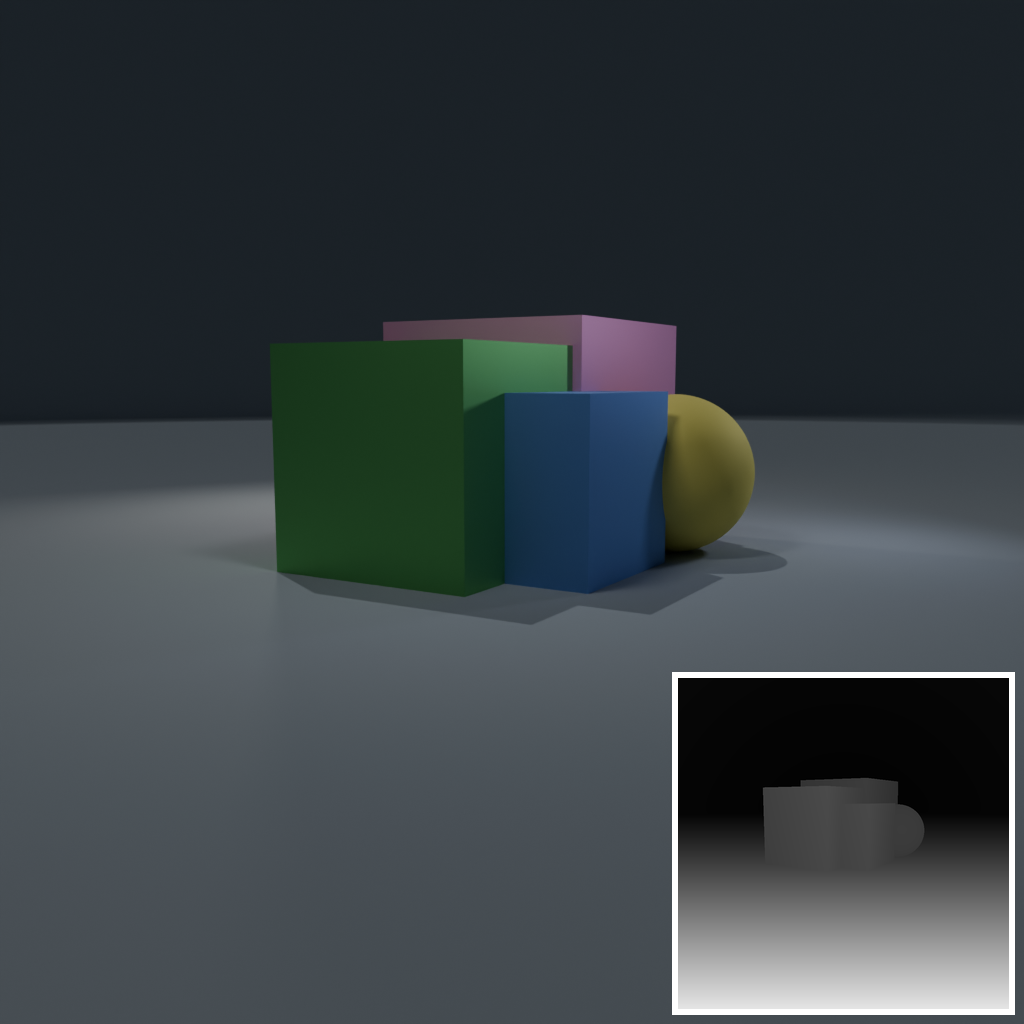} &
\includegraphics[width=0.19\linewidth]{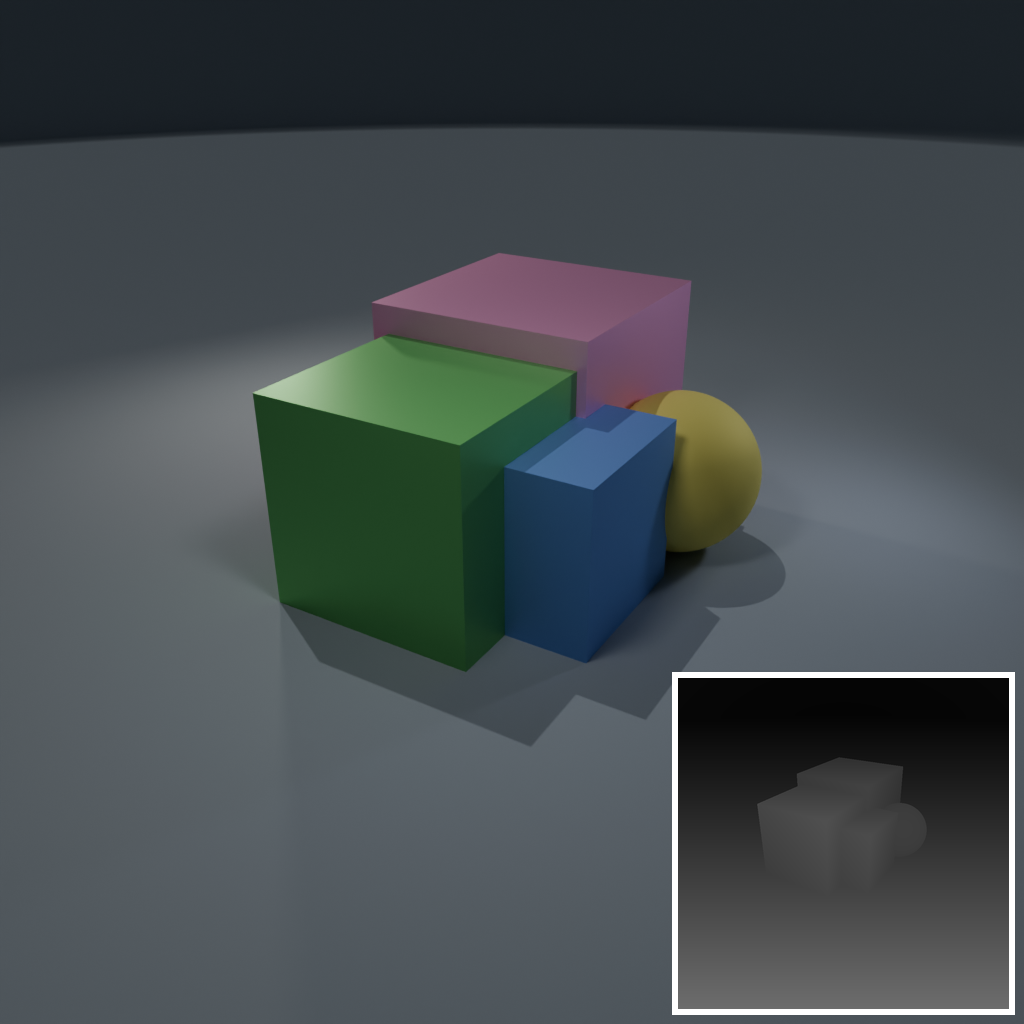} &
\includegraphics[width=0.19\linewidth]{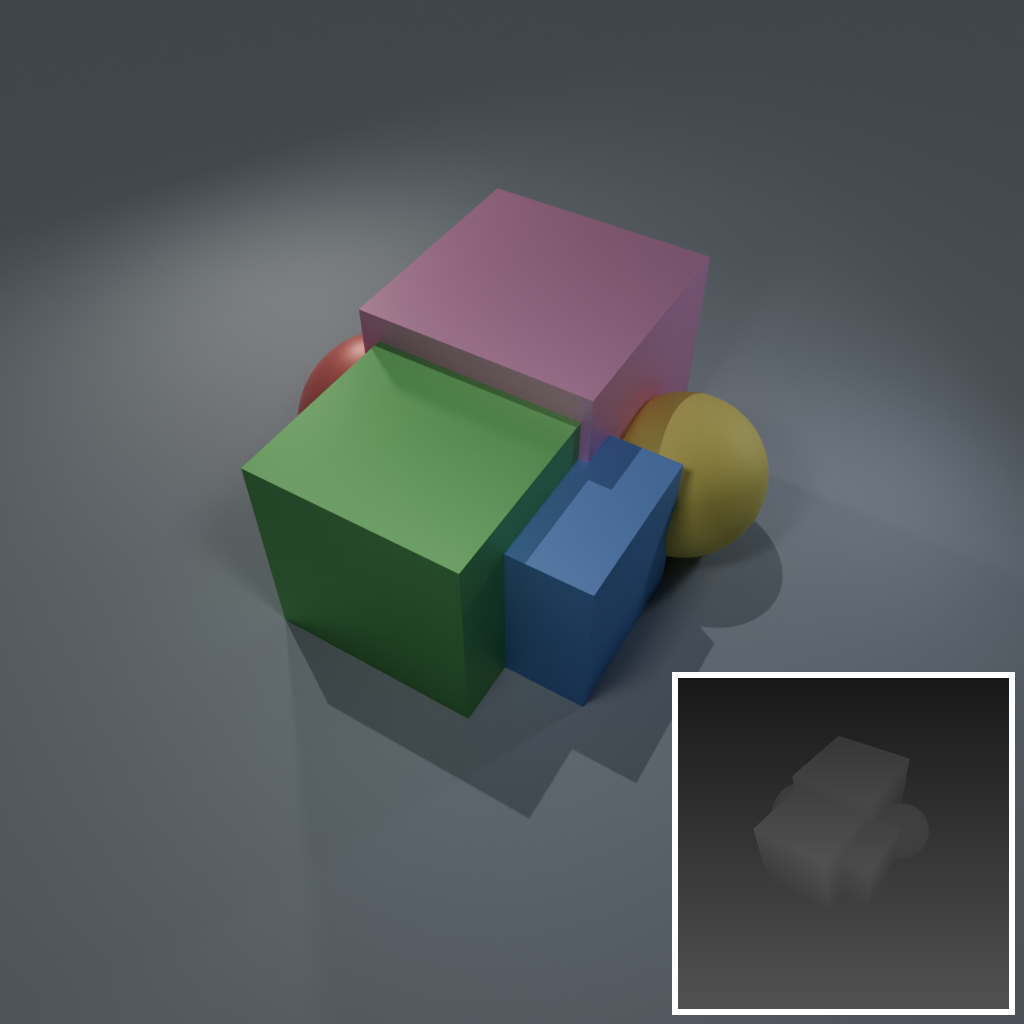} &
\includegraphics[width=0.19\linewidth]{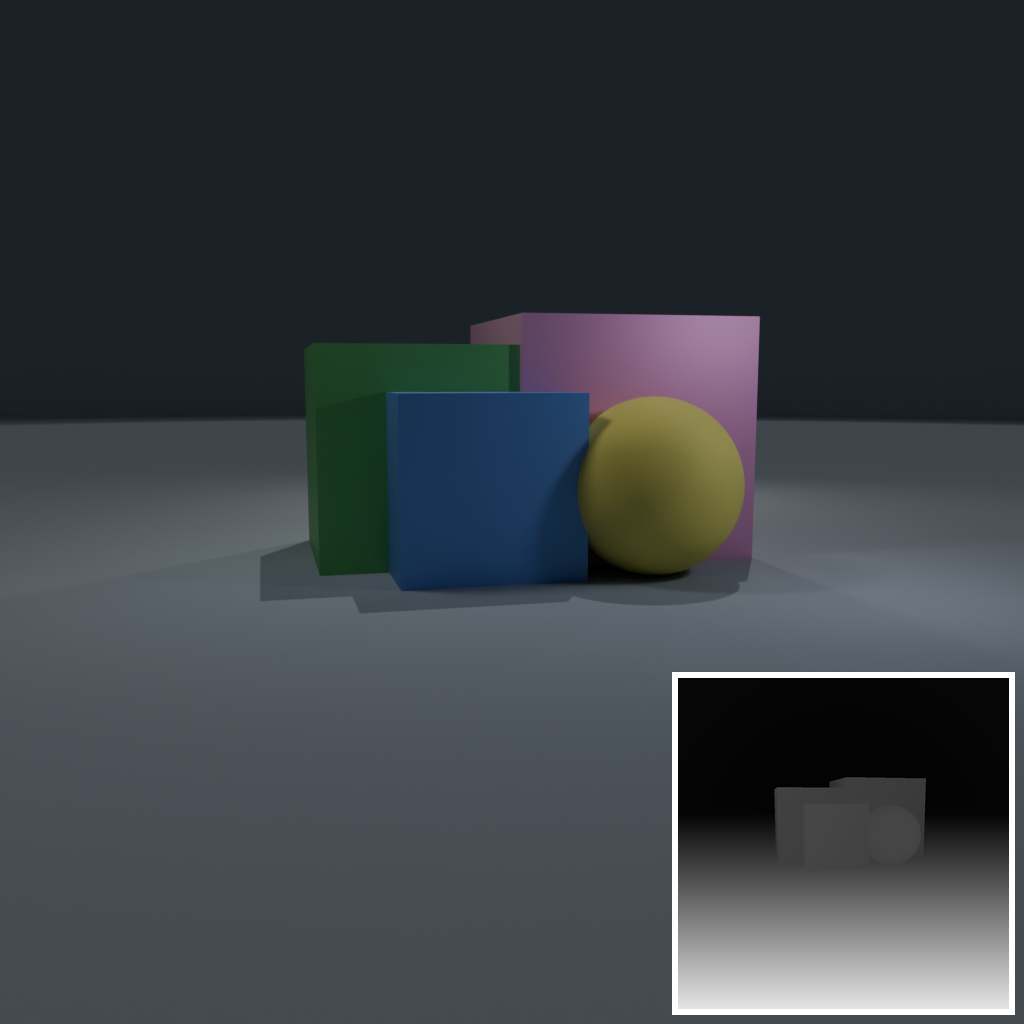} \\
\includegraphics[width=0.19\linewidth]{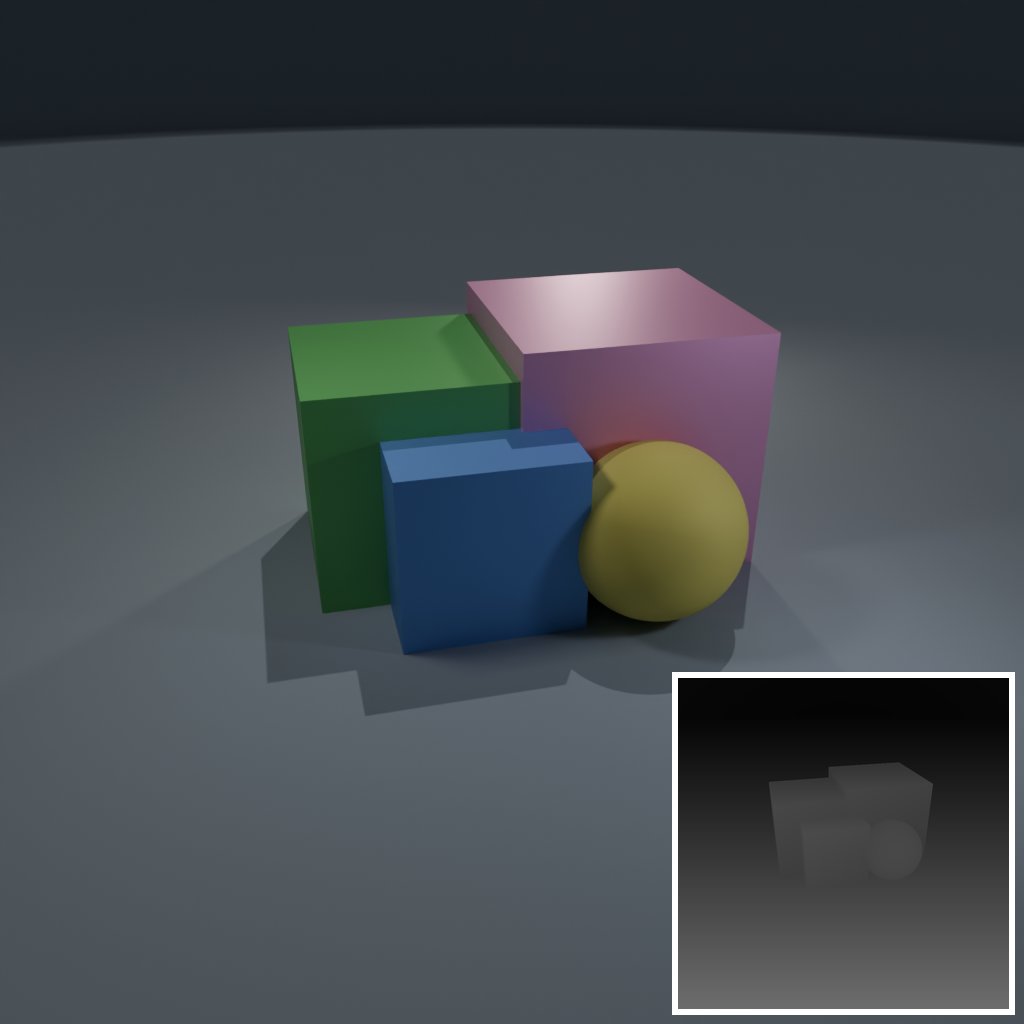} &
\includegraphics[width=0.19\linewidth]{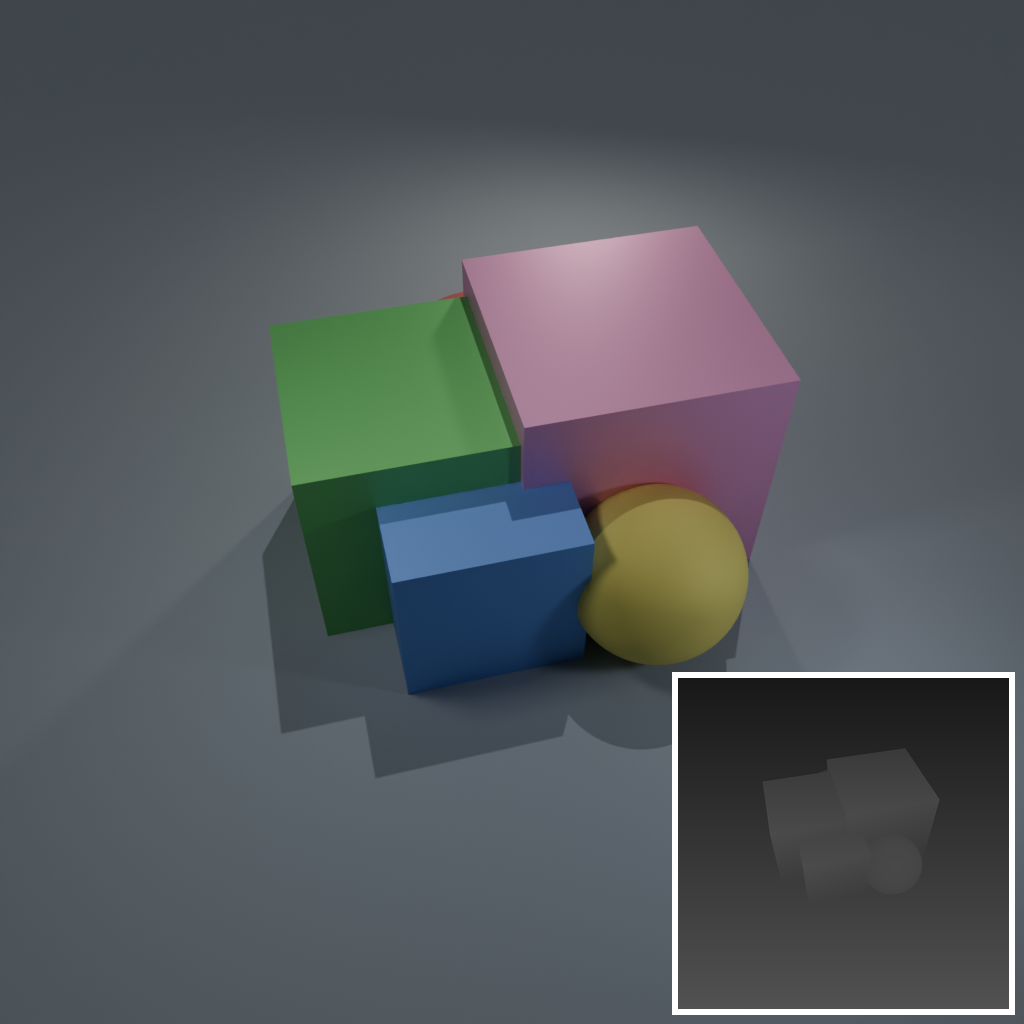} &
\includegraphics[width=0.19\linewidth]{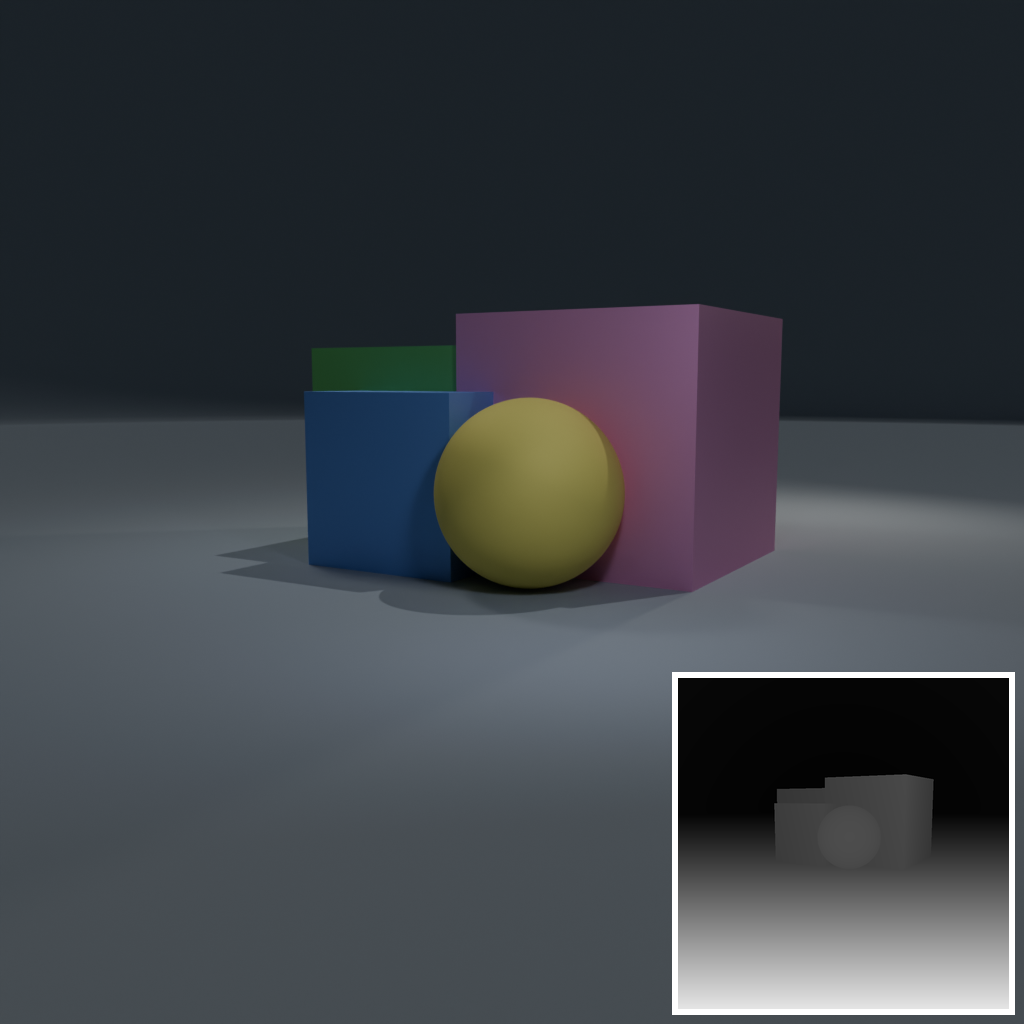} &
\includegraphics[width=0.19\linewidth]{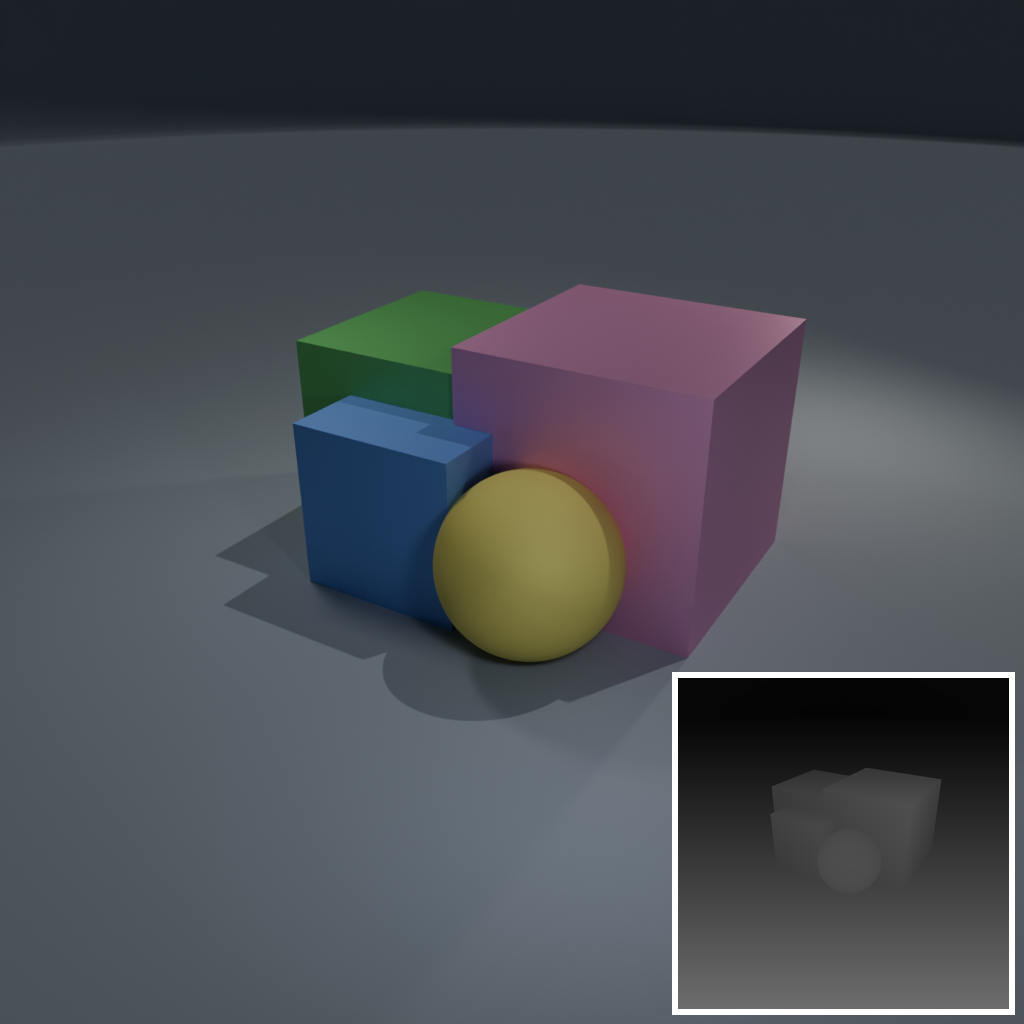} &
\includegraphics[width=0.19\linewidth]{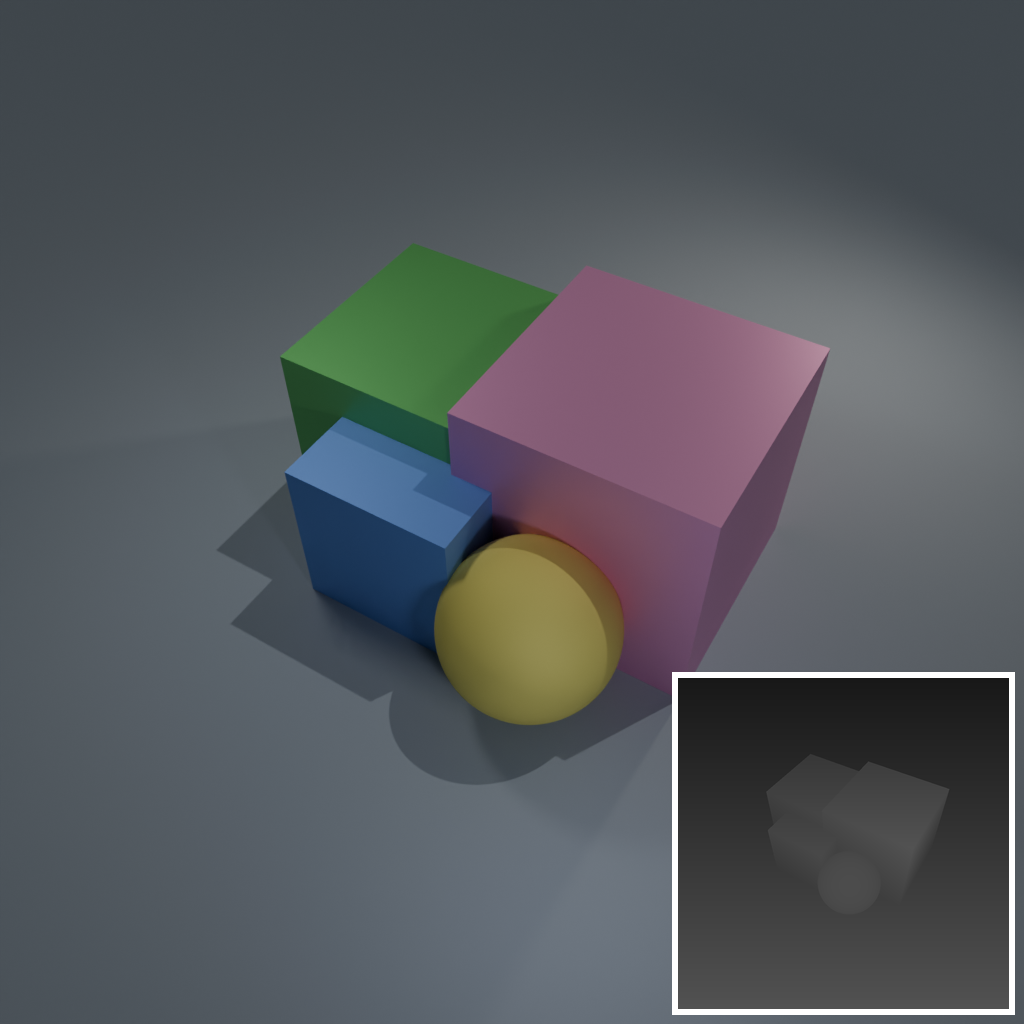} \\
\end{tabular}
\caption{\textbf{One Blender + Kubric scene rendered from all 15 grid viewpoints, with ground-truth disparity overlaid.} Each tile shows the rendered RGB view; the bottom-right is the corresponding ground-truth depth converted to inverse-depth (disparity). Rows / columns follow the shared yaw $\times$ pitch grid (yaws $\{-90^\circ, -45^\circ, 0^\circ, +45^\circ, +90^\circ\}$, pitches $\{-20^\circ, 0^\circ, +20^\circ\}$).}
\label{fig:blender_rgbd}
\end{figure}

\paragraph{Per-scene generation procedure.} We provide a visual demonstration of a single generated scene in \Cref{fig:blender_rgbd} and outline the generation process via pseudocode in \Cref{alg:blender_kubric_scene}.

\begin{algorithm}[t]
\caption{Procedural rendering of one Blender + Kubric scene.}
\label{alg:blender_kubric_scene}
\begin{algorithmic}[1]
\Require seed $s$, object count $N_{\text{obj}} \in \{5, 6\}$, view count
  $N_{\text{view}}{=}15$, resolution $1024{\times}1024$, shared view
  grid $\mathcal{V} = \{(\text{yaw}_i, \text{pitch}_i)\}_{i=1}^{N_{\text{view}}}$,
  colour palette $\mathcal{C}$
\State $\textsc{rng} \leftarrow \mathrm{seed}(s)$
\State scene $\leftarrow$ \textsc{kb.Scene}(resolution, frame\_end$=N_{\text{view}}$)
\State \textsc{AddGroundPlane}(scene)
\State \textsc{AddLights}(scene) \Comment{CLEVR rig (jitter $0.5$) + ambient $0.05$}
\For{$i \leftarrow 1$ to $N_{\text{obj}}$}
    \State cls $\leftarrow$ \textsc{rng}.choice(\{\texttt{Cube}, \texttt{Sphere}\})
    \State scale $\leftarrow$ \textsc{rng}.uniform($0.8$, $1.6$)
    \State colour $\leftarrow$ $\mathcal{C}[i \bmod |\mathcal{C}|]$
    \State $\theta \leftarrow 2\pi i / N_{\text{obj}}$;\ \
           $r \leftarrow$ \textsc{rng}.uniform($1.0$, $2.5$)
    \State pos $\leftarrow$ $(r\cos\theta,\, r\sin\theta,\, \text{scale})$
    \State $o \leftarrow$ cls(scale, pos, PrincipledBSDF(colour, $m,\,\rho$))
    \State scene $\mathrel{+}=$ $o$
\EndFor
\State \textsc{SetupCameraKeyframes}(scene, $\mathcal{V}$, ref\_elev$=30^\circ$, focal$=35$mm)
\State \textsc{SetupCompositorForDepth}(scene)
\For{$k \leftarrow 1$ to $N_{\text{view}}$}
    \State \textsc{Render}(scene, frame$=k$)
    \State \textsc{Save}(rgb$_k$, depth$_k$)
\EndFor
\end{algorithmic}
\end{algorithm}

% \paragraph{Data Documentation.} Every view is rendered at
% $1024{\times}1024$. For each scene we save the RGB as \emph{*.npy} and the depth as \emph{*.npy}. We save a shared \texttt{metadata.json} at the dataset root records the $15$ (yaw, pitch) offsets and per-view indices as well.

% \paragraph{Reproducing the data.} The full generation pipeline ships
% under \texttt{/storage/home/oooolga/Projects/TWM/scripts/}:
% \begin{itemize}
% \item \texttt{generate\_shared\_metadata.py} --- writes the shared
%   $15$-view grid to \texttt{metadata.json};
% \item \texttt{generate\_kubric\_scenes.py} --- orchestrates one or
%   many scenes, forwarding flags to the per-scene Blender script;
% \item \texttt{generate\_kubric\_scenes.slurm} --- the SLURM array
%   wrapper used to render all $2{,}001$ training scenes in parallel
%   (one task per scene).
% \end{itemize}
% Internally each task launches
% \verb|blender --background --python twm/kubric/generate_multiview_kubric.py|
% with the flags \verb|--no-hdri --use-grid-mode --no-gso| (forcing the
% primitives-only configuration described above) plus the per-task seed.

% ==========================================================================
\subsection{Mixed Real-World Training Dataset}
\label{app:datasets:mixed}

\paragraph{Composition.} The primary fine-tuning runs use a weighted mixture of two real-world datasets: DL3DV and RE10K. We train on a selected subset of these datasets, with the exact number of scenes reported in \Cref{tab:mixed_training_config_sources}. The mixture is governed by a single \emph{per-source} weight $w_k$ (\Cref{tab:mixed_training_config_sources}). Operationally, the data loader proceeds as follows:

\begin{enumerate}
\item It converts each source weight into a \emph{per-scene} weight $w_k / |\mathcal{S}_k|$, where $|\mathcal{S}_k|$ denotes the number of scenes available from source $k$ on disk.
\item It passes these per-scene weights to the \texttt{WeightedSceneGroupedSampler}, which samples the next scene with a probability proportional to its per-scene weight.
\end{enumerate}

\begin{table}[h]
\centering
\small
\begin{tabular}{l c c c c}
\toprule
\textbf{Source} & \textbf{Num. of Scenes Loaded} & \textbf{Weight} & \textbf{Min gap} & \textbf{Max gap} \\
\midrule
DL3DV-10K & 8,559 & $0.35$ & $5$ & $25$ \\
RealEstate10K & 26,628 & $0.65$ & $10$ & $50$ \\
\bottomrule
\end{tabular}
\caption{\textbf{Mixed training mixture.} Two real-world sources are
sampled in proportion to the listed \emph{weight}. \emph{Min/Max gap} is
the temporal stride (in frames) between the source and target frame of a
training pair; the wider RE10K range reflects its faster intra-scene
camera motion.}
\label{tab:mixed_training_config_sources}
\end{table}
\paragraph{Depth and Pose Pre-extraction.}

We pre-extract depth maps and camera trajectories (intrinsics and extrinsics) offline using the DepthAnything-3 (DA3)~\citep{lin2025da3} backbone (\texttt{depth-anything/da3nested-giant-large}) in ``video'' mode with \texttt{process\_res\_method="upper\_bound\_resize"}. During preprocessing, we observe that DA3 estimations exhibit greater stability in multi-view settings than in monocular contexts; specifically, increasing the sequence length significantly enhances prediction precision and reliability. Motivated by this, we partition the DL3DV and RE10K datasets into 60-frame chunks, processing each independently at $960$\,px to maximize the temporal context provided to DA3 simultaneously.

Because DA3 jointly estimates both scene depth and camera geometry, the outputs within each chunk inherently share a consistent coordinate frame. To maintain this consistency, training pairs are sampled exclusively from within the same chunk using these DA3-estimated poses. For evaluation, however, we follow standard benchmark protocols and utilize the ground-truth COLMAP or Structure-from-Motion (SfM) poses provided natively by the datasets. Crucially, this offline precomputation allows the training loaders to stream static on-disk artifacts directly, eliminating the runtime computational overhead of re-running DA3..

\paragraph{Pre-processing.}
All RGB video frames are resized to $384 \times 384$. Depth supervision is provided using the precomputed DA3 predictions as soft targets. To normalize the camera configurations, we map them to a physical disparity range of $[d_{\min}, d_{\max}] = [0.02, 2.0]$.

\paragraph{Training-Time Pair Selection.}
To construct training pairs, both dataset loaders employ a \emph{chunk-aware random sampling} strategy. For any selected scene, a frame pair $(f_{\text{src}}, f_{\text{tgt}})$ is generated via the following pipeline:
\begin{enumerate}
    \item \textbf{Anchor Selection:} We randomly sample a source anchor frame $f_{\text{src}}$ from the set of frames covered by a valid DA3 chunk.
    \item \textbf{Candidate Filtering:} We construct a valid target candidate set $\mathcal{C} = \{f : |f - f_{\text{src}}| \in [g_{\min}, g_{\max}]\}$. Critically, this set is strictly restricted to the same chunk as $f_{\text{src}}$ to prevent mixing inconsistent coordinate frames across chunk boundaries.
    \item \textbf{Target Sampling:} The final target frame is drawn uniformly from the candidate set, denoted as $f_{\text{tgt}} \sim \mathcal{U}(\mathcal{C})$.
\end{enumerate}
% \noindent\textbf{Fallback Logic:} If the candidate set $\mathcal{C}$ is initially empty, we dynamically decrease the minimum frame gap $g_{\min}$ within the chunk boundaries. For degenerate, single-frame chunks where no other frames exist, the system falls back to a self-pair by reusing the anchor ($f_{\text{tgt}} = f_{\text{src}}$).

% \paragraph{Scene-Grouped Batching.}
% Loading per-scene DA3 outputs from disk on every iteration introduces significant I/O overhead. To amortize this cost, we implement a \emph{scene-grouped sampler}. Instead of sampling completely random scenes per batch, the data loader maintains a rolling pool of $\texttt{num\_cached\_scenes} = 16$ recently loaded scenes. Each training batch is then constructed by drawing $\texttt{pairs\_per\_scene\_batch} = 8$ distinct frame pairs from this cached pool.

% ==========================================================================
\subsection{Evaluation Pair Selection}
\label{app:datasets:eval}

\paragraph{Per-Benchmark Evaluation Configurations.}
\Cref{tab:benchmark_configuration} summarizes the configurations used across our evaluation sets. Pairs are sampled from within the 60-frame chunks defined during DA3 preprocessing, adhering to the dataset-specific stride and frame-gap constraints listed below.For DL3DV, we utilize the entire evaluation set, excluding a single corrupted scene. For RealEstate10K (RE10k), we evaluate on a test subset. Finally, for the Tanks~\&~Temples dataset, our evaluation is strictly restricted to the following seven scenes: \texttt{Barn}, \texttt{Caterpillar}, \texttt{Church}, \texttt{Courthouse}, \texttt{Ignatius}, \texttt{Meetingroom}, and \texttt{Truck}.

\begin{table}[ht]
\centering
\small
\begin{tabular}{l c c c c c}
\toprule
\textbf{Dataset} & \textbf{Stride} & \textbf{Min Gap} & \textbf{Max Gap} & \textbf{Num. Pairs} & \textbf{Num. Scenes} \\
\midrule
DL3DV-10K & $5$  & $5$  & $25$ & 28,780 & 54 \\
RealEstate10K   & $10$ & $10$ & $50$ & 18,084 & 1,991 \\
Tanks \& Temples       & $1$  & $1$  & $25$ & 5,642  & 7  \\
\bottomrule
\end{tabular}
\caption{Evaluation benchmark configurations, denoting the sequence frame stride, minimum/maximum frame separation gaps, total evaluated pairs, and scene counts.}
\label{tab:benchmark_configuration}
\end{table}

% \paragraph{Pseudo-code.}
% \begin{algorithm}[H]
% \caption{Deterministic evaluation pair selection per scene
% (mirrors \texttt{\_build\_eval\_pairs} in
% \texttt{src/datasets/dl3dv\_dataset.py} and the equivalent loops in
% \texttt{re10k\_da3\_dataset.py} / \texttt{tnt\_colmap\_dataset.py}).}
% \label{alg:eval_pairs}
% \begin{algorithmic}[1]
% \Require valid-frame count $n_{\text{raw}}$, stride $s$,
%   post-stride view count $N = \lfloor n_{\text{raw}} / s \rfloor$,
%   raw chunk windows $\mathcal{K} = \{(c_s, c_e)\}$
%   (strict DA3 partition; reconstructed from $n_{\text{raw}}$
%   for DL3DV-eval), max raw gap $g$
% \State $g_{\text{eff}} \leftarrow g \,/\, s$
%        \Comment{integer division; cap on strided index distance}
% \State for each strided index $i \in \{0, \ldots, N-1\}$:
%        $\text{cid}[i] \leftarrow$ id of the chunk containing
%        raw position $i \cdot s$, else $-1$
% \State $\mathcal{P} \leftarrow []$
% \ForAll{$(i, j) \in \mathrm{permutations}(\{0, \ldots, N-1\},\ 2)$}
%     \If{$\text{cid}[i] < 0$ \textbf{or} $\text{cid}[i] \ne \text{cid}[j]$}
%         \State \textbf{continue} \Comment{cross-chunk or orphaned}
%     \EndIf
%     \If{$|i - j| > g_{\text{eff}}$} \State \textbf{continue} \EndIf
%     \State append $(i, j)$ to $\mathcal{P}$
% \EndFor
% \State $\mathcal{P} \leftarrow \mathrm{sort}(\mathrm{set}(\mathcal{P}))$
%        \Comment{de-dup across overlapping chunks, then deterministic order}
% \State \Return $\mathcal{P}$
% \end{algorithmic}
% \end{algorithm}

%% file: appendix/training_config.tex
The full training configuration used to produce the evaluated checkpoints is reported in ~\Cref{tab:mixed_training_config} for the mixed-training run.

\begin{table}[h]
\centering
\small
\begin{tabular}{l l}
\toprule
\textbf{Hyper-parameter} & \textbf{Value} \\
\midrule
Base model & \texttt{Qwen/Qwen-Image-Edit-2509} \\
Transformer backbone & \texttt{linoyts/Qwen-Image-Edit-Rapid-AIO} \\
Lightning weights & \texttt{Qwen-Image-Edit-2511-Lightning-8steps-V1.0-bf16} \\
LoRA rank / $\alpha$ & $128$ / $128$ ($64$ / $64$ for Blender training)\\
LoRA target modules & q, k, v, out, MLP, mod, in, proj\_out \\
LoRA learning rate & $5 \times 10^{-5}$ \\
ControlNet layers / scale & $5$ / $1.0$ \\
% Image size & $384 \times 384$ \\
% Optimizer & AdamW, $\beta=(0.9,0.999)$, wd $=10^{-2}$ \\
Learning rate & $1\times10^{-4}$ (cosine, $\min=10^{-6}$) \\
% Batch size $\times$ grad accum & $1\times1$ (per GPU) \\
Mixed precision & bf16 \\
Gradient clipping & $1.0$ \\
EMA decay (from step $10$k) & $0.99995$ \\
Loss weights & $w_{\text{fm}}{=}1.0$, $w_{\text{fm,tgt-rgb}}{=}1.0$, $w_{\text{fm,tgt-dep}}{=}0.75$, $w_{\text{recon}}{=}0.5$, $w_{\text{lpips}}{=}0.25$, $w_{\text{dino}}{=}0.02$ \\
Timestep sampling & logit-normal, shift $=1.5$, base $=4096$ \\
\bottomrule
\end{tabular}
\caption{\textbf{Mixed training configuration.} Values reproduced verbatim from the released \texttt{config.json} (mixed-DA3 run).}
\label{tab:mixed_training_config}
\end{table}

\subsection{Pl\"ucker Ray Map Conditioning}
To condition the ControlNet, target camera rays are processed and injected via three steps:
\begin{itemize}
    \item 
    \textbf{Pl\"ucker Mapping:} Source and target rays are back-projected into a 6D Pl\"ucker coordinate map.
    \item 
    \textbf{Lossless Tokenization:} The map is deterministically reshaped into flat geometric tokens, bypassing learned convolutions.
    \item
    \textbf{Zero-Init Injection:} A zero-initialized linear layer projects these tokens to the transformer width. They are then added element-wise to the image latents, ensuring a stable "no-op" initialization at the start of training.
\end{itemize}

%% file: appendix/PCC_materials/main.tex
\subsection{Scale Jitter Ablation}\label{app:jitter_test}
\input{appendix/PCC_materials/jitter_test}

\newpage
\subsection{Scene Visualizations of the Occlusion Ablation}\label{app:occlusion_test}
\input{appendix/PCC_materials/occlution_test}

\newpage
\subsection{Scale Alignment for PCC}\label{app:scale_alignment_pcc}
\input{appendix/PCC_materials/scale_alignment_and_auc}

\subsection{Camera-Pose Accuracy}\label{app:pose_auc}
\input{appendix/PCC_materials/pose_auc}

\subsection{Visualization of Scale-Alignment for DA3}
\begin{figure}[h]
  \centering
  \includegraphics[width=0.90\linewidth]{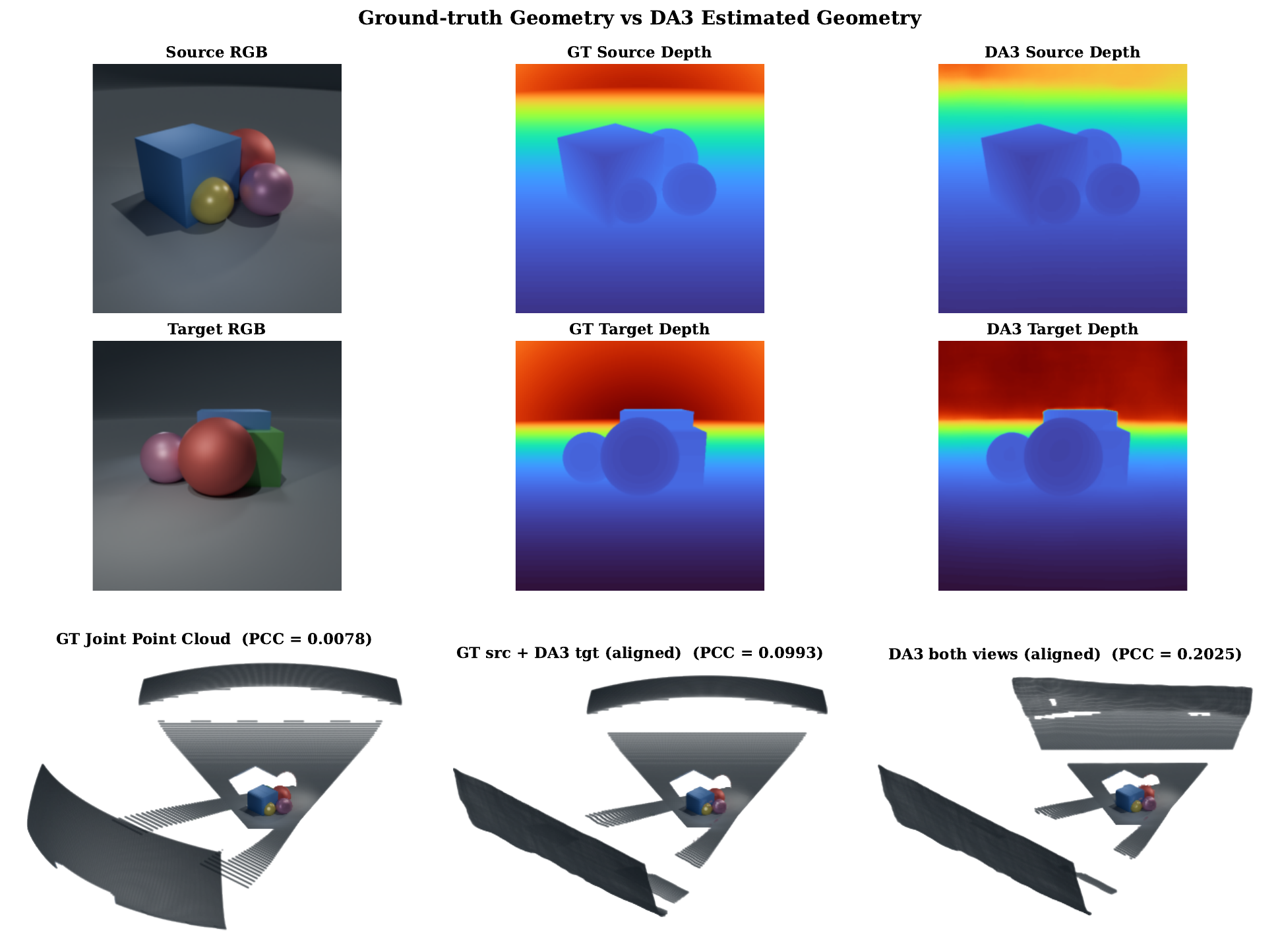}
  \caption{\textbf{Ground-truth versus DA3-estimated geometry for a two-view pair.} We visualize three distinct 3D reconstructions of a two-view scene. \textbf{Top two rows:} The source and target RGB inputs alongside their corresponding ground-truth (GT) depth maps (from Blender) and scale-aligned DA3 depth predictions, all color-coded by distance. \textbf{Bottom row (left to right):} (1) The reference GT joint point cloud, obtained by unprojecting both GT depths using GT camera poses (PCC = 0.008). (2) A mixed reconstruction utilizing the source GT depth and the target DA3 depth. The target depth is affine-aligned to the GT scale using a foreground/background split (determined via Otsu's thresholding on the pooled GT depth) to specifically isolate the error introduced by DA3 (PCC = 0.099). (3) The fully estimated point cloud, unprojected using DA3 depths and GT cameras for both views(PCC = 0.203).}
  \label{app:fig:gt_vs_da3}
\end{figure}

\subsection{PCC Evaluation in Practice}
\input{appendix/PCC_materials/pcc_in_practice}

\subsection{PCC Pseudo Code}
\input{appendix/PCC_materials/pcc_pseudocode}

\subsection{Differentiable PCC Loss via Inverse Rendering}\label{app:differentiable_pcc}
\input{appendix/PCC_materials/differentiable_pcc}

\subsection{Quantitative Results of the Jitter and Occlusion Experiments}
\begin{table}[h]
\centering
\small
\setlength{\tabcolsep}{4pt}
\begin{tabular}{l ccc c cccc}
\toprule
& \multicolumn{3}{c}{2D image metrics} & & \multicolumn{4}{c}{3D / camera pose metrics} \\
\cmidrule(lr){2-4}\cmidrule(lr){6-9}
& PSNR$\uparrow$ & SSIM$\uparrow$ & LPIPS$\downarrow$ & PCC$_\text{GT}\!\downarrow$
& PCC$_\text{DA3}\!\downarrow$ & AUC@5$\uparrow$ & RPE-R$\downarrow$ & RPE-T$\downarrow$ \\
\midrule
Baseline            & $\infty$ & 1.000 & 0.000 & 0.002 & 0.074 & 0.743 & 12.79 & 0.77 \\
\midrule
Jitter $\pm$1\%     & 42.1     & 0.996 & 0.004 & 0.004 & 0.095 & 0.700 & 13.08 & 0.79 \\
Jitter $\pm$2\%     & 39.1     & 0.990 & 0.008 & 0.006 & 0.123 & 0.686 & 12.93 & 0.93 \\
Jitter $\pm$5\%     & 34.6     & 0.976 & 0.021 & 0.015 & 0.232 & 0.486 & 13.23 & 2.22 \\
Jitter $\pm$10\%    & 31.1     & 0.962 & 0.043 & 0.029 & 0.346 & 0.329 & 13.87 & 4.12 \\
Jitter $\pm$20\%    & 28.1     & 0.948 & 0.088 & 0.052 & 0.501 & 0.086 & 14.31 & 8.76 \\
Jitter $\pm$40\%    & 25.5     & 0.935 & 0.149 & 0.083 & 0.615 & 0.057 & 14.70 & 12.35 \\
\midrule
Occlusion (avg.)    & 34.7     & 0.978 & 0.064 & 0.005 & --    & --    & --    & --    \\
\bottomrule
\end{tabular}
\caption{Mean metrics across views for the scale-jitter and occlusion experiments. Under scale jitter, 2D quality (PSNR/SSIM/LPIPS) and the DA3-based geometry/pose metrics (PCC$_\text{DA3}$, AUC@5, RPE-T) all degrade monotonically as the perturbation grows; rotation error RPE-R stays in a narrow $\sim$13–15 band, confirming that scale ambiguity manifests primarily as translation drift. Under occlusion, where each scene variant is internally 3D-consistent but visually different from the baseline, the 2D metrics respond: \textbf{PCC$_\text{GT}$ remains at the baseline level} ($0.005\!\approx\!0.002$), correctly recognising each variant as geometrically consistent.}
\label{tab:pcc_summary}
\end{table}

As shown in \Cref{tab:pcc_summary}, all four metrics degrade monotonically with increasing jitter. At $\pm 40\%$ jitter, PCC rises from $0.001$ to $0.044$, PSNR drops from $\infty$ to $25.5$\,dB, SSIM from $1.0$ to $0.935$, and LPIPS from $0$ to $0.149$. This confirms that when geometric distortion directly changes pixel content, all metrics respond in agreement. 

% \subsection{Scene Visualizations of the Counterfactual Ablation}
% \input{appendix/PCC_materials/counterfactual_test}

\input{appendix/PCC_materials/multiview}

%% file: appendix/PCC_materials/jitter_test.tex
\begin{figure}[h]
\centering
\setlength{\tabcolsep}{1pt}
\renewcommand{\arraystretch}{0.3}
\begin{tabular}{@{} m{1.5em} @{\hspace{3pt}} m{0.23\linewidth} m{0.23\linewidth} m{0.23\linewidth} m{0.23\linewidth} @{}}
 & \centering\small View 01 & \centering\small View 02 & \centering\small View 03 & \centering\arraybackslash\small View 04 \\[4pt]

\rotatebox[origin=c]{90}{\small Baseline} &
\includegraphics[width=\linewidth]{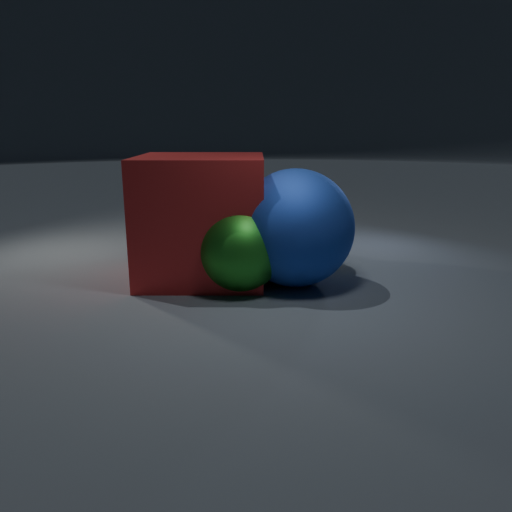} &
\includegraphics[width=\linewidth]{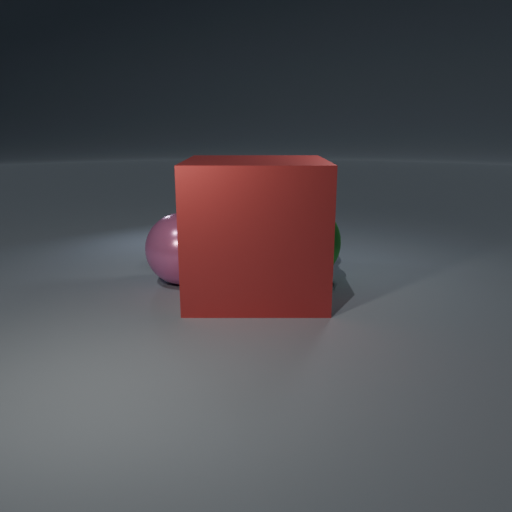} &
\includegraphics[width=\linewidth]{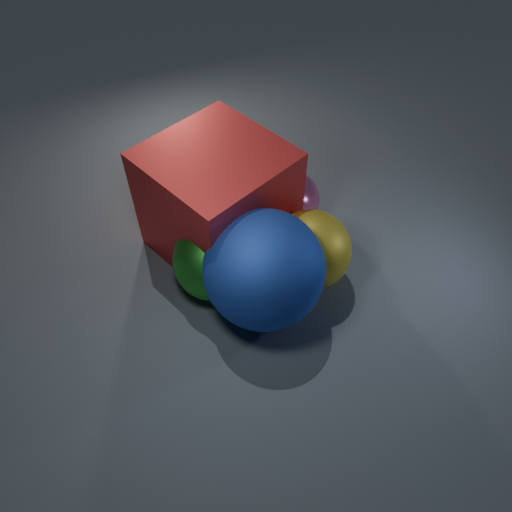} &
\includegraphics[width=\linewidth]{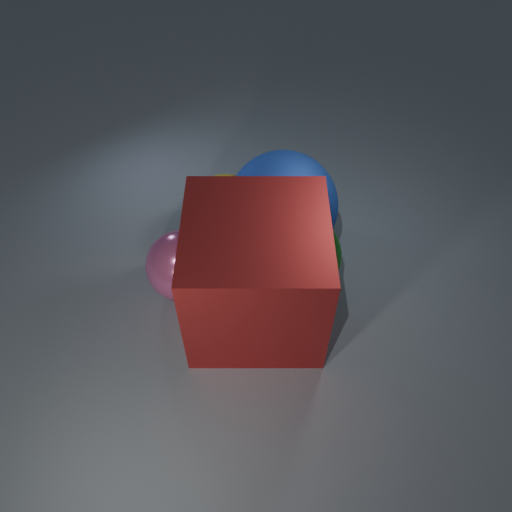} \\[-1pt]

& \centering\small\textcolor{pcccolor}{PCC:0.001}&
\centering\small\textcolor{pcccolor}{PCC:0.001} &
\centering\small\textcolor{pcccolor}{PCC:0.001} &
\centering\arraybackslash\small\textcolor{pcccolor}{PCC:0.002} \\[3pt]

\rotatebox[origin=c]{90}{\small Jitter $\pm$5\%} &
\includegraphics[width=\linewidth]{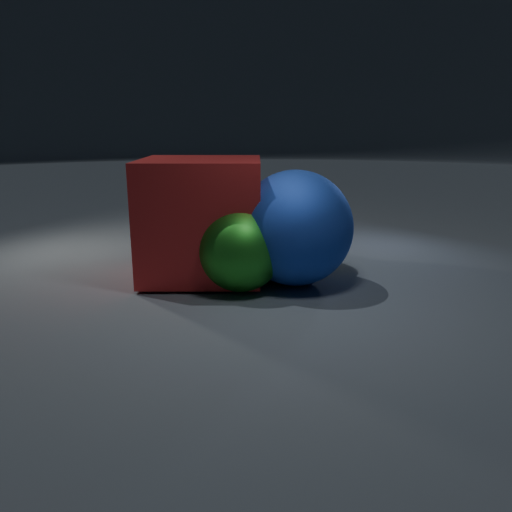} &
\includegraphics[width=\linewidth]{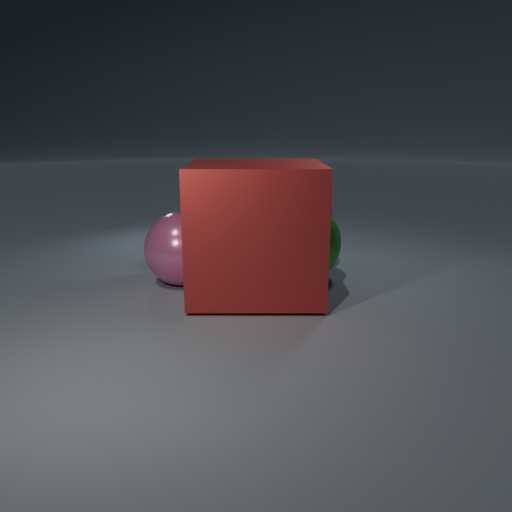} &
\includegraphics[width=\linewidth]{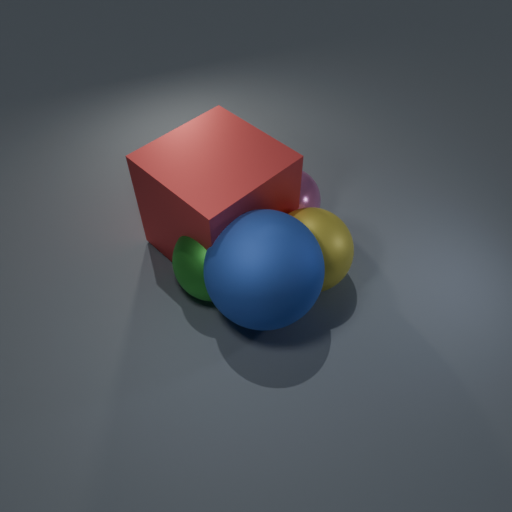} &
\includegraphics[width=\linewidth]{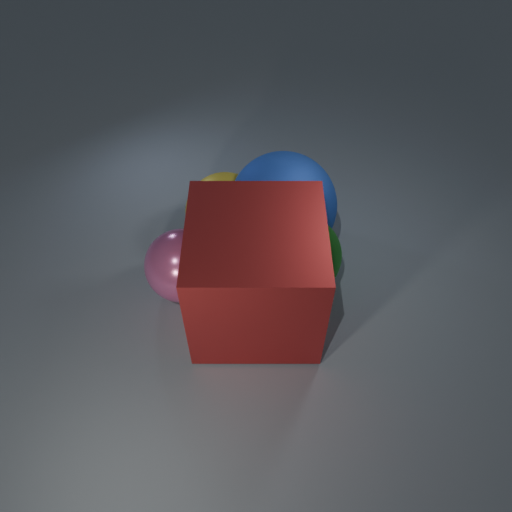} \\

& \centering\small\textcolor{pcccolor}{PCC:0.017} \quad\textcolor{psnrcolor}{PSNR:35.8} &
\centering\small\textcolor{pcccolor}{PCC:0.019} \quad\textcolor{psnrcolor}{PSNR:35.2} &
\centering\small\textcolor{pcccolor}{PCC:0.010} \quad\textcolor{psnrcolor}{PSNR:34.9} &
\centering\arraybackslash\small\textcolor{pcccolor}{PCC:0.020}\quad \textcolor{psnrcolor}{PSNR:35.5} \\[3pt]

& \centering\small\textcolor{ssimcolor}{SSIM:0.98} \quad\textcolor{lpipscolor}{LPIPS:0.02} &
\centering\small\textcolor{ssimcolor}{SSIM:0.98} \quad\textcolor{lpipscolor}{LPIPS:0.03} &
\centering\small\textcolor{ssimcolor}{SSIM:0.98} \quad\textcolor{lpipscolor}{LPIPS:0.02} &
\centering\arraybackslash\small\textcolor{ssimcolor}{SSIM:0.98}\quad \textcolor{lpipscolor}{LPIPS:0.02} \\[3pt]

\rotatebox[origin=c]{90}{\small Jitter $\pm$20\%} &
\includegraphics[width=\linewidth]{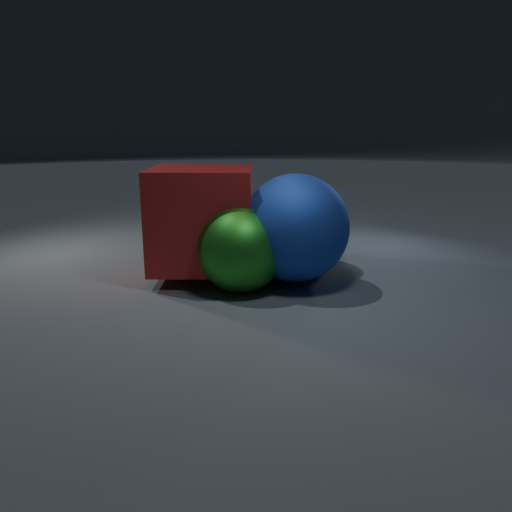} &
\includegraphics[width=\linewidth]{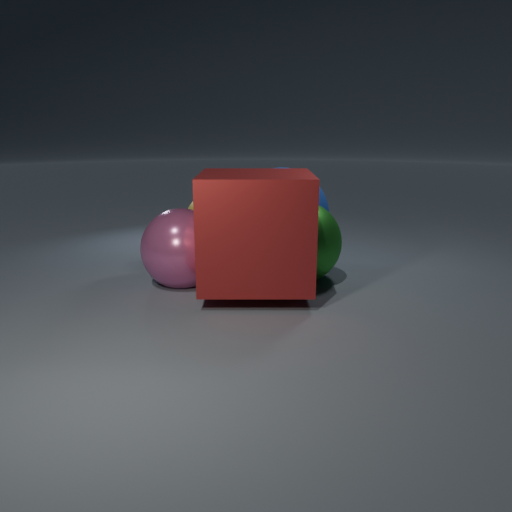} &
\includegraphics[width=\linewidth]{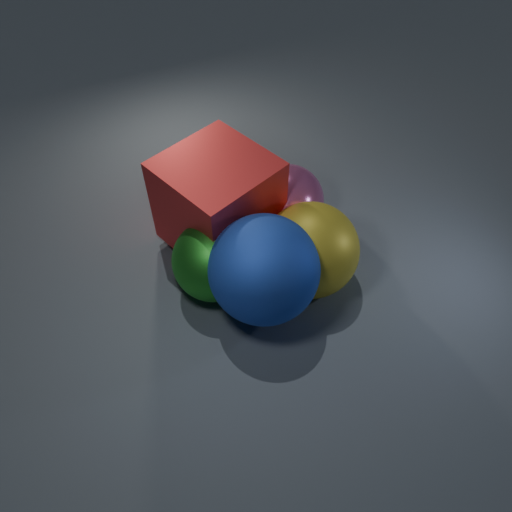} &
\includegraphics[width=\linewidth]{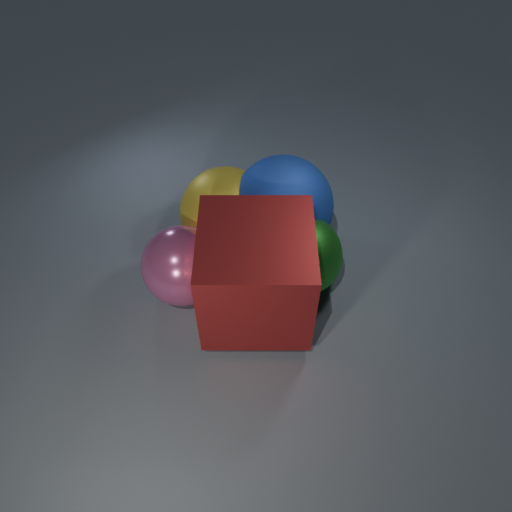} \\

& \centering\small\textcolor{pcccolor}{PCC:0.057}\quad \textcolor{psnrcolor}{PSNR:29.6} &
\centering\small\textcolor{pcccolor}{PCC:0.052}\quad \textcolor{psnrcolor}{PSNR:28.9} &
\centering\small\textcolor{pcccolor}{PCC:0.038}\quad\textcolor{psnrcolor}{PSNR:28.0} &
\centering\arraybackslash\small\textcolor{pcccolor}{PCC:.070} \quad\textcolor{psnrcolor}{PSNR:28.5}\\[3pt]

& \centering\small\textcolor{ssimcolor}{SSIM:0.95} \quad\textcolor{lpipscolor}{LPIPS:0.07} &
\centering\small\textcolor{ssimcolor}{SSIM:0.96}\quad\textcolor{lpipscolor}{LPIPS:0.11} &
\centering\small\textcolor{ssimcolor}{SSIM:0.94}\quad \textcolor{lpipscolor}{LPIPS:0.09} &
\centering\arraybackslash\small \textcolor{ssimcolor}{SSIM:0.94} \quad\textcolor{lpipscolor}{LPIPS:0.08} \\

\end{tabular}

\caption{Scale jitter experiment. Baseline (top) with no jitter, and two variants where each object is randomly scaled by $\pm$5\% and $\pm$20\%. All metrics degrade together: \textcolor{pcccolor}{PCC} increases (worse 3D consistency), while \textcolor{psnrcolor}{PSNR}/\textcolor{ssimcolor}{SSIM}/\textcolor{lpipscolor}{LPIPS} decrease (worse visual similarity). Unlike the occlusion experiment, scale jitter is detectable by all metrics since it changes both geometry and appearance uniformly.}
\label{fig:jitter_comparison}
\end{figure}

%% file: appendix/PCC_materials/occlution_test.tex
\begin{figure}[h]
\centering
\setlength{\tabcolsep}{1pt}
\renewcommand{\arraystretch}{0.3}
% C = centered, justified m-column used for all four image/metric cells, so
% every cell has IDENTICAL alignment behaviour (avoids the top-right tile
% drifting because only the last cell carried \arraybackslash before).
\newcolumntype{C}{>{\centering\arraybackslash}m{0.23\linewidth}}
\begin{tabular}{@{} m{1.5em} @{\hspace{3pt}} C C C C @{}}
 & \small View 01 (anchor) & \small View 02 & \small View 03 & \small View 04 \\[4pt]

\rotatebox[origin=c]{90}{\small Scene A} &
\includegraphics[width=\linewidth]{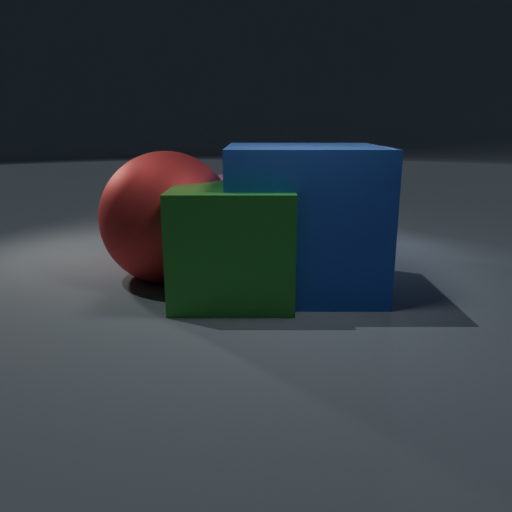} &
\includegraphics[width=\linewidth]{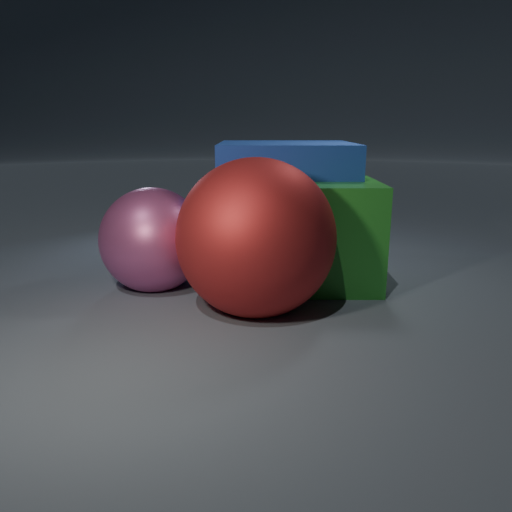} &
\includegraphics[width=\linewidth]{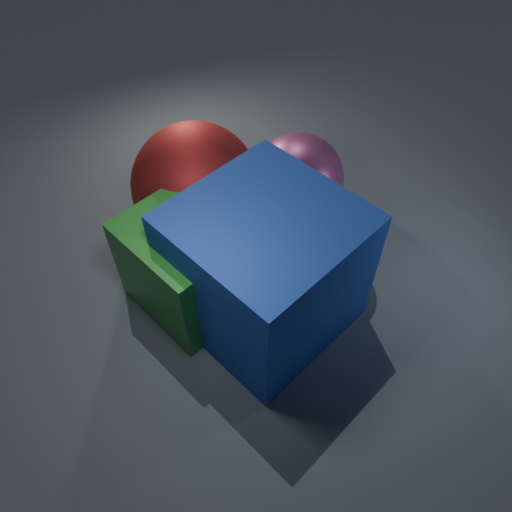} &
\includegraphics[width=\linewidth]{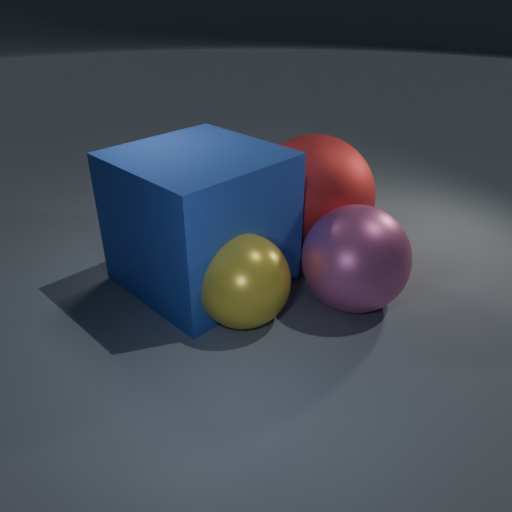} \\[3pt]

& &
\small\textcolor{pcccolor}{PCC:0.007} &
\small\textcolor{pcccolor}{PCC:0.002} &
\small\textcolor{pcccolor}{PCC:0.007} \\[6pt]

\rotatebox[origin=c]{90}{\small Scene B\textsubscript{1}} &
\includegraphics[width=\linewidth]{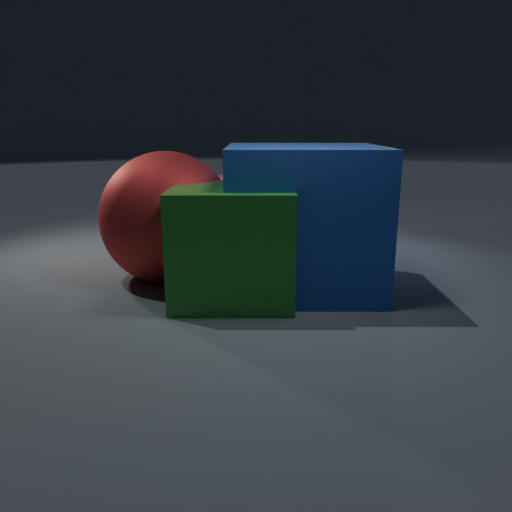} &
\includegraphics[width=\linewidth]{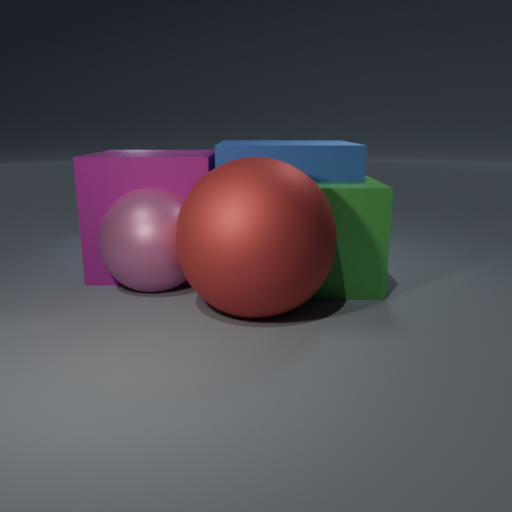} &
\includegraphics[width=\linewidth]{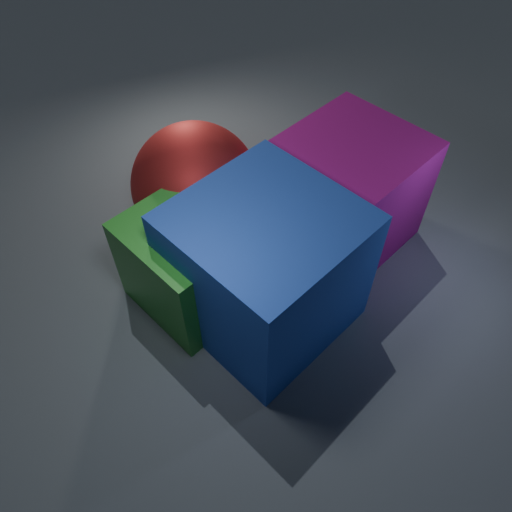} &
\includegraphics[width=\linewidth]{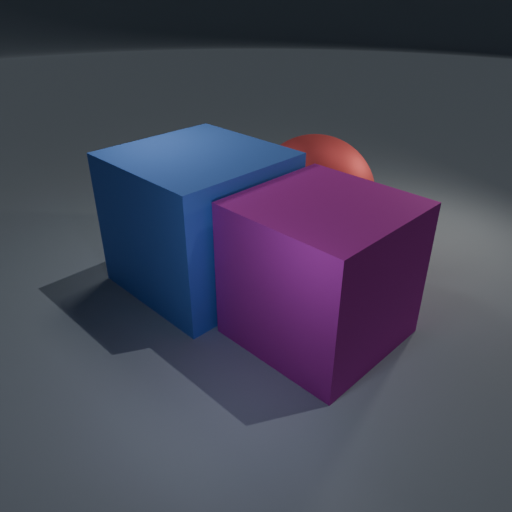} \\

& \small\textcolor{pcccolor}{PCC:0.000} \quad \textcolor{psnrcolor}{PSNR:56.9} &
\small\textcolor{pcccolor}{PCC:0.007} \quad \textcolor{psnrcolor}{PSNR:30.0} &
\small\textcolor{pcccolor}{PCC:0.002} \quad \textcolor{psnrcolor}{PSNR:30.4} &
\small\textcolor{pcccolor}{PCC:0.008} \quad \textcolor{psnrcolor}{PSNR:23.0} \\[3pt]

& \small \textcolor{ssimcolor}{SSIM:1.00} \quad \textcolor{lpipscolor}{LPIPS:0.00} &
\small \textcolor{ssimcolor}{SSIM:0.98} \quad \textcolor{lpipscolor}{LPIPS:0.08} &
\small\textcolor{ssimcolor}{SSIM:0.98} \quad \textcolor{lpipscolor}{LPIPS:0.12} &
\small\textcolor{ssimcolor}{SSIM:0.93} \quad\textcolor{lpipscolor}{LPIPS:0.22} \\[3pt]

\rotatebox[origin=c]{90}{\small Scene B\textsubscript{2}} &
\includegraphics[width=\linewidth]{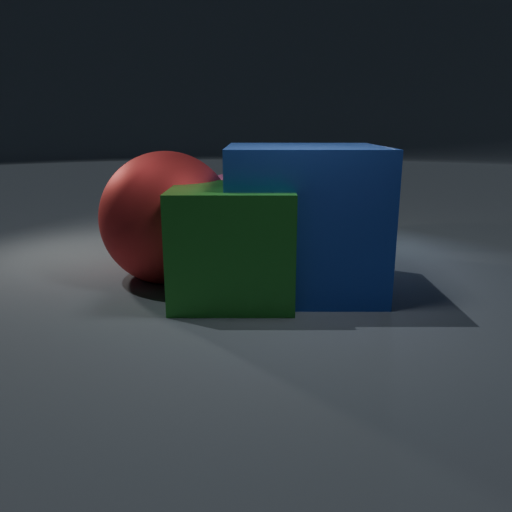} &
\includegraphics[width=\linewidth]{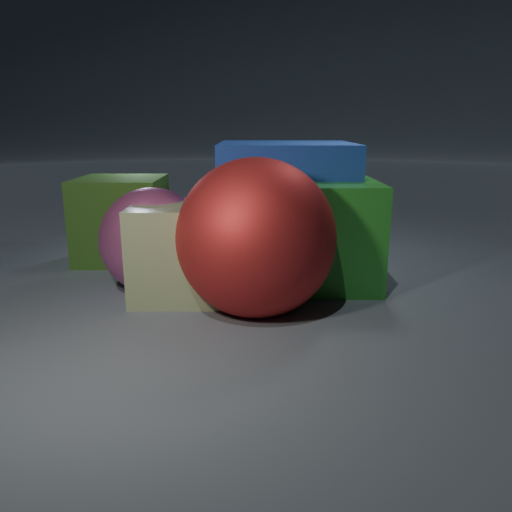} &
\includegraphics[width=\linewidth]{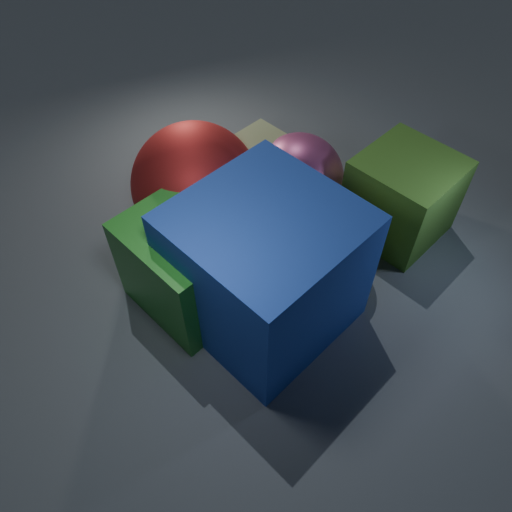} &
\includegraphics[width=\linewidth]{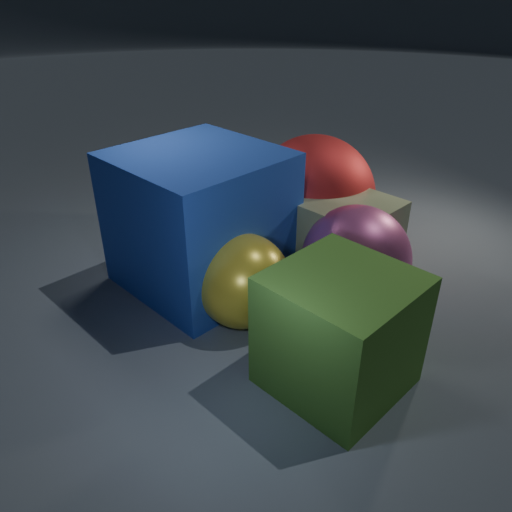} \\

& \small\textcolor{pcccolor}{PCC:0.000} \quad \textcolor{psnrcolor}{PSNR:57.3} &
\small\textcolor{pcccolor}{PCC:0.007} \quad \textcolor{psnrcolor}{PSNR:30.3} &
\small\textcolor{pcccolor}{PCC:0.002} \quad \textcolor{psnrcolor}{PSNR:29.1} &
\small\textcolor{pcccolor}{PCC:0.007} \quad \textcolor{psnrcolor}{PSNR:23.8} \\

& \small\textcolor{ssimcolor}{SSIM:1.00} \quad\textcolor{lpipscolor}{LPIPS:0.00} &
\small\textcolor{ssimcolor}{SSIM:0.98}\quad \textcolor{lpipscolor}{LPIPS:0.07} &
\small\textcolor{ssimcolor}{SSIM:0.98} \quad \textcolor{lpipscolor}{LPIPS:0.10} &
\small \textcolor{ssimcolor}{SSIM:0.94} \quad \textcolor{lpipscolor}{LPIPS:0.20} \\
\end{tabular}
\caption{Occlusion experiment. Scene~A (top) and two variants B\textsubscript{1}, B\textsubscript{2} where objects are added behind the anchor camera (View~01). Metrics: \textcolor{pcccolor}{PCC}~/ \textcolor{psnrcolor}{PSNR}~/ \textcolor{ssimcolor}{SSIM}~/ \textcolor{lpipscolor}{LPIPS}. For Scene~A, 2D metrics are perfect since each view is compared to itself. For Scene~B variants, PCC remains unchanged ($\approx$Scene~A) because each scene is internally 3D-consistent, while \textcolor{psnrcolor}{PSNR}/\textcolor{ssimcolor}{SSIM}/\textcolor{lpipscolor}{LPIPS} correctly detect the visual differences from added objects.}
\label{fig:occlusion_comparison}
\end{figure}

%% file: appendix/PCC_materials/scale_alignment_and_auc.tex
Because the multi-view geometry estimator (i.e DA3~\citep{lin2025da3}) jointly predicts depth and camera poses from RGB input alone, its outputs are defined only up to a global similarity transformation. Since each inference step minimizes multi-view photometric and geometric inconsistencies, processing frames from two different versions of a scene (e.g., a reference and a perturbed variant) forces the model to artificially deform the recovered geometry to ensure the reconstructions overlap. As a result, the estimator silently absorbs the true 3D inconsistency into its pose and scale predictions, meaning the final measured residual reflects only the leftover error rather than the actual 3D mismatch.

To resolve this ambiguity, we fit a single affine transformation, shared by the source and target views, that aligns the DA3-predicted depths to ground-truth metric depths over all valid pixels in both views:
\begin{equation}
\alpha^\star,\, \beta^\star \;=\; \operatorname*{arg\,min}_{\alpha,\,\beta}
  \sum_{v \in \{s, t\}}\sum_{p \in \mathcal{V}_v}
    \bigl(\alpha\,\hat{D}_v(p) + \beta - D_v^{\mathrm{gt}}(p)\bigr)^2,
\label{eq:scale_align}
\end{equation}
where $v \in \{s,t\}$ indexes the source ($s$) or target ($t$) view, $p$ denotes a pixel location, $\hat{D}_v(p)$ is the depth predicted by DA3 at pixel $p$ in view $v$, and $D_v^{\mathrm{gt}}(p)$ is the corresponding ground-truth metric depth. The set $\mathcal{V}_v = \{p \mid \hat{D}_v(p) > 0 \wedge D_v^{\mathrm{gt}}(p) > 0\}$ contains pixels for which both depths are valid and positive. The variables $\alpha$ and $\beta$ are, respectively, the shared multiplicative scale and additive shift, and $(\alpha^\star,\beta^\star)$ are their least-squares-optimal values. Using one shared pair for both views preserves their relative depth scale rather than aligning each depth map independently.

The fitted parameters $(\alpha^{*}, \beta^{*})$ are then applied to both the source- and target-view depths:$$\tilde{\mathbf{D}} \;=\; \max\!\bigl(\alpha^{*}\hat{\mathbf{D}} + \beta^{*},\;0\bigr).$$This operation ensures that the relative depth structure predicted by DA3 across the two views is preserved, while the global scale and offset remain strictly locked to the GT metric scale.

\paragraph{Foreground/background scale alignment.}
In synthetic scenes with a plain backdrop, we fit \emph{two} independent affines, one for the foreground and one for the background. The foreground/background partition is obtained automatically by 1-D \emph{Otsu thresholding}~\citep{otsu1979otsu} on the pooled GT depths of the source and target views: Otsu picks the threshold $t^\ast$ that maximizes the between-class variance of the depth histogram. Within each set we pool DA3/GT pixel
pairs from both views and solve a $2\!\times\!2$ least-squares system for $(\alpha_{\text{fg}}, \beta_{\text{fg}})$ and $(\alpha_{\text{bg}}, \beta_{\text{bg}})$.
At inference time each pixel is aligned with the affine corresponding to
its own GT bin:
\[
\hat d_{\text{aligned}}(p) =
\begin{cases}
\alpha_{\text{bg}}\,\hat d_{\text{DA3}}(p) + \beta_{\text{bg}}, & \text{if } d_{\text{GT}}(p) \ge t^\ast,\\
\alpha_{\text{fg}}\,\hat d_{\text{DA3}}(p) + \beta_{\text{fg}}, & \text{otherwise.}
\end{cases}
\]
This keeps foreground geometry tight while still placing the dome at a
plausible distance behind it, removing the FG/BG cluster split that a
single global affine would otherwise produce.

\subsection{Far Depth Pruning}
Multi-view geometry estimators frequently produce unbounded or thresholded depth predictions in sky and far-background regions. To prevent these outliers from skewing both the scale-alignment computation and the reprojection loss, we occasionally mask out the highest $\rho=3-5\%$ of valid \emph{target-view} depths prior to alignment. We specify in the experimental section whenever this far-depth masking is utilized.

%% file: appendix/PCC_materials/pose_auc.tex
To quantify camera-pose accuracy, we compare the DA3-predicted camera poses from the two views against the ground truth. We adopt established pose evaluation metrics~\citep{lin2025da3, wang2024posediffusion, he2024cameractrl, song2026worldforge}, utilizing DA3 as our designated geometry estimator. Specifically, we calculate the relative rotation error (RPE-R) $e_R$ and relative translation error(RPE-T) $e_t$ for the relative pose between views:
$$\begin{aligned} e_R &= \arccos\!\left( \frac{ \mathrm{tr}\!\bigl( \hat{\mathbf{R}}_\mathrm{rel}^\top \mathbf{R}_\mathrm{rel}^\mathrm{gt} \bigr) - 1 }{2} \right), \\[4pt] e_t &= \arccos\!\bigl( \hat{\mathbf{t}}_\mathrm{rel} \cdot \mathbf{t}_\mathrm{rel}^\mathrm{gt} \bigr), \end{aligned}$$
where $\mathbf{R}_\mathrm{rel}$ and $\mathbf{t}_\mathrm{rel}$ represent the rotation and unit translation direction of the relative transformation, and $\hat{\cdot}$ denotes DA3 estimates. The maximum of these forms the pairwise pose error: $e_\mathrm{pose} = \max(e_R,\, e_t)$. We report the area under the cumulative error curve (AUC) up to threshold $\theta$:$$\mathrm{AUC}@\theta \;=\; \frac{1}{\theta}\int_0^{\theta} \mathbf{1}\!\bigl[e_\mathrm{pose} \le \alpha\bigr]\, \mathrm{d}\alpha,$$for $\theta \in \{3, 5, 15, 30\}$. Higher AUC values confirm that the generated view possesses sufficient geometric consistency for accurate relative pose recovery.

%% file: appendix/PCC_materials/pcc_in_practice.tex
\begin{figure}[H]
  \centering
  \includegraphics[width=.9\linewidth]{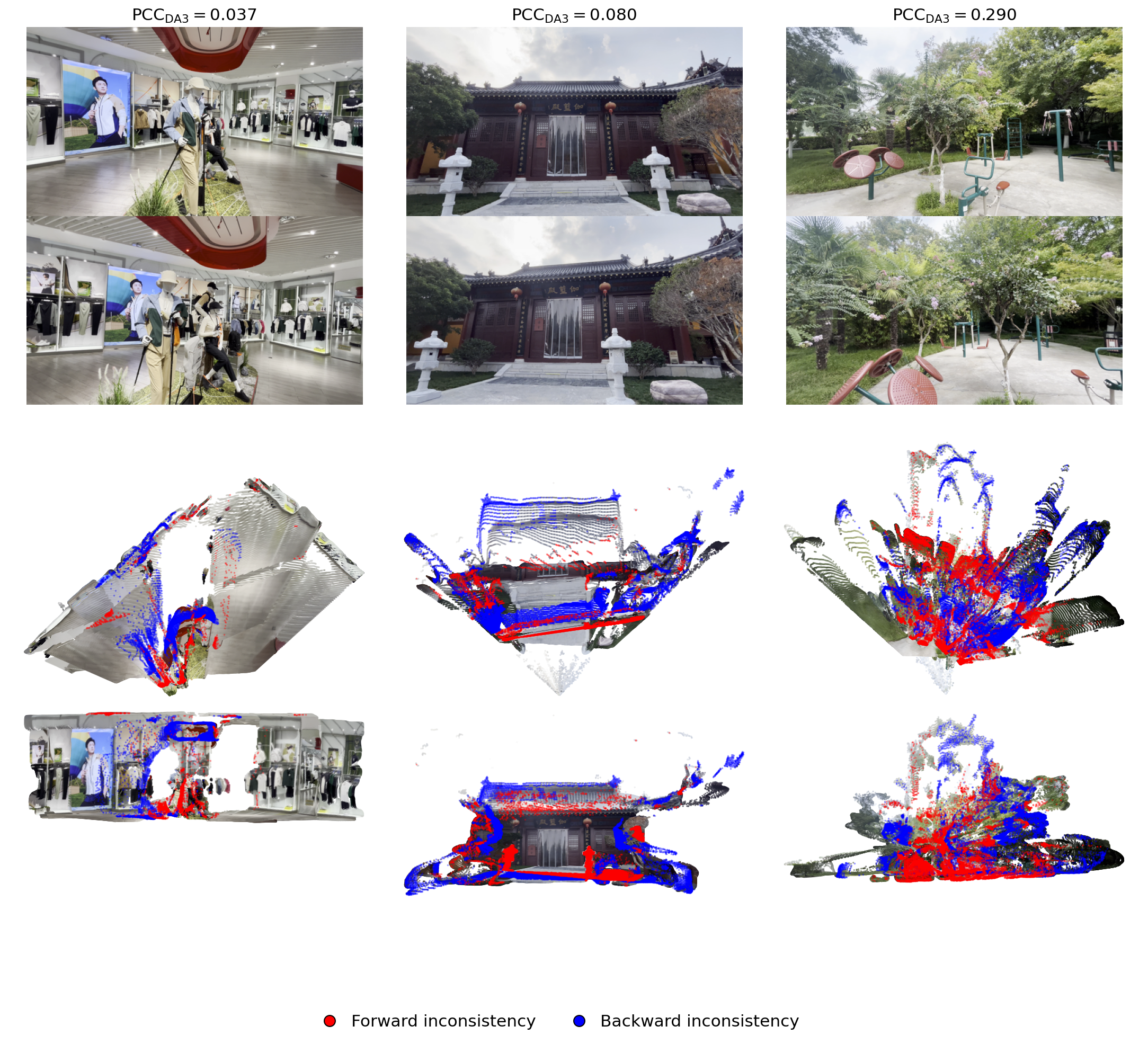}
  \caption{\textbf{Projection Consistency Constraint (PCC$_{\text{DA3}}$) on DL3DV.} Each column shows a source–target view pair: the top two rows are the source and target RGB frames, and the bottom two rows show the fused two-view point cloud (top-down and front) built from the GT source depth and DA3-estimated target depth (FG/BG affine-aligned), reprojected under the ground-truth poses. Consistent points keep their RGB color; red/blue highlight forward/backward reprojection inconsistencies, and the per-column PCC$_{\text{DA3}}$ score (lower is better) quantifies them.}\label{app:fig:projection_consistency_dl3dv}
\end{figure}
\begin{figure}[h]
  \centering
  \includegraphics[width=.9\linewidth]{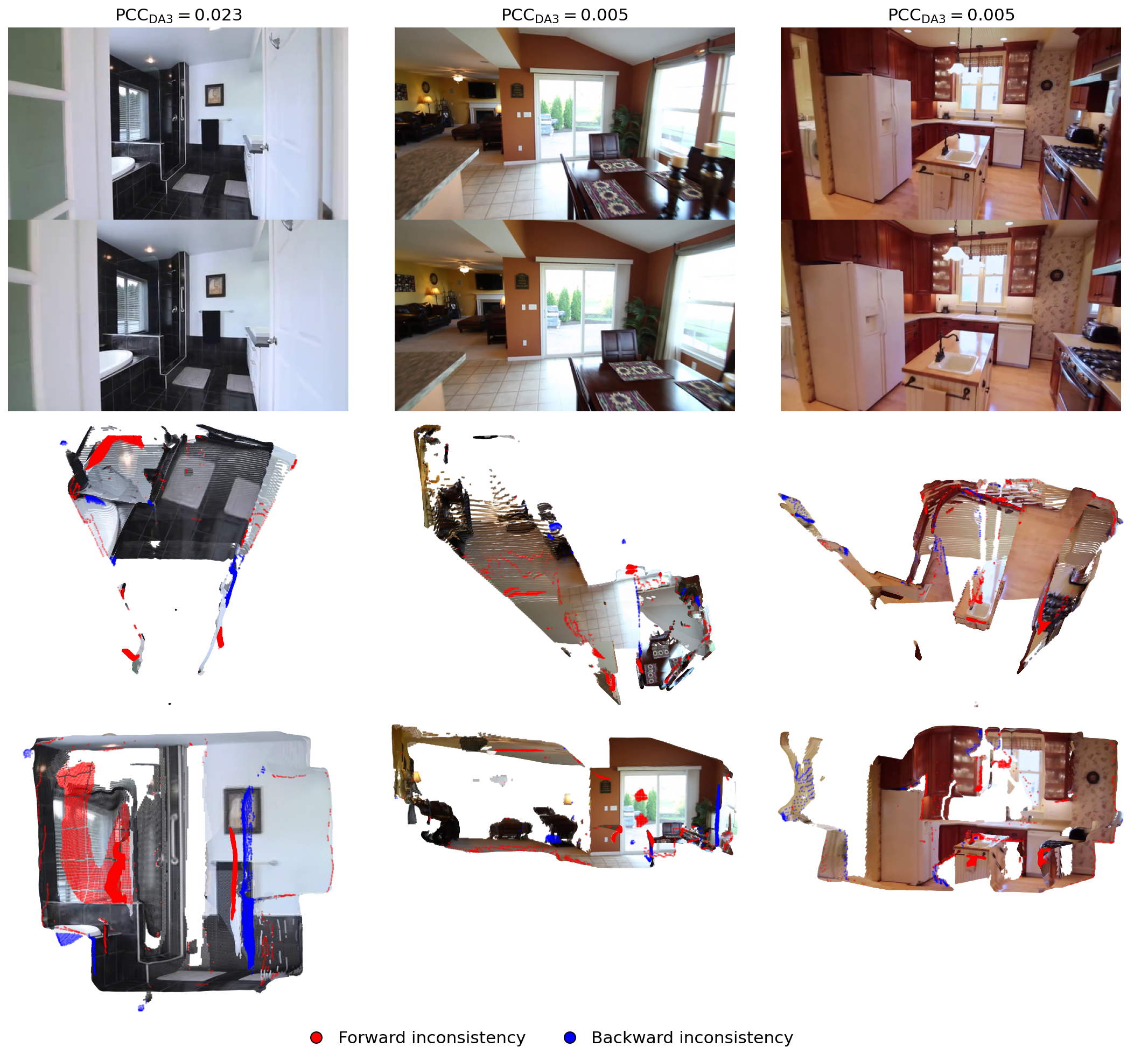}
  \caption{\textbf{Projection Consistency Constraint (PCC$_{\text{DA3}}$) on re10k.} Each column shows a source–target view pair: the top two rows are the source and target RGB frames, and the bottom two rows show the fused two-view point cloud (top-down and front) built from the GT source depth and DA3-estimated target depth (FG/BG affine-aligned), reprojected under the ground-truth poses. Consistent points keep their RGB color; red/blue highlight forward/backward reprojection inconsistencies, and the per-column PCC$_{\text{DA3}}$ score (lower is better) quantifies them.}\label{app:fig:projection_consistency_re10k}
\end{figure}
Computing the PCC requires four primary inputs: the source depth, the target depth, and their respective camera poses. Crucially, the origin of these inputs—specifically, whether they are extracted from ground-truth data or estimated by a 3D reconstruction model—heavily influences the resulting metric values.

Because we employ a 3D reconstruction framework (DA3) to obtain these inputs, we outline our specific input configurations, dataset-dependent processing pipelines, and baselines below.

\paragraph{Camera Pose Configuration} While an alternative approach would be to utilize the camera poses estimated directly by DA3, we strictly enforce the use of the dataset's ground-truth camera poses (derived from either COLMAP or the simulation environment) during the PCC evaluation. This design choice is critical for alignment with our core task: since our objective is to evaluate whether the generated output maintains 3D consistency conditioned on a specific camera prompt, we must measure performance against the exact geometric trajectory requested by that prompt.

\paragraph{Depth Processing and DA3 View-Sensitivity} The pipeline for obtaining source and target depths depends on whether the underlying dataset is simulated or real-world. A key characteristic of DA3 is its high sensitivity to the number of input views; its depth and geometry estimations become significantly more stable and accurate as the number of viewpoints increases. Because Blender datasets provide exact ground-truth depth while real-world, in-the-wild datasets do not, we handle them using two distinct paradigms:
\begin{itemize}
    \item 
    \textbf{Blender Simulated Scenes:} In these environments, exact ground-truth depth is available for the source view. Although DA3 can process a sparse 2-view input (the source and the generated target view) to estimate both depths simultaneously, its performance degrades severely in such few-view settings due to inherent reconstruction ambiguities (as illustrated in the rightmost panel of \Cref{app:fig:gt_vs_da3}). To prevent this degradation from polluting our evaluation, we only utilize DA3 to estimate the depth of the generated target view. We then align this estimated target depth with Blender’s ground-truth target depth using the foreground-background alignment algorithm detailed in \Cref{app:scale_alignment_pcc}. \Cref{app:fig:gt_vs_da3} visualizes the final 3D reconstructed scene alongside its corresponding PCC value.
    \item 
    \textbf{In-the-Wild Static Scenes:} Real-world video sequences lack ground-truth depth entirely. To leverage DA3’s strength in dense-view scenarios, we preprocess the video data by partitioning sequences into 60-frame chunks. We treat this high-fidelity, 60-view reconstruction as the pseudo-ground-truth depth for each view, subsequent to which we apply the same alignment and PCC computation workflow used for the Blender scenes.
\end{itemize}
\paragraph{Error Propagation and the Oracle PCC Baseline} It is important to note that because our evaluation pipeline relies on DA3-estimated depths, any inherent reconstruction errors from the model will systematically affect the final PCC values. This limitation is visible even when validating on the original dataset pairs (where the target views are the actual ground-truth frames); due to estimation noise in DA3, the metric fails to achieve a perfect error score of $0$ (see \Cref{app:fig:projection_consistency_dl3dv} and \Cref{app:fig:projection_consistency_re10k}).To decouple our model’s generation quality from the systematic errors of the reconstruction tool, we establish and report an Oracle PCC baseline across all experiments (\Cref{tab:blender_transposed_benchmarks_with_std} and \Cref{tab:full_comprehensive_landscape_benchmarks_no_std}). This oracle score represents the practical lower bound of the metric under DA3's geometric constraints, providing a realistic benchmark for perfect generation.

%% file: appendix/PCC_materials/pcc_pseudocode.tex
\begin{algorithm}[t]
\caption{Projection Consistency Check (PCC)}\label{alg:app:pcc}
\label{alg:pcc}
\begin{algorithmic}[1]
\Require source/target depth $D_s, D_t$; camera poses $T_s, T_t$; intrinsics $K_s, K_t$
\Statex \textbf{Params:} dead zone $\tau$; max clamp $c$
\Ensure consistency score (lower is better; $0=$ perfectly consistent)
\vspace{2pt}
\Function{Reproject}{$D,\, T_{\text{from}},\, T_{\text{to}}$}
    \State $P \gets \textsc{Unproject}(D, T_{\text{from}})$
        \Comment{depth $\to$ 3D points in world}
    \State \Return z-buffer render of $P$ into the $T_{\text{to}}$ camera
\EndFunction
\vspace{3pt}
\LComment{Forward: reproject source into target, compare to target depth}
\State $\hat D_{s\to t},\, M_f \gets \textsc{Reproject}(D_s, T_s, T_t)$
\State $\Delta_f \gets (D_t - \hat D_{s\to t})$ \textbf{ on valid pixels } $M_f$
\vspace{2pt}
\LComment{Backward: reproject target into source, compare to source depth}
\State $\hat D_{t\to s},\, M_b \gets \textsc{Reproject}(D_t, T_t, T_s)$
\State $\Delta_b \gets (D_s - \hat D_{t\to s})$ \textbf{ on valid pixels } $M_b$
\vspace{3pt}
\LComment{Robust, thresholded penalty (ignore noise, saturate outliers)}
\State $\phi(\Delta) \gets \min\!\big(\log(1 + \max(\Delta - \tau,\, 0)),\; c\big)$
\State $\text{score} \gets
        \dfrac{\sum_{M_f}\phi(\Delta_f) + \sum_{M_b}\phi(\Delta_b)}
              {|M_f| + |M_b|}$
\State \Return $\text{score}$
\end{algorithmic}
\end{algorithm}

The pseudocode for the PCC algorithm is provided in ~\Cref{alg:app:pcc}.

%% file: appendix/PCC_materials/differentiable_pcc.tex
The PCC score used for \emph{evaluation} renders a
depth map from one view into the other with a hard $z$-buffer: every source
pixel is unprojected to $3$D, projected into the target camera, rounded to the
nearest integer pixel, and the closest point per target pixel is kept by
sorting the splatted depths and letting nearer points overwrite farther ones.
Writing $\mathrm{round}(\cdot)$ for the pixel snapping and $\arg\!\operatorname{sort}$
for the depth ordering, the rendered target depth is
\begin{equation}
D_t^{\mathrm{zbuf}}(p) \;=\;
  \min_{\,q\,:\,\mathrm{round}(\pi_{s\to t}(q))\,=\,p}\; z_{s\to t}(q),
\label{eq:zbuffer}
\end{equation}
where $\pi_{s\to t}$ is the source-to-target reprojection and $z_{s\to t}$ the
transformed depth. The three operations that make Eq.~\eqref{eq:zbuffer} an
honest occlusion test --- the $\mathrm{round}(\cdot)$ pixel snapping, the
$\arg\!\operatorname{sort}$ depth ordering, and the scatter-assignment (a
per-pixel $\min$ over a data-dependent index set) --- are all piecewise-constant
in the depth, so $\partial D_t^{\mathrm{zbuf}} / \partial D = 0$ almost
everywhere. The metric is therefore non-differentiable and cannot be used as a
training objective on generated depth.

\paragraph{Inverse rendering via differentiable inverse warping.}
To obtain a trainable surrogate we replace the forward $z$-buffer
\emph{splat} (which scatters points \emph{to} the target grid at rounded
coordinates) with an \emph{inverse warp} that, for every target pixel,
analytically computes where it lands in the other view and \emph{gathers} the
other view's depth from those sub-pixel coordinates by bilinear sampling. Let
$\mathbf{P}_s, \mathbf{P}_t \in SE(3)$ denote the OpenGL camera-to-world poses
($X$ right, $Y$ up, camera looking down $-Z$) and $K_v=(f_x^v, f_y^v, c_x^v,
c_y^v)$ the pinhole intrinsics of view $v$. For a pixel $p=(u,v)$ with depth
$d$ we unproject to the source camera frame,
\begin{equation}
\mathbf{x}_c \;=\;
  \Bigl(\tfrac{(u-c_x)\,d}{f_x},\; \tfrac{(c_y-v)\,d}{f_y},\; -d\Bigr)^{\!\top},
\label{eq:unproject}
\end{equation}
transform it into the other camera with the single relative pose
$\mathbf{T}_{f\to t}=\mathbf{P}_t^{-1}\mathbf{P}_f$, giving
$\mathbf{x}'=(X',Y',Z')^{\!\top}=\mathbf{T}_{f\to t}\,\mathbf{x}_c$ and a
reprojected depth $z'=-Z'$, and project back down,
\begin{equation}
u' = f_x'\,\frac{X'}{z'} + c_x', \qquad
v' = f_y'\,\frac{Y'}{z'} + c_y', \qquad
\mathbf{g}(p) = \Bigl(\tfrac{2u'}{W}-1,\; \tfrac{2v'}{H}-1\Bigr),
\label{eq:project}
\end{equation}
where $\mathbf{g}(p)\in[-1,1]^2$ is the normalized sampling location
(\texttt{align\_corners=False}). A pixel is \emph{valid} when its reprojection
is finite, in front of the camera ($z'>\epsilon$), and lands inside the image.
Given a depth map $D$ and the grid $\mathbf{g}$, the inverse warp is the
bilinear sampling operator $\mathcal{S}(D;\mathbf{g})$. Unlike
Eq.~\eqref{eq:zbuffer}, $\mathcal{S}$ is differentiable both with respect to the
sampled \emph{values} and to the sampling \emph{coordinates}
$\mathbf{g}$, so gradients propagate back into the predicted depth through the
full projection chain in Eqs.~\eqref{eq:unproject}--\eqref{eq:project}.

\paragraph{Bidirectional gradient routing.}
Let $\hat D_t$ be the generated (target-view) depth we optimize and $D_s$ the
ground-truth source depth. We accumulate two terms, each of which places the
gradient on $\hat D_t$ but routes it differently, so that the predicted surface
is penalized whenever it pokes \emph{in front of} the reprojected reference.

\emph{(i) Forward, source $\to$ target.} We reproject the GT source depth into
the target view. Because the coordinates are driven by GT data, the grid
$\mathbf{g}_{s\to t}$ and the transformed reference depth $z_t^{\mathrm{ref}}$
are computed under $\operatorname{stop\text{-}grad}$; the gradient enters only
through the \emph{gathered prediction}
\begin{equation}
\hat D_t^{\rightarrow}(p) \;=\; \mathcal{S}\bigl(\hat D_t;\,\mathbf{g}_{s\to t}(p)\bigr),
\qquad
\Delta_{\mathrm{fwd}}(p) \;=\; \hat D_t^{\rightarrow}(p) - z_t^{\mathrm{ref}}(p).
\label{eq:fwd}
\end{equation}

\emph{(ii) Backward, target $\to$ source.} We reproject the \emph{predicted}
target depth into the source view. Here both the grid $\mathbf{g}_{t\to s}$ and
the transformed depth $z_s^{\mathrm{pred}}$ depend on $\hat D_t$ (gradient flows
through the projection), while the GT source depth is gathered as a detached
reference,
\begin{equation}
D_s^{\leftarrow}(p) \;=\; \mathcal{S}\bigl(D_s;\,\mathbf{g}_{t\to s}(p)\bigr),
\qquad
\Delta_{\mathrm{bwd}}(p) \;=\; D_s^{\leftarrow}(p) - z_s^{\mathrm{pred}}(p).
\label{eq:bwd}
\end{equation}

\paragraph{Signed penalty.}
Both residuals are passed through the same one-sided, saturated
$\log$ penalty as the evaluation metric. With a scale $s$ that maps native
depth units to metres, a dead-zone $\delta$ (optionally depth-proportional,
$\delta=\delta_{\mathrm{rel}}\,|D_{\mathrm{ref}}|\,s$), and an upper clamp
$\tau$,
\begin{equation}
\ell(\Delta) \;=\;
  \min\!\Bigl(\log\!\bigl(1 + \max(s\,\Delta - \delta,\,0)\bigr),\; \tau\Bigr).
\label{eq:penalty}
\end{equation}
Only the positive part beyond $\delta$ is penalized (the predicted surface in
front of the reference); the scaling acts on the \emph{difference} alone, so
the geometry in Eqs.~\eqref{eq:unproject}--\eqref{eq:project} stays in native
units. Summing over the valid pixels $\mathcal{V}$ of both directions and
normalizing by their count gives the per-sample loss
\begin{equation}
\mathcal{L}_{\mathrm{PCC}} \;=\;
  \frac{\displaystyle\sum_{p\in\mathcal{V}_{\mathrm{fwd}}}\!\ell\bigl(\Delta_{\mathrm{fwd}}(p)\bigr)
        + \sum_{p\in\mathcal{V}_{\mathrm{bwd}}}\!\ell\bigl(\Delta_{\mathrm{bwd}}(p)\bigr)}
       {\lvert\mathcal{V}_{\mathrm{fwd}}\rvert + \lvert\mathcal{V}_{\mathrm{bwd}}\rvert}.
\label{eq:pcc_loss}
\end{equation}
An optional per-pixel confidence map $w(p)\in[0,1]$, sampled at
$\mathbf{g}_{s\to t}$ for the forward term and used natively for the backward
term, multiplies each violation so that low-confidence regions contribute less.

\paragraph{Remark (surrogate gap).}
Because the inverse warp gathers depth by bilinear interpolation without a
$z$-test, it is occlusion-na\"ive and bleeds across depth discontinuities.
$\mathcal{L}_{\mathrm{PCC}}$ is therefore a differentiable \emph{surrogate} of
Eq.~\eqref{eq:zbuffer} and is not expected to match the metric numerically; we
optimize $\mathcal{L}_{\mathrm{PCC}}$ during training but always report the hard
$z$-buffer PCC at evaluation time.

%% file: appendix/PCC_materials/multiview.tex
\subsection{Multi-View PCC}
\label{sec:multiview-pcc}

\begin{figure}[t]
  \centering
  \includegraphics[width=0.95\linewidth]{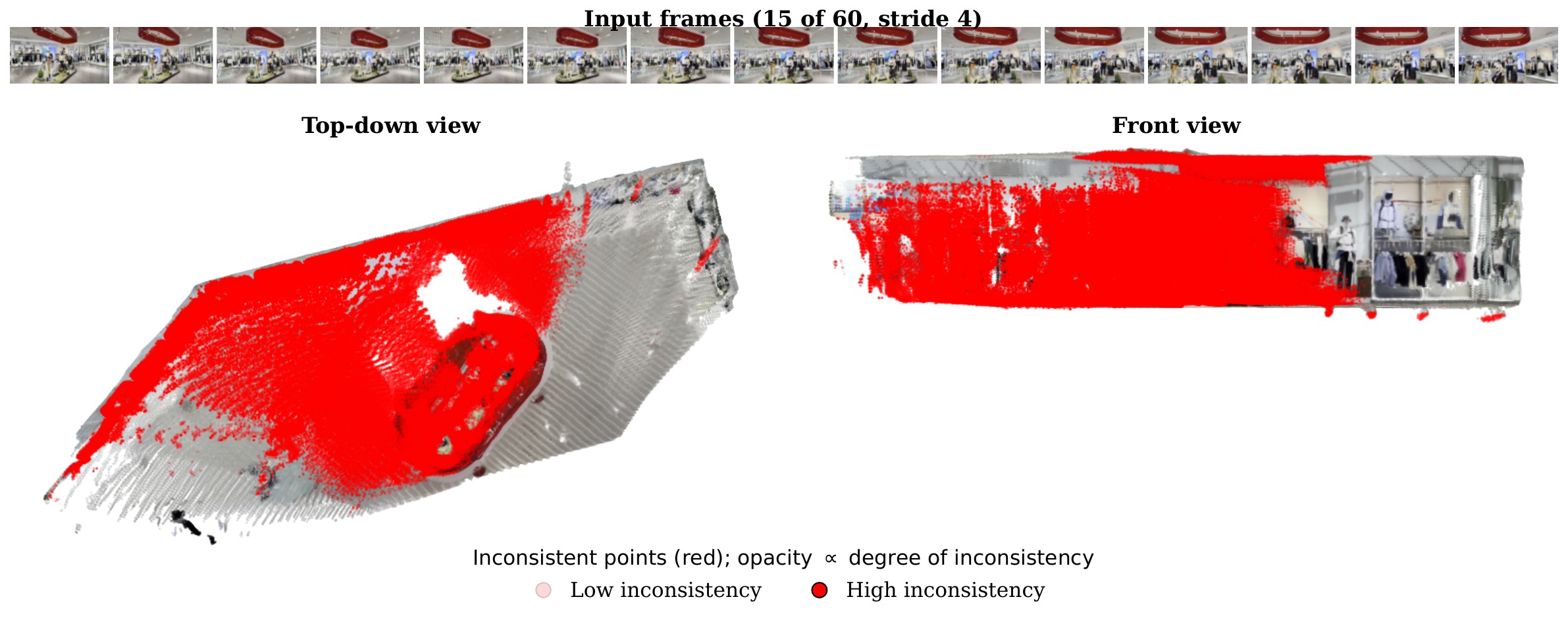}
  \caption{\textbf{Multi-view geometric inconsistency of a 60-frame accumulated point cloud (DL3DV).} \textbf{Top:} 15 of the 60 input frames (stride 4) used to build the cloud. \textbf{Bottom:} the accumulated point cloud, unprojected from the per-frame DA3 depth and DA3-estimated camera poses and shown from two viewpoints (top-down and front, in the reference-camera frame). Each frame's depth is scored by reprojecting the global cloud into that view (z-buffer) and measuring the relative depth violation $\mathrm{rel} = (D - \hat{D})/\hat{D}$; points with $\mathrm{rel} > \tau$ ($\tau{=}0.15$) are marked \textcolor{red}{red}, with \textbf{opacity proportional to the degree of inconsistency} $\max(\mathrm{rel}-\tau,\,0)$ (faint = borderline, solid = strongly inconsistent). Geometrically consistent points keep their RGB color.}
\label{app:fig:accum_inconsistency_dl3dv}
\end{figure}
\begin{figure}[h]
  \centering
  \includegraphics[width=0.95\linewidth]{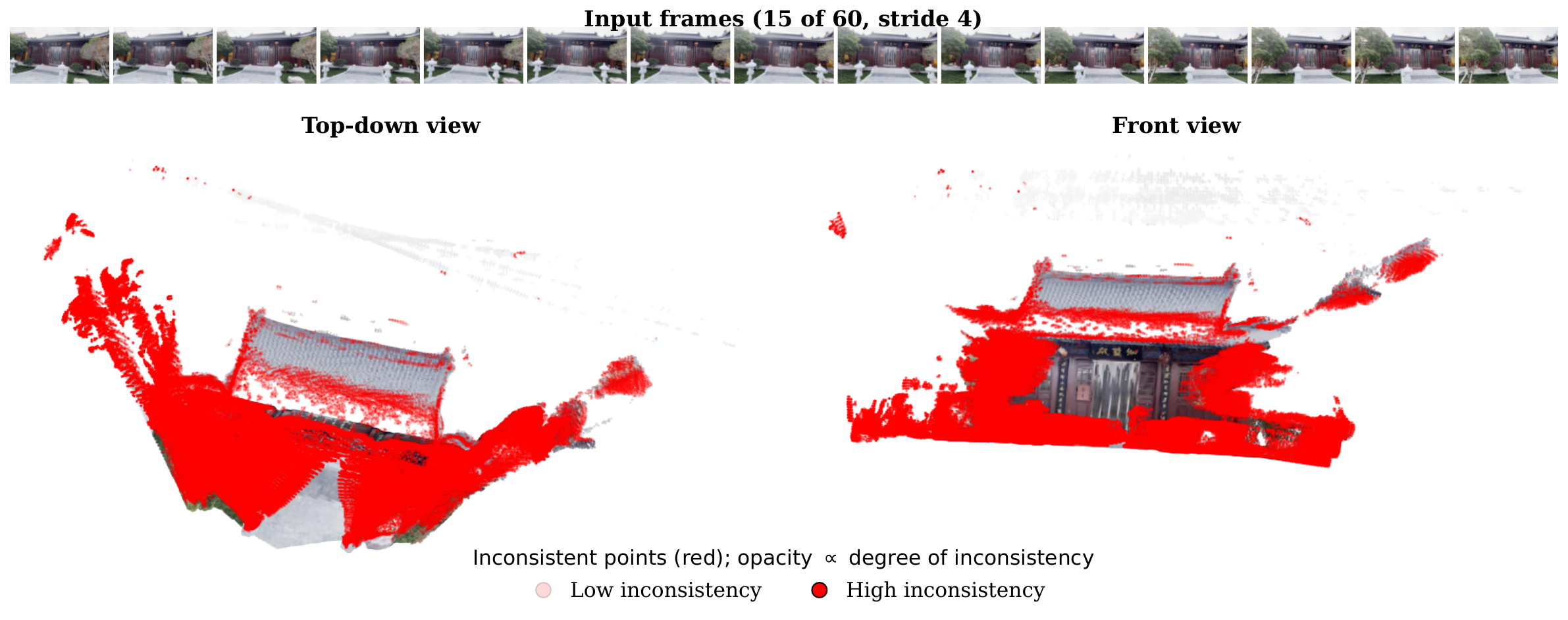}
  \caption{\textbf{Multi-view geometric inconsistency of a 60-frame accumulated point cloud (DL3DV).}}
\label{app:fig:accum_inconsistency_dl3dv_1}
\end{figure}
\begin{figure}[h]
  \centering
  \includegraphics[width=0.95\linewidth]{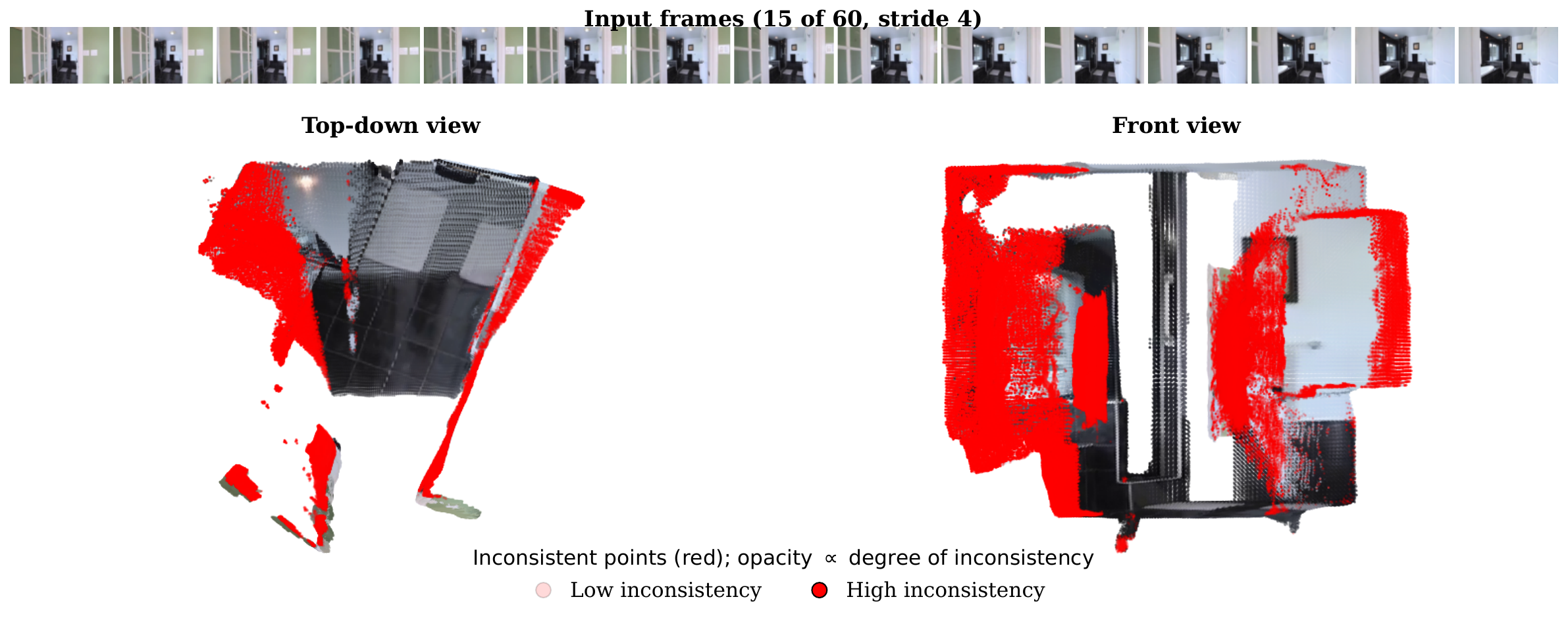}
  \caption{\textbf{Multi-view geometric inconsistency of a 60-frame accumulated point cloud (DL3DV).}}
\label{app:fig:accum_inconsistency_re10k}
\end{figure}

\paragraph{Setup.}
Consider a scene with $V = N + M$ posed views, comprising $N$ source views and $M$ generated views. For each view $v_i$, where $i \in \{1,\dots,V\}$, we are given an RGB image $I_{v_i}$, a depth map $D_{v_i}\!:\Omega\!\to\!\mathbb{R}_{\ge 0}$ (with image domain $\Omega$), and a camera-to-world pose $C_{v_i}\!\in\!\mathrm{SE}(3)$.

\paragraph{Global reprojection.}
For any given view $v_i$, we construct a global point cloud $P_{\text{global}}$ aggregated from all the views $v_j$:
\begin{equation}
    P_{\text{global}} \;=\; \bigcup_{i} P_{v_i},
    \qquad \text{where} \quad
    P_{v_i} \;=\; \bigl\{\, C_{v_i}\,\pi^{-1}\!\bigl(\mathbf{u},\,D_{v_i}(\mathbf{u})\bigr) \;\big|\; \mathbf{u}\in\Omega \,\bigr\},
\end{equation}
and $\pi^{-1}(\mathbf{u},d)$ unprojects pixel $\mathbf{u}\in\Omega$ at depth $d$ into the camera coordinate frame. We then project $P_{\text{global}}$ into camera $v_i$ and rasterize it using a $z$-buffer (where the nearest point per pixel wins). This yields the \emph{reprojected depth} of the scene from the perspective of $v_i$:
\begin{equation}
    \widehat{D}_{v_i}(\mathbf{u})
    \;=\;
    \min_{\substack{\mathbf{p}\in P_{\text{global}} \\ \pi(C_{v_i}^{-1}\mathbf{p}) = \mathbf{u}}}
    \bigl[\,C_{v_i}^{-1}\mathbf{p}\,\bigr]_{z}.
\end{equation}

\paragraph{Per-pixel consistency criterion.}
For opaque scenes, the only physically admissible relationship between the observed depth $D_{v_i}$ and the reprojected depth $\widehat{D}_{v_i}$ is:
\begin{equation}
    D_{v_i}(\mathbf{u}) \;\le\; \widehat{D}_{v_i}(\mathbf{u}) + \delta,
    \qquad \forall\, \mathbf{u}\in M_i,
    \label{eq:mvpcc-constraint}
\end{equation}
where $\delta\ge 0$ is a tolerance for depth noise and rasterization quantization.

\paragraph{Why the test suffices.}
Although Eq.~\eqref{eq:mvpcc-constraint} is one-sided, applying it across \emph{all} views systematically detects both primary error modes in novel view generation. Let $v_i$ be a generated view containing a hallucinated surface $H \subset \mathbb{R}^3$ at pixel $\mathbf{u}_i \in \Omega$.

\begin{enumerate}
    \item[\textbf{(a)}] \textbf{Occlusion Violation} (\emph{$H$ lies behind the rest of the scene}).
    At view $v_i$, the reprojected depth $\widehat{D}_{v_i}(\mathbf{u}_i)$ records the closer, real surface seen by other views, whereas the predicted depth is $D_{v_i}(\mathbf{u}_i) = \mathrm{depth}(H)$. Because $H$ is behind the real surface, $D_{v_i}(\mathbf{u}_i) > \widehat{D}_{v_i}(\mathbf{u}_i)$, triggering a violation directly at view $i$.

    \item[\textbf{(b)}] \textbf{Free-Space Violation} (\emph{$H$ lies in front of the rest of the scene; a ``floater''}).
    At view $v_i$, $D_{v_i}(\mathbf{u}_i) = \mathrm{depth}(H) \le \widehat{D}_{v_i}(\mathbf{u}_i)$, so no violation is raised at $i$ (from $v_i$'s own perspective, the scene appears self-consistent). However, for any \emph{other} view $v_j$ whose ray passes through $H$ at pixel $\mathbf{u}_j$ first before any other points. The $z$-buffer pulls the reprojected depth $\widehat{D}_{v_j}(\mathbf{u}_j)$ forward to $\mathrm{depth}(H)$, while the real-scene observation $D_{v_j}(\mathbf{u}_j)$ corresponds to the true, farther surface. Thus, $D_{v_j}(\mathbf{u}_j) > \widehat{D}_{v_j}(\mathbf{u}_j)$, and the violation is successfully caught at view $j$.
\end{enumerate}

\paragraph{Multiview PCC loss.}
We turn the one-sided criterion of Eq.~\eqref{eq:mvpcc-constraint} into a
score by penalizing only its violations. For each view $v_i$ define the
per-pixel free-space residual
\begin{equation}
    r_{v_i}(\mathbf{u})
    \;=\;
    \max\!\bigl(0,\; D_{v_i}(\mathbf{u}) - \widehat{D}_{v_i}(\mathbf{u}) - \delta \bigr),
\end{equation}
which is positive exactly when the observed depth lies \emph{behind} the
$z$-buffered scene depth by more than the tolerance $\delta$. To bound the
influence of gross outliers (e.g.\ depth-discontinuity fringes), we apply a
saturating penalty
\begin{equation}
    \rho(r) \;=\; \min\!\bigl(\log(1 + r),\; \tau \bigr),
\end{equation}
with cap $\tau$. The multi-view PCC loss is the mean penalty over all covered
pixels, aggregated across every view:
\begin{equation}
    \mathcal{L}_{\mathrm{PCC}}
    \;=\;
    \frac{\displaystyle\sum_{i=1}^{V}\sum_{\mathbf{u}\in M_i} \rho\bigl(r_{v_i}(\mathbf{u})\bigr)}
         {\displaystyle\sum_{i=1}^{V} \lvert M_i \rvert},
    \qquad
    M_i \;=\; \bigl\{\, \mathbf{u}\in\Omega \;\big|\; \widehat{D}_{v_i}(\mathbf{u}) < \infty \,\bigr\},
\end{equation}
where $M_i$ is the set of pixels in view $v_i$ that receive at least one
reprojected point under the $z$-buffer. Lower $\mathcal{L}_{\mathrm{PCC}}$
indicates greater multi-view consistency.
\newpage

%% file: appendix/baseline_models.tex
We compare our ESC model against a representative set of recent novel-view-synthesis (NVS) baselines spanning the main generative paradigms. \textbf{OVIE}~\citep{rahary2026ovie} and \textbf{LagerNVS}~\citep{szymanowicz2026lagernvs} are feed-forward regressors that predict a target view in a single forward pass (OVIE additionally performs a $16$-scale search), prioritizing speed over photorealism. \textbf{VIVID}~\citep{elata2024vivid} is a latent diffusion model with classifier-free guidance, while \textbf{PhotoNVS}~\citep{yu2023photonvs} is a pixel-space DDPM requiring many denoising steps. \textbf{GLD}~\citep{jang2026gld} adopts a multi-stage cascade diffusion design, and \textbf{Gen3C}~\citep{ren2025gen3c} casts NVS as conditional video diffusion at high resolution. For each baseline we use the authors' released checkpoints and recommended inference settings. We evaluate two variants of our own model: \textbf{ESC-Blender}, specialized on the Blender scenes, and \textbf{ESC-Mixed}, trained on the mixture of RealEstate10K and DL3DV training sets and evaluated on Tanks-and-Temples, RealEstate10K, and DL3DV.

\subsection{Inference Efficiency}
\label{sec:inference_efficiency}
\begin{table}[h]
  \centering
  
  \setlength{\aboverulesep}{0pt}
  \setlength{\belowrulesep}{0pt}
  \renewcommand{\arraystretch}{1.3} 
  
  \begin{tabular}{llccc}
    \toprule
    \textbf{Model} & \textbf{Paradigm} & \textbf{NFE} & \textbf{Native Res.} & \textbf{Per-view Time} \\
    \midrule
    OVIE & feed-forward + 16-scale search & $\sim$1 ($\times$16) & $256$ & $0.1$--$0.3$\,s \\
    LagerNVS & feed-forward & $\sim$1 & $512$ & $0.2$--$0.5$\,s \\
    \midrule
    \rowcolor{green!12} ESC-Blender (rank 64) & diffusion & $20$ & $384$ & $\sim$3\,s \\
    \rowcolor{green!12} ESC-Mixed (rank 128) & diffusion & $20$ & $384$ & $\sim$3\,s \\
    \midrule
    VIVID & diffusion + CFG ($g{=}1.5$) & $32$ ($\sim$64) & $384$ & $3$--$4$\,s \\
    GLD & cascade diffusion (L1/L2/L3) & multi-stage & --- & $5$--$15$\,s \\
    % Gen3C & video diffusion & $50$ & $704{\times}1280$ & $30$--$60$\,s \\
    PhotoNVS & DDPM & $2000$ & $384$ & $60$--$120$\,s \\
    \bottomrule
  \end{tabular}
  \caption{Comparison of inference paradigms, total number of function evaluations (NFE), native resolution, and per-view latency. Per-view times are normalized to an H200 GPU and represent startup-corrected estimates. For our ESC models, we observe similar inference latency across the evaluated LoRA ranks (rank 64 for Blender; rank 128 for Mixed).}
  \label{tab:runtime}
\end{table}

Table~\ref{tab:runtime} compares the inference paradigm, total number of function evaluations (NFE), native resolution, and per-view latency across all evaluated methods. Feed-forward approaches (OVIE, LagerNVS) offer the lowest latency, generating novel views in well under one second. Conversely, diffusion-based methods prioritize fidelity, with computational cost largely dictated by the number of function evaluations; consequently, PhotoNVS ($2000$ steps) and Gen3C ($50$ steps at $704{\times}1280$) are one to two orders of magnitude slower. Our ESC models strike a favorable balance, synthesizing a $384{\times}384$ view in approximately three seconds using only $20$ NFE. This approach achieves latency comparable to VIVID while requiring fewer function evaluations and remains significantly more efficient than cascade and video-diffusion baselines. Notably, this inference latency remains consistent for both our Blender-specialized (LoRA rank $64$) and mixed-data (LoRA rank $128$) variants.

%% file: appendix/results.tex
% ---------------------------------------------------------------------------
% Qualitative ESC variants on the synthetic Blender benchmark.
% Columns: Source, GT, ESC-Shadow, ESC-PCC-BoN, ESC-PCC-FT.
% ---------------------------------------------------------------------------
\subsection{Compute Time}
\input{appendix/compute_time}
\begin{table*}[p]
\centering
\setlength{\tabcolsep}{1.5pt}
\renewcommand{\arraystretch}{0.5}
\begin{tabularx}{\textwidth}{Y Y | Y Y Y}
\demohd{\textcolor{legendinput}{Source}} & \demohd{\textcolor{legendtarget}{GT}} & \demohd{\textcolor{legendours}{\textbf{ESC-Shadow}}} & \demohd{\textcolor{legendours}{\textbf{ESC-PCC-BoN}}} & \demohd{\textcolor{legendours}{\textbf{ESC-PCC-FT}}} \\
\hline
% Scene 1: sample_001853
\rescellsrc{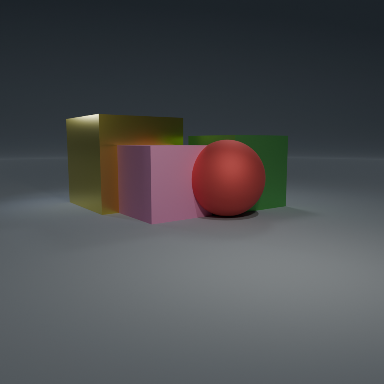} &
\rescellgt{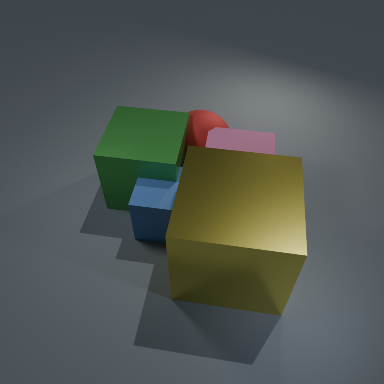} &
\rescellours{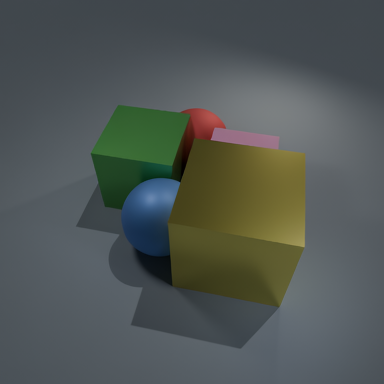} &
\rescellours{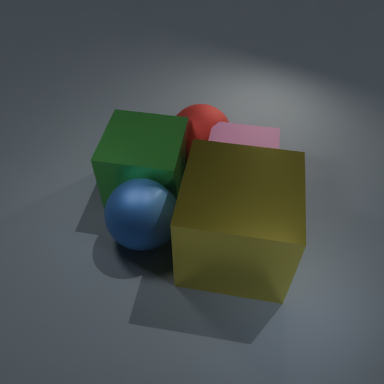} &
\rescellours{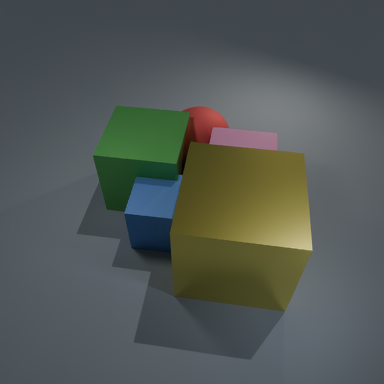} \\
% Scene 2: sample_005888
\rescellsrc{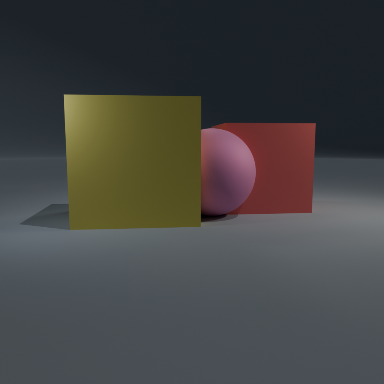} &
\rescellgt{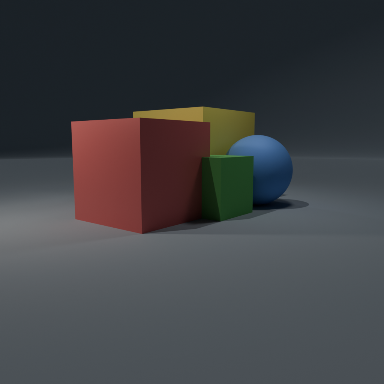} &
\rescellours{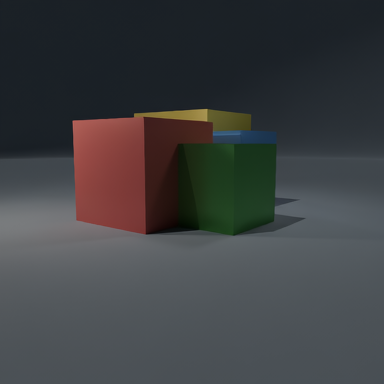} &
\rescellours{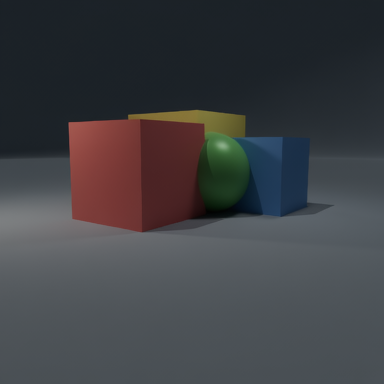} &
\rescellours{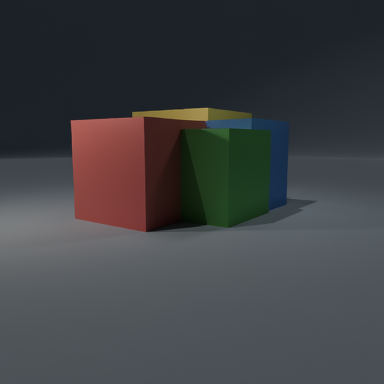} \\
% Scene 3: sample_013869
\rescellsrc{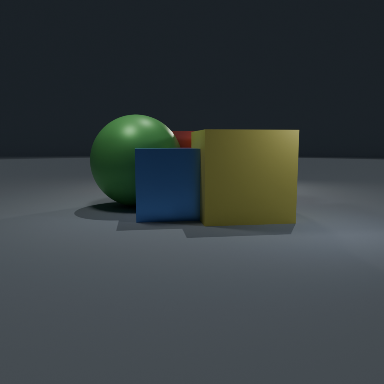} &
\rescellgt{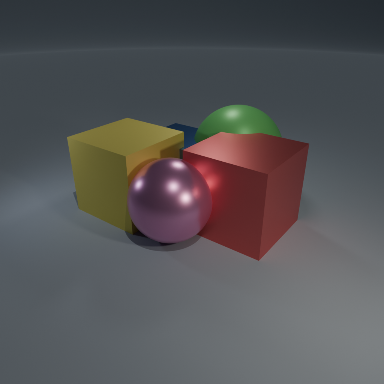} &
\rescellours{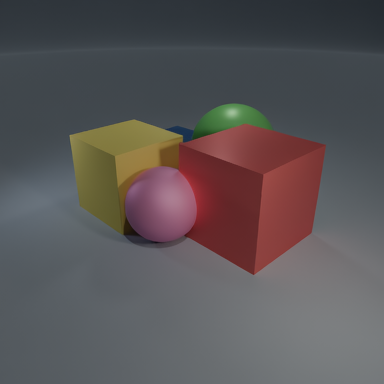} &
\rescellours{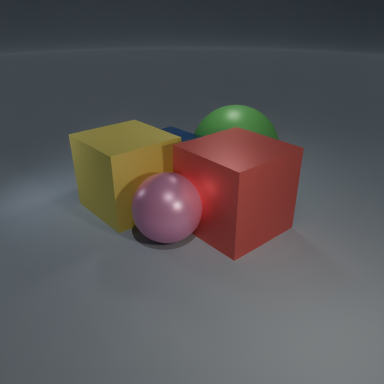} &
\rescellours{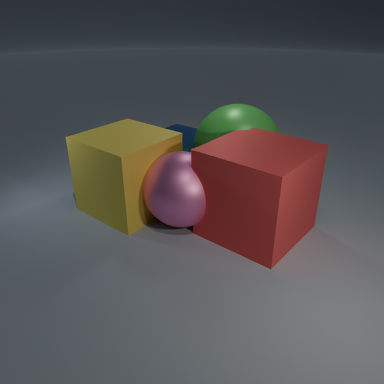} \\
% Scene 4: sample_018919
\rescellsrc{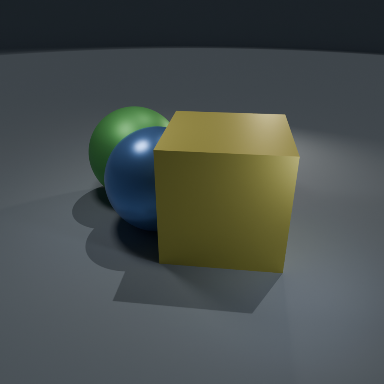} &
\rescellgt{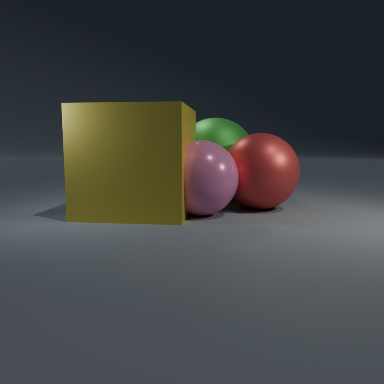} &
\rescellours{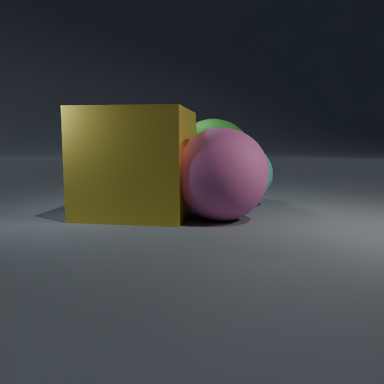} &
\rescellours{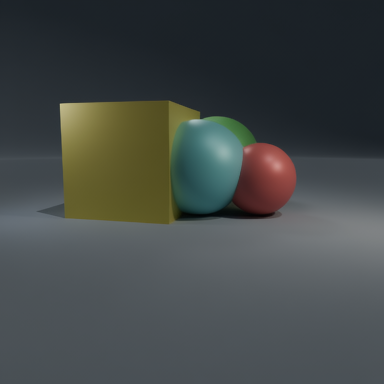} &
\rescellours{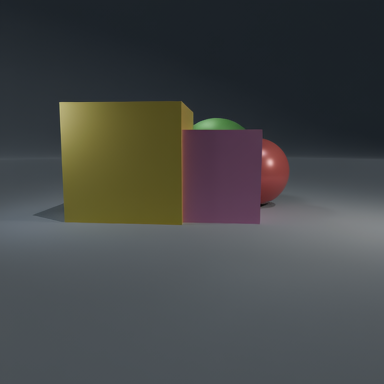} \\
\end{tabularx}
\caption{\textbf{Qualitative comparison of ESC variants on the synthetic Blender benchmark.} Each row shows the \textcolor{legendinput}{source view (blue border)}, the \textcolor{legendtarget}{ground-truth target (green border)}, and three \textcolor{legendours}{ESC variants (orange border)}: ESC-Shadow, ESC-PCC-BoN, and ESC-PCC-FT. PCC-aware selection and fine-tuning yield novel views that most faithfully reproduce the scene geometry.}
\label{tab:blender_qualitative}
\end{table*}

\clearpage
\newpage

\begin{sidewaystable*}[p]
\centering
\small

\vspace{10pt}
\begin{tabular}{l cc ccc c cccc}
\toprule
& \multicolumn{2}{c}{\textbf{Synthesis Quality}} & \multicolumn{3}{c}{\textbf{Perceptual Quality}} & \textbf{Geometry} & \multicolumn{4}{c}{\textbf{Pose \& View Prompt Consistency}} \\
\cmidrule(r){2-3} \cmidrule(lr){4-6} \cmidrule(lr){7-7} \cmidrule(l){8-11}
\textbf{Model} & PSNR $\uparrow$ & SSIM $\uparrow$ & LPIPS $\downarrow$ & DINO-Sim $\uparrow$ & DreamSim $\downarrow$ & PCC $\downarrow$ & AUC@15 $\uparrow$ & AUC@30 $\uparrow$ & RPE-R $\downarrow$ & RPE-T $\downarrow$ \\
\midrule

\multicolumn{11}{l}{\textbf{DL3DV Dataset}} \\
\midrule
PhotoNVS & $12.38$ & \third{$0.3465$} & $0.5743$ & $0.5647$ & $0.3876$ & $0.2628$ & $0.0148$ & $0.0558$ & $70.79$ & $47.18$ \\
VIVID & $10.88$ & $0.3001$ & $0.6407$ & $0.6909$ & $0.2953$ & \third{$0.2475$} & $0.0551$ & $0.1949$ & $22.61$ & $37.21$ \\
OVIE & $11.58$ & $0.3365$ & $0.7099$ & $0.2755$ & $0.6480$ & $0.3122$ & $0.0084$ & $0.0403$ & $54.14$ & $49.18$ \\
GLD & $12.58$ & \second{$0.3598$} & $0.5665$ & $0.7019$ & $0.2549$ & $0.2487$ & \second{$0.1134$} & \second{$0.2894$} & \third{$16.03$} & \second{$31.58$} \\
LagerNVS & \first{$16.14$} & \first{$0.5102$} & \third{$0.4714$} & $0.6734$ & $0.3231$ & $0.2487$ & \first{$0.1231$} & \first{$0.2932$} & $16.74$ & \first{$30.70$} \\
\modelsep
ESC-shadow & $12.16$ & $0.3082$ & $0.4952$ & \third{$0.8035$} & \third{$0.1676$} & $0.2533$ & $0.0492$ & $0.2012$ & $19.62$ & $32.91$ \\
ESC (\textit{Ours}) & \third{$12.69$} & $0.3343$ & \second{$0.4448$} & \second{$0.8366$} & \second{$0.1420$} & \second{$0.2448$} & $0.0886$ & $0.2575$ & \second{$12.97$} & $33.28$ \\
ESC-PCC-BoN (\textit{Ours}) & \second{$12.76$} & $0.3347$ & \first{$0.4340$} & \first{$0.8386$} & \first{$0.1402$} & \first{$0.2257$} & \third{$0.0922$} & \third{$0.2643$} & \first{$12.91$} & \third{$32.52$} \\
\modelsep
Oracle (\textit{GT-vs-DA3}) & $\cdot$ & $\cdot$ & $\cdot$ & $\cdot$ & $\cdot$ & $0.1199$ & $0.4090$ & $0.6631$ & $5.14$ & $9.87$ \\

\midrule
\multicolumn{11}{l}{\textbf{RE10K Dataset}} \\
\midrule
PhotoNVS & \third{$17.51$} & \third{$0.5886$} & $0.3660$ & $0.7669$ & $0.2176$ & $0.0899$ & $0.4734$ & $0.6234$ & $3.96$ & $16.4$ \\
VIVID & $16.46$ & $0.5514$ & $0.3741$ & $0.8841$ & $0.1088$ & $0.0879$ & $0.3997$ & $0.6119$ & $1.51$ & $14.61$ \\
OVIE & $13.80$ & $0.4638$ & $0.5009$ & $0.7168$ & $0.7881$ & $0.1344$ & $0.0243$ & $0.0923$ & $12.93$ & $47.12$ \\
GLD & $15.77$ & $0.5527$ & $0.4413$ & $0.8128$ & $0.1394$ & $0.0898$ & \first{$0.7273$} & \first{$0.8422$} & \second{$0.95$} & \first{$5.86$} \\
LagerNVS & \first{$18.46$} & \first{$0.6350$} & $0.3239$ & $0.8838$ & $0.1256$ & \third{$0.0783$} & \second{$0.6800$} & \second{$0.7980$} & \first{$0.56$} & \second{$8.02$} \\
\modelsep
ESC-shadow & $15.74$ & $0.5180$ & \third{$0.3166$} & \third{$0.8982$} & \third{$0.0780$} & $0.0870$ & \third{$0.6142$} & \third{$0.7679$} & $1.61$ & \third{$8.48$} \\
ESC (\textit{Ours}) & $16.61$ & $0.5464$ & \second{$0.2613$} & \second{$0.9157$} & \second{$0.0669$} & \second{$0.0773$} & $0.5661$ & $0.7346$ & $1.09$ & $9.97$ \\
ESC-PCC-BoN (\textit{Ours}) & \second{$17.73$} & \second{$0.5912$} & \first{$0.2150$} & \first{$0.9222$} & \first{$0.0655$} & \first{$0.0656$} & $0.5787$ & $0.7429$ & \third{$0.98$} & $9.68$ \\
\modelsep
Oracle (\textit{GT-vs-DA3}) & $\cdot$ & $\cdot$ & $\cdot$ & $\cdot$ & $\cdot$ & $0.0539$ & $0.8098$ & $0.8862$ & $0.45$ & $4.62$ \\

\midrule
\multicolumn{11}{l}{\textbf{Tanks \& Temples Dataset}} \\
\midrule
PhotoNVS & $10.13$ & $0.2692$ & $0.7688$ & $0.2345$ & $0.5902$ & $0.4467$ & $0.0275$ & $0.0796$ & $49.70$ & $43.41$ \\
VIVID & $10.38$ & $0.2859$ & $0.7527$ & $0.3793$ & $0.5040$ & $0.4416$ & $0.0833$ & $0.2310$ & $33.89$ & $31.86$ \\
OVIE & $10.58$ & \second{$0.3093$} & $0.8219$ & $0.0640$ & $0.7881$ & $0.5102$ & $0.0052$ & $0.0229$ & $53.01$ & $53.35$ \\
GLD & $10.88$ & $0.2891$ & $0.6365$ & $0.5099$ & $0.4027$ & \third{$0.4184$} & \first{$0.4989$} & \first{$0.6208$} & \third{$15.92$} & \first{$15.33$} \\
LagerNVS & \first{$13.48$} & \first{$0.4669$} & $0.6420$ & $0.2738$ & $0.6630$ & $0.4901$ & $0.1937$ & $0.2736$ & $35.25$ & $33.84$ \\
\modelsep
ESC-shadow & $10.78$ & $0.2692$ & \third{$0.6336$} & \third{$0.6549$} & \third{$0.2928$} & $0.4291$ & $0.1694$ & $0.3837$ & $22.00$ & $20.08$ \\
ESC (\textit{Ours}) & \third{$11.27$} & $0.2882$ & \second{$0.5835$} & \second{$0.7242$} & \second{$0.2362$} & \second{$0.3826$} & \third{$0.3287$} & \second{$0.5501$} & \first{$14.62$} & \second{$15.64$} \\
ESC-PCC-BoN (\textit{Ours}) & \second{$11.42$} & \third{$0.2892$} & \first{$0.5596$} & \first{$0.7339$} & \first{$0.2232$} & \first{$0.2849$} & \second{$0.3329$} & \third{$0.5482$} & \second{$14.74$} & \third{$16.28$} \\
\modelsep
Oracle (\textit{GT-vs-DA3}) & $\cdot$ & $\cdot$ & $\cdot$ & $\cdot$ & $\cdot$ & $0.1899$ & $0.7700$ & $0.8357$ & $7.55$ & $6.33$ \\
\bottomrule
\end{tabular}
\caption{\textbf{Full Comprehensive Evaluation across View Synthesis Benchmarks.} We present the complete evaluation metrics across DL3DV, RE10K, and Tanks \& Temples datasets. Arrow ($\downarrow / \uparrow$) indicates the direction of better performance. The top-three entries within each dataset are highlighted using distinct progressive pastel red backgrounds, with the \first{1st} place values also emphasized in \textbf{bold} text. "Oracle" indicates the ground-truth label compared against the paired DA3 estimate on the source images.}
\label{tab:full_comprehensive_landscape_benchmarks_no_std}
\end{sidewaystable*}

\clearpage
\newpage

% ---------------------------------------------------------------------------
% Extended qualitative comparison (fourteen scenes x ten methods) for the
% appendix. Structure follows fig:qualitative_nvs in the main paper.
% ---------------------------------------------------------------------------
\newlength{\appdemoimg}
\setlength{\appdemoimg}{0.095\textwidth}
\newcommand{\cmproot}{figures/results/comparisons}
\newcommand{\appdemocell}[2]{\includegraphics[width=\appdemoimg]{\cmproot/#1/#2.png}}
\newcommand{\appdemocellsrc}[1]{\srcframe{\includegraphics[width=\appdemoimg]{\cmproot/#1/src.png}}}
\newcommand{\appdemocellgt}[1]{\gtframe{\includegraphics[width=\appdemoimg]{\cmproot/#1/gt.png}}}
\newcommand{\appdemocellours}[2]{\eacframe{\includegraphics[width=\appdemoimg]{\cmproot/#1/#2.png}}}
\newcommand{\appdemocellbase}[2]{\baseframe{\includegraphics[width=\appdemoimg]{\cmproot/#1/#2.png}}}
\newcommand{\appdemohd}[1]{\makebox[\appdemoimg][c]{\scriptsize #1}}
\newcommand{\appdemorow}[2]{%
  \rotatebox{90}{\makebox[\appdemoimg][c]{\scriptsize #2}}
  & \appdemocellsrc{#1}
  & \appdemocellbase{#1}{photonvs}
  & \appdemocellbase{#1}{vivid}
  & \appdemocellbase{#1}{ovie}
  & \appdemocellbase{#1}{gld}
  & \appdemocellbase{#1}{lagernvs}
  & \appdemocellours{#1}{eac_shadow}
  & \appdemocellours{#1}{eac_base}
  & \appdemocellours{#1}{eac_bon}
  & \appdemocellgt{#1} \\%
}

\begin{figure*}[p]
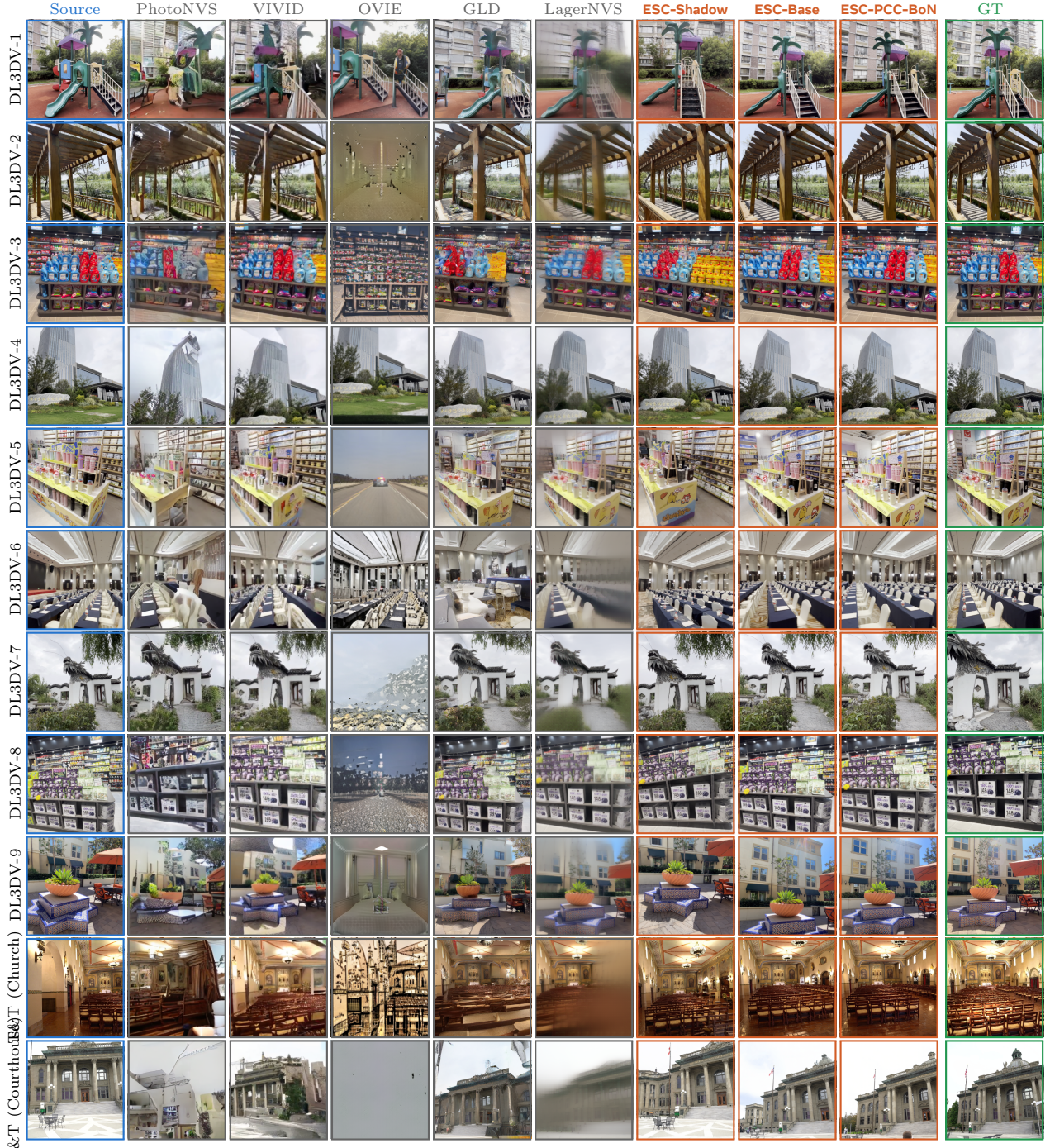

\centering
\renewcommand{\arraystretch}{0.45}
\begin{tabular}{@{}c@{}*{10}{@{\hspace{1.5pt}}c}@{}}
& \appdemohd{\textcolor{legendinput}{Source}} & \appdemohd{\textcolor{legendbaseline}{PhotoNVS}} & \appdemohd{\textcolor{legendbaseline}{VIVID}} & \appdemohd{\textcolor{legendbaseline}{OVIE}}
& \appdemohd{\textcolor{legendbaseline}{GLD}} & \appdemohd{\textcolor{legendbaseline}{LagerNVS}} & \appdemohd{\textcolor{legendours}{\textbf{ESC-Shadow}}}
& \appdemohd{\textcolor{legendours}{\textbf{ESC-Base}}} & \appdemohd{\textcolor{legendours}{\textbf{ESC-PCC-BoN}}} & \appdemohd{\textcolor{legendtarget}{GT}} \\[1pt]
\appdemorow{dl3dv_0238de14089be825251ca9a7c91c471a941bde76f78ae7b637ef306f30df544e_src-frame_00033_tgt-frame_00166_idx011460}{DL3DV-1}
\appdemorow{dl3dv_14452271416631599c6e7a2362b2c5330e08fbf0ec2adfea7930d84cba4e3d69_src-frame_00042_tgt-frame_00037_idx019400}{DL3DV-2}
\appdemorow{dl3dv_599b9dc15771ada7ce5eed6c2bc69c90977a2fa072a82f7121714f4a882f6ee5_src-frame_00030_tgt-frame_00035_idx013140}{DL3DV-3}
\appdemorow{dl3dv_924228bfa8983b5aa7921a609d5046036c96022f8458c037460386e74f3ffec6_src-frame_00213_tgt-frame_00015_idx017646}{DL3DV-4}
\appdemorow{dl3dv_9d746529972b9f3957bb5a4895984946638bd1945880934c448c2431523dd1d1_src-frame_00008_tgt-frame_00155_idx005419}{DL3DV-5}
\appdemorow{dl3dv_a126efe829a3872bb092e09b58ece5c87d464dd3c3ec5665b1b5fddb696d55ff_src-frame_00069_tgt-frame_00078_idx019861}{DL3DV-6}
\appdemorow{dl3dv_ae90ea23130506b51f0aa19e518d2cc920ebb8c9a2b89a2178c00d3f5959440b_src-frame_00332_tgt-frame_00006_idx020438}{DL3DV-7}
\appdemorow{dl3dv_f20fc26d6d46029dc7dac01435523d4d015b27ef2d53bc415007ee695edc8a17_src-frame_00028_tgt-frame_00204_idx026627}{DL3DV-8}
\appdemorow{dl3dv_fae2d0d5927ef4e8a06398ce9ba6dd52ee4fee78131bb9d2910ca5dbfbfa2362_src-frame_00016_tgt-frame_00155_idx028334}{DL3DV-9}
\appdemorow{tnt_church_src-000001_tgt-000006_idx001350}{T\&T (Church)}
\appdemorow{tnt_courthouse_src-000006_tgt-000031_idx002230}{T\&T (Courthouse)}
\end{tabular}
\caption{\textbf{Extended qualitative comparison on novel view synthesis (DL3DV and Tanks \& Temples).} For each scene we show the input source view (\textcolor{legendinput}{blue border}), the renderings produced by the baselines (\textcolor{legendbaseline}{gray border}: PhotoNVS, VIVID, OVIE, GLD, LagerNVS), our ESC variants (\textcolor{legendours}{orange border}: ESC-Shadow, ESC-Base, and the geometry-guided ESC-PCC-BoN (Best-of-N)), and the ground-truth target view (\textcolor{legendtarget}{green border}). Rows 1--9 are scenes from DL3DV, and the last two rows show the \emph{Church} and \emph{Courthouse} scenes from Tanks \& Temples. Across datasets, our variants reproduce scene geometry and fine detail more faithfully than the baselines, with ESC-PCC-BoN consistently best matching the ground-truth target.}
\label{fig:qualitative_nvs_dl3dv_tnt}
\end{figure*}

\clearpage
\newpage

% ---------------------------------------------------------------------------
% Extended qualitative comparison on RE10K (nine scenes x ten methods).
% Reuses the \appdemorow / \appdemocell / \appdemohd macros defined above.
% ---------------------------------------------------------------------------
\begin{figure*}[p]
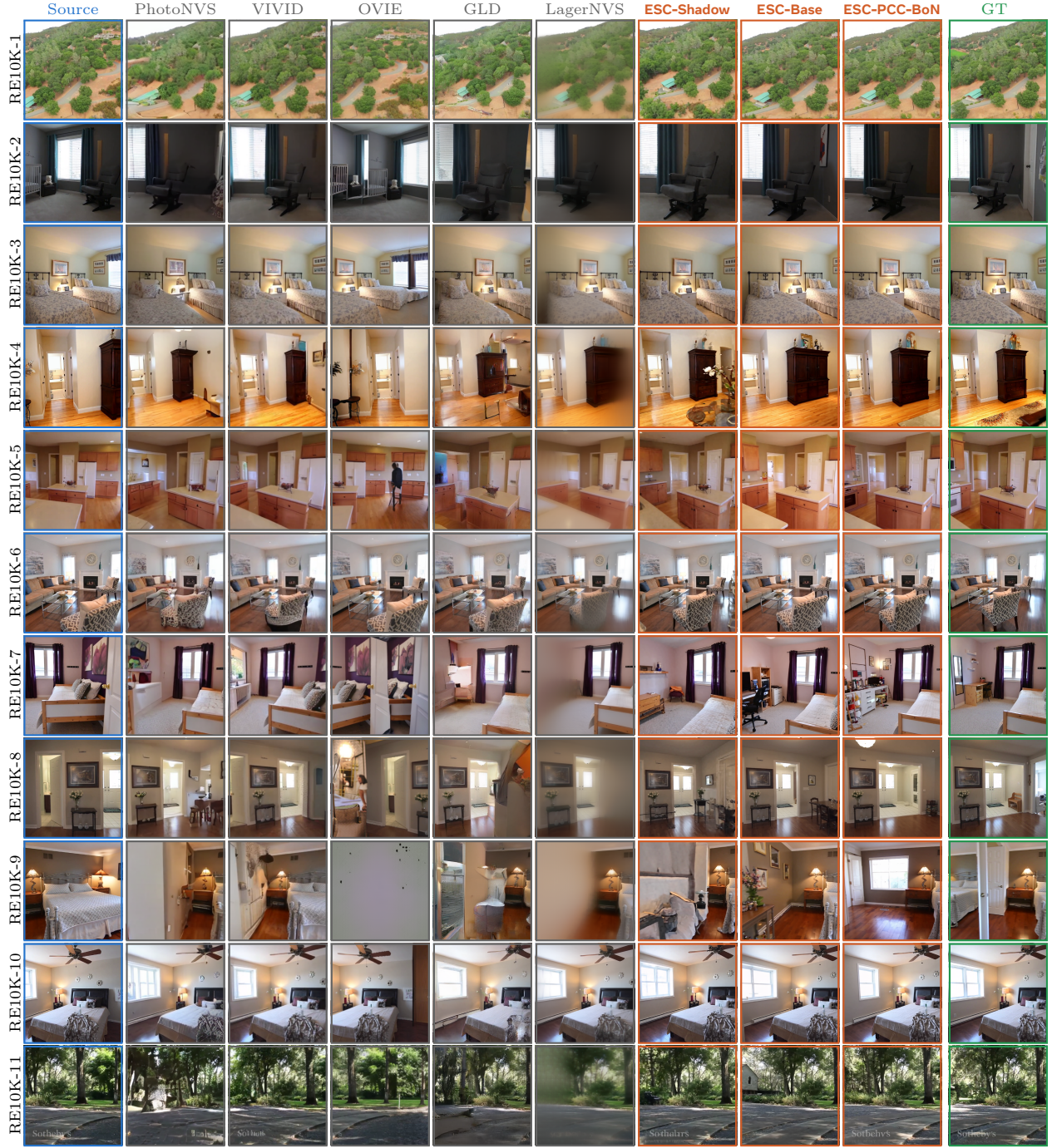

\centering
\renewcommand{\arraystretch}{0.45}
\begin{tabular}{@{}c@{}*{10}{@{\hspace{1.5pt}}c}@{}}
& \appdemohd{\textcolor{legendinput}{Source}} & \appdemohd{\textcolor{legendbaseline}{PhotoNVS}} & \appdemohd{\textcolor{legendbaseline}{VIVID}} & \appdemohd{\textcolor{legendbaseline}{OVIE}}
& \appdemohd{\textcolor{legendbaseline}{GLD}} & \appdemohd{\textcolor{legendbaseline}{LagerNVS}} & \appdemohd{\textcolor{legendours}{\textbf{ESC-Shadow}}}
& \appdemohd{\textcolor{legendours}{\textbf{ESC-Base}}} & \appdemohd{\textcolor{legendours}{\textbf{ESC-PCC-BoN}}} & \appdemohd{\textcolor{legendtarget}{GT}} \\[1pt]
\appdemorow{re10k_0b429a4733089487_src-frame_00040_tgt-frame_00090_idx001914}{RE10K-1}
\appdemorow{re10k_0ed25f15cbccd939_src-frame_00000_tgt-frame_00050_idx008564}{RE10K-2}
\appdemorow{re10k_2c9018ef57c6b061_src-frame_00040_tgt-frame_00090_idx025644}{RE10K-3}
\appdemorow{re10k_2e4cac06a4f92261_src-frame_00070_tgt-frame_00020_idx027492}{RE10K-4}
\appdemorow{re10k_3e8dd5a6930ecb92_src-frame_00140_tgt-frame_00190_idx036212}{RE10K-5}
\appdemorow{re10k_3f68a1e365e94eb4_src-frame_00080_tgt-frame_00030_idx037469}{RE10K-6}
\appdemorow{re10k_4b0fdb10ae15684b_src-frame_00030_tgt-frame_00080_idx040069}{RE10K-7}
\appdemorow{re10k_4bc7ca44cc62b8b1_src-frame_00180_tgt-frame_00130_idx041119}{RE10K-8}
\appdemorow{re10k_6a2423d943574e11_src-frame_00040_tgt-frame_00090_idx057786}{RE10K-9}
\appdemorow{re10k_0a9f2831a3e73de8_src-frame_00110_tgt-frame_00130_idx000409}{RE10K-10}
\appdemorow{re10k_4cc48509585e4157_src-frame_00010_tgt-frame_00060_idx043072}{RE10K-11}
\end{tabular}
\caption{\textbf{Extended qualitative comparison on novel view synthesis (RE10K).} For each scene we show the input source view (\textcolor{legendinput}{blue border}), the renderings produced by the baselines (\textcolor{legendbaseline}{gray border}: PhotoNVS, VIVID, OVIE, GLD, LagerNVS), our ESC variants (\textcolor{legendours}{orange border}: ESC-Shadow, ESC-Base, and the geometry-guided ESC-PCC-BoN (Best-of-N)), and the ground-truth target view (\textcolor{legendtarget}{green border}). All rows are scenes from the RealEstate10K (RE10K) benchmark. Our variants reproduce indoor scene geometry and fine detail more faithfully than the baselines, with ESC-PCC-BoN consistently best matching the ground-truth target.}
\label{fig:qualitative_nvs_re10k}
\end{figure*}

\clearpage
\newpage
% ---------------------------------------------------------------------------
% Best-of-N qualitative table on DL3DV (+ one Tanks & Temples scene).
% Same column layout as tab:best_of_n_re10k.
% ---------------------------------------------------------------------------
\begin{figure*}[p]
\centering
\setlength{\tabcolsep}{1.5pt}
\renewcommand{\arraystretch}{0.5}
\begin{tabularx}{\textwidth}{Y Y | Y | Y Y Y}
\demohd{\textcolor{legendinput}{Source}} & \demohd{\textcolor{legendtarget}{GT}} & \demohd{\textcolor{legendours}{\textbf{Best}}} & \multicolumn{3}{c}{\demohd{\textcolor{legendours}{Candidates (Best-of-N)}}} \\
\hline
% Scene 1: dl3dv_004811
\rescellsrc{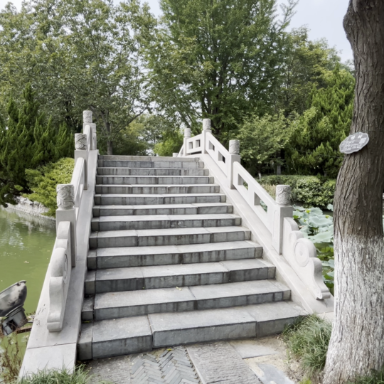} &
\rescellgt{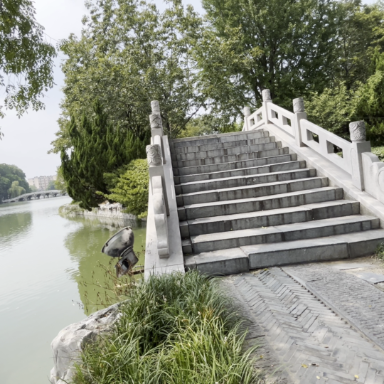} &
\rescellours{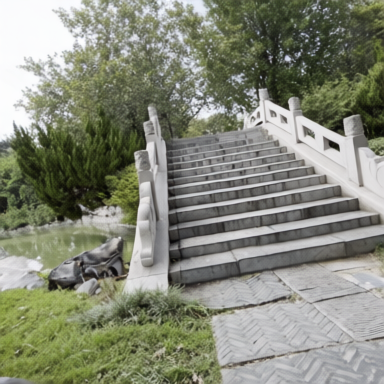} &
\rescellours{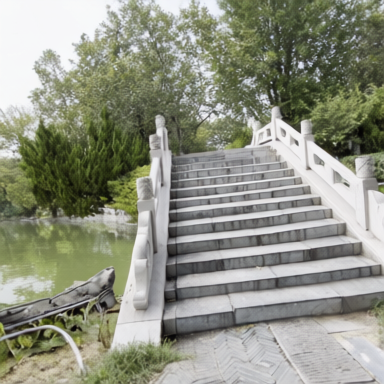} &
\rescellours{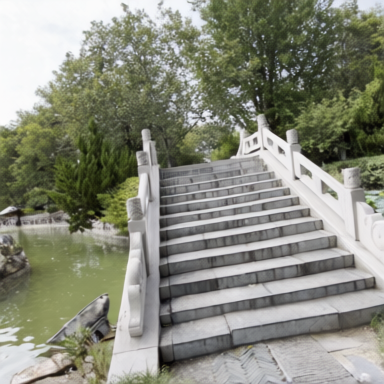} &
\rescellours{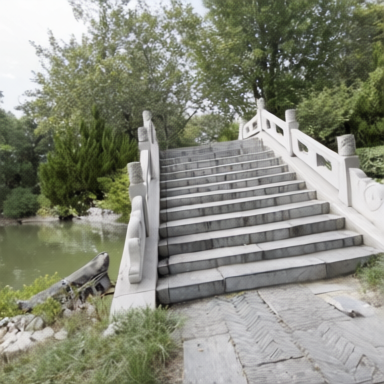} \\
% Scene 2: dl3dv_011340
\rescellsrc{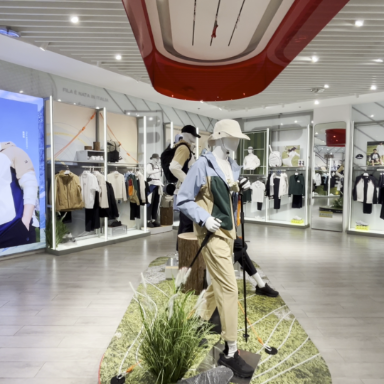} &
\rescellgt{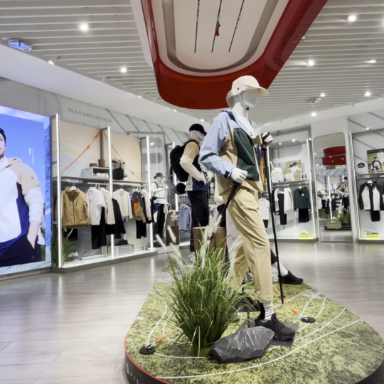} &
\rescellours{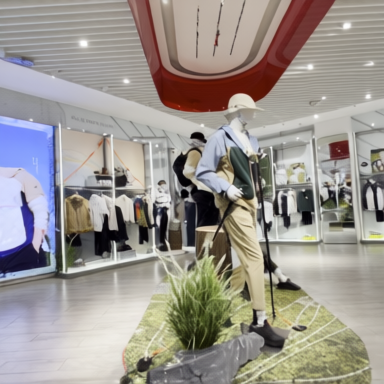} &
\rescellours{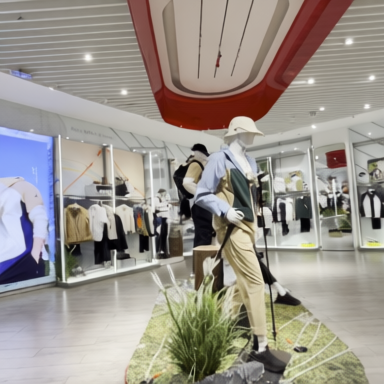} &
\rescellours{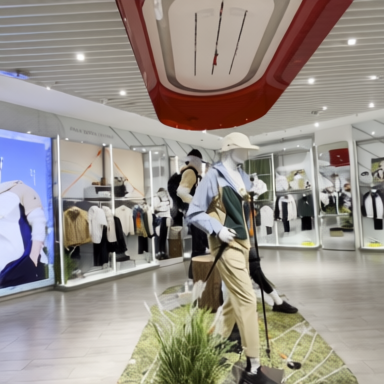} &
\rescellours{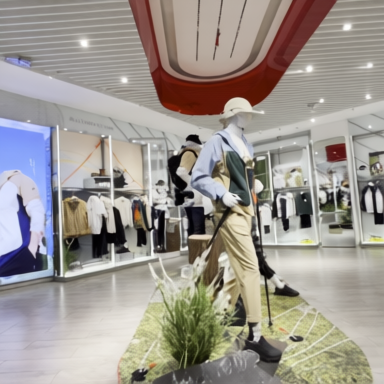} \\
% Scene 3: dl3dv_ae90ea23_020170
\rescellsrc{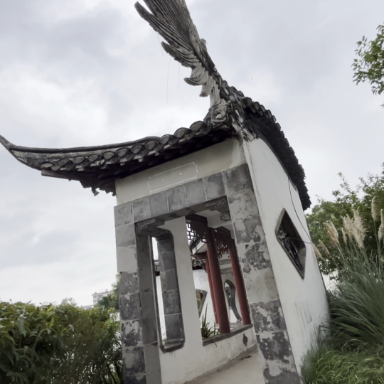} &
\rescellgt{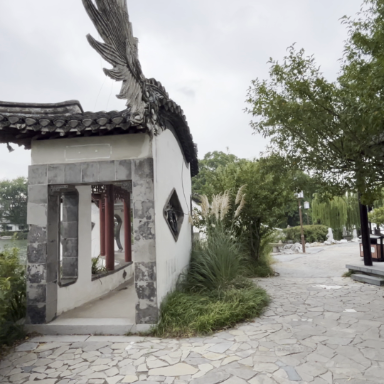} &
\rescellours{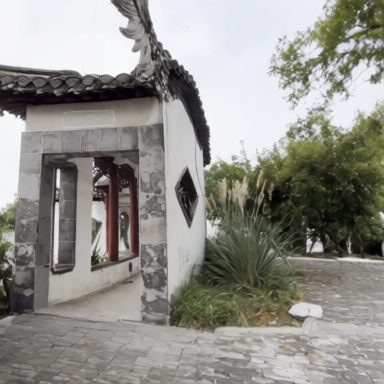} &
\rescellours{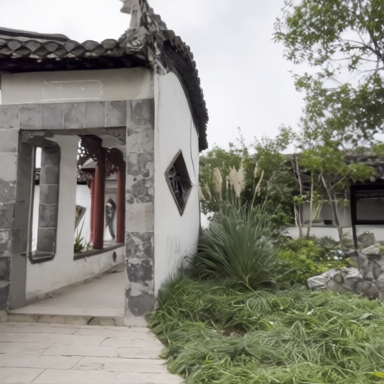} &
\rescellours{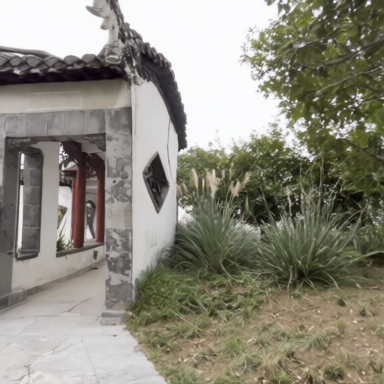} &
\rescellours{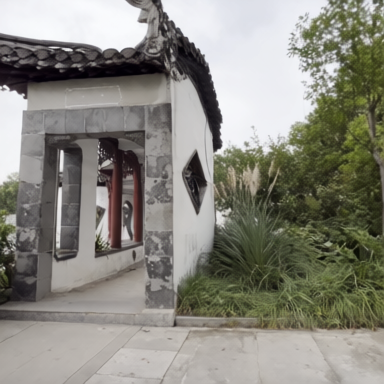} \\
% Scene 4: dl3dv_b93f5691_021151
\rescellsrc{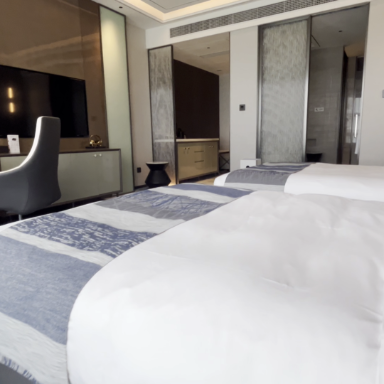} &
\rescellgt{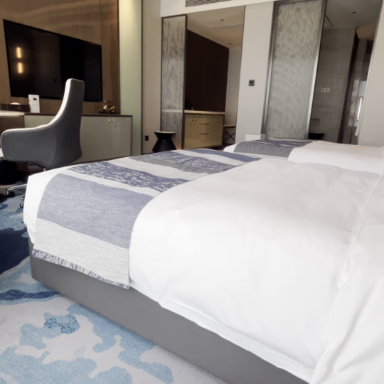} &
\rescellours{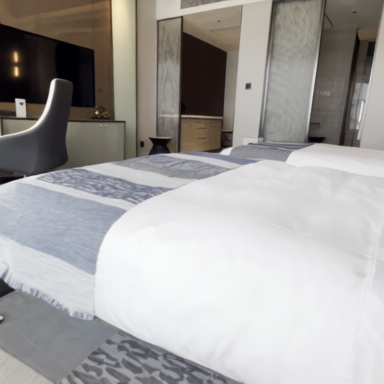} &
\rescellours{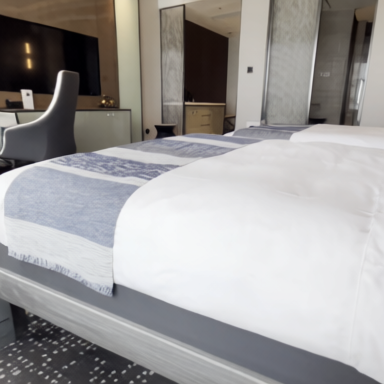} &
\rescellours{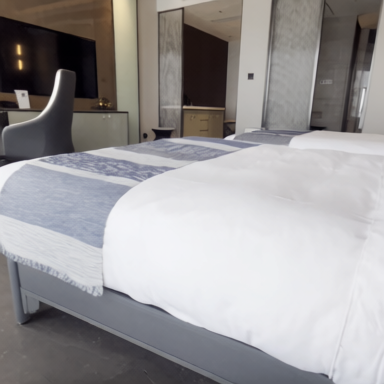} &
\rescellours{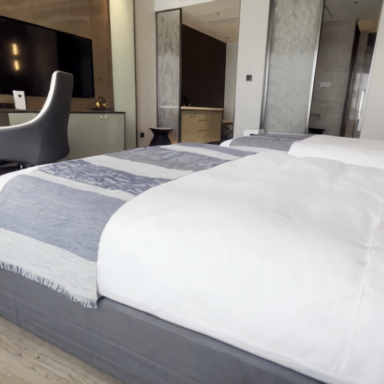} \\
% Scene 5: dl3dv_005426
\rescellsrc{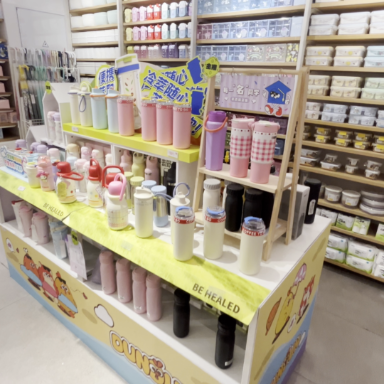} &
\rescellgt{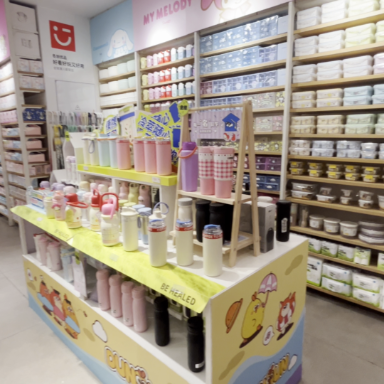} &
\rescellours{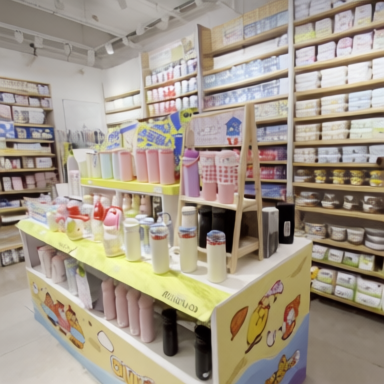} &
\rescellours{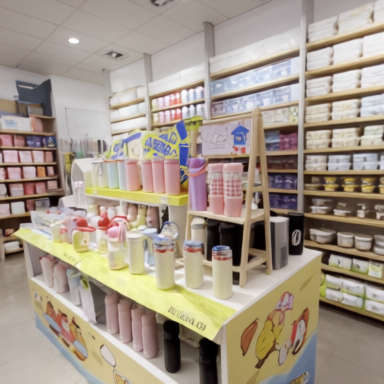} &
\rescellours{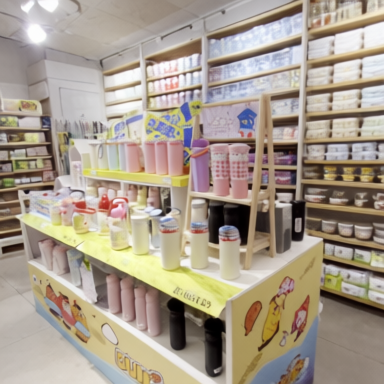} &
\rescellours{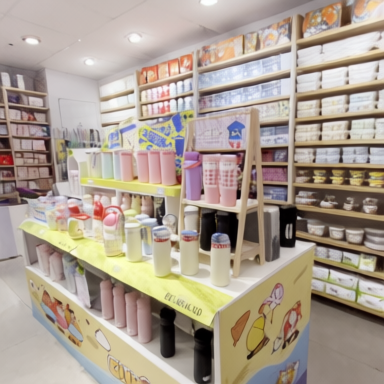} \\
% Scene 6: dl3dv_025318
\rescellsrc{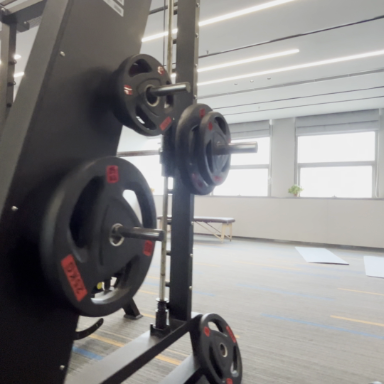} &
\rescellgt{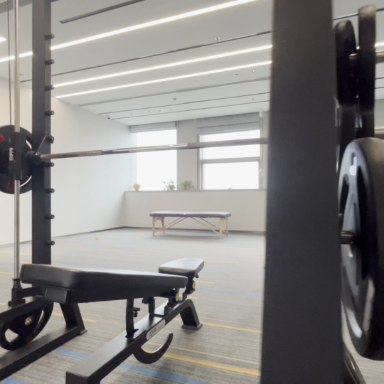} &
\rescellours{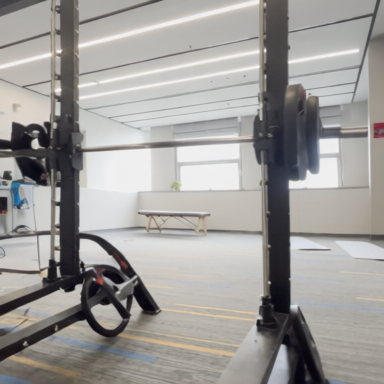} &
\rescellours{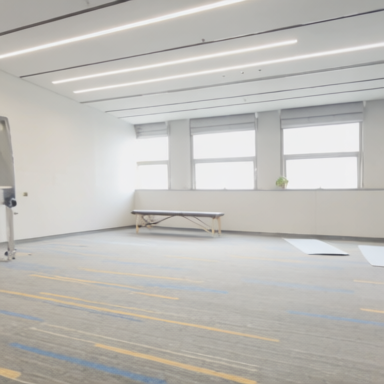} &
\rescellours{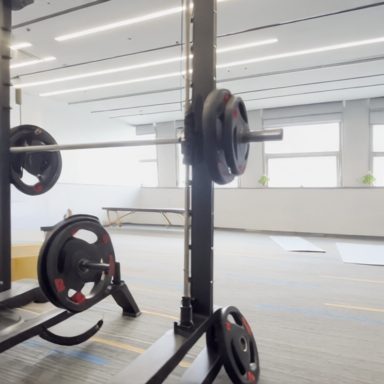} &
\rescellours{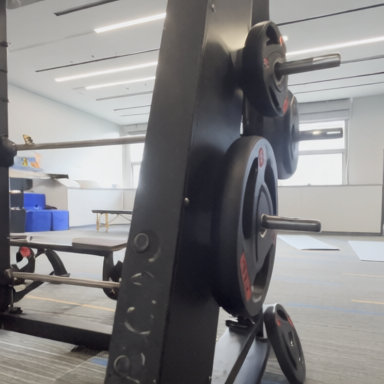} \\
% Scene 7: tnt_barn (Tanks & Temples)
\rescellsrc{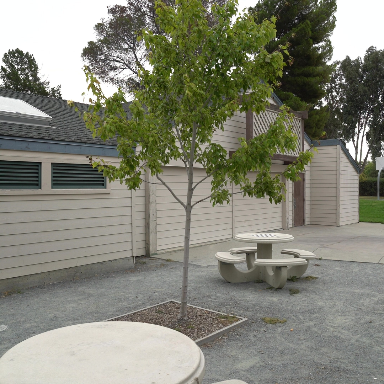} &
\rescellgt{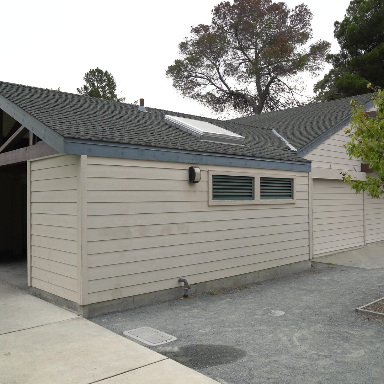} &
\rescellours{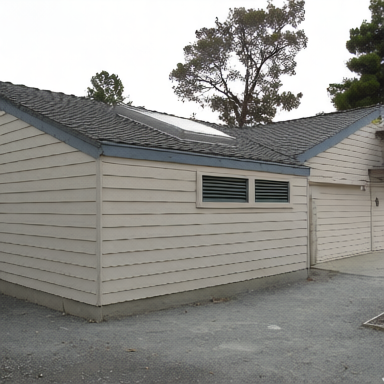} &
\rescellours{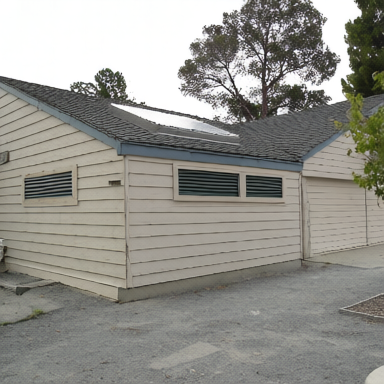} &
\rescellours{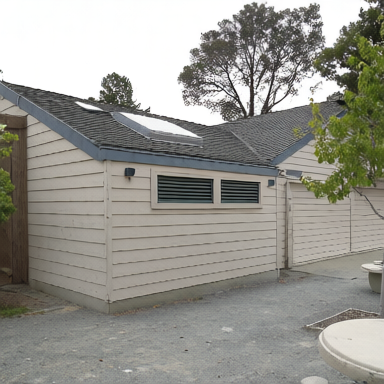} &
\rescellours{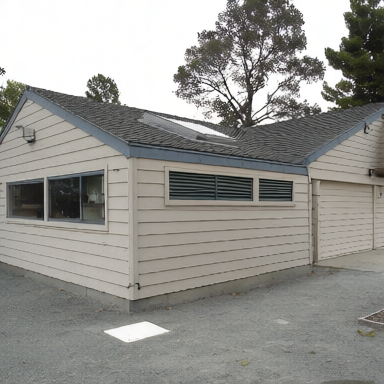} \\
\end{tabularx}
\caption{\textbf{Best-of-N novel-view synthesis on DL3DV.} Each row shows the \textcolor{legendinput}{source view (blue border)}, \textcolor{legendtarget}{ground truth (green border)}, the \textcolor{legendours}{PCC-selected best sample (orange border)}, and three additional \textcolor{legendours}{candidates (orange border)} from the same Best-of-N pool. The first six rows are DL3DV scenes; the last row is the \emph{Barn} scene from Tanks \& Temples.}
\label{tab:best_of_n_dl3dv}
\end{figure*}

% ---------------------------------------------------------------------------
% Extended Best-of-N qualitative table (remaining 7 scenes not shown in
% tab:best_of_n in the main paper). Same column layout as the main table.
% ---------------------------------------------------------------------------
\begin{figure*}[p]
\centering
\setlength{\tabcolsep}{1.5pt}
\renewcommand{\arraystretch}{0.5}
\begin{tabularx}{\textwidth}{Y Y | Y | Y Y Y}
\demohd{\textcolor{legendinput}{Source}} & \demohd{\textcolor{legendtarget}{GT}} & \demohd{\textcolor{legendours}{\textbf{Best}}} & \multicolumn{3}{c}{\demohd{\textcolor{legendours}{Candidates (Best-of-N)}}} \\
\hline
% Scene 1: re10k_0fd536fc3c8fdf19
\rescellsrc{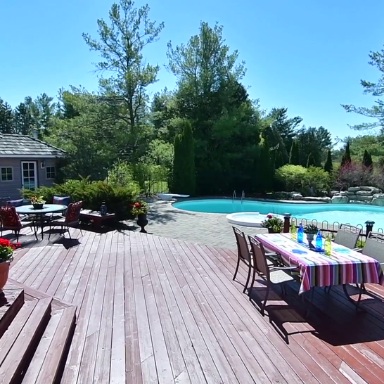} &
\rescellgt{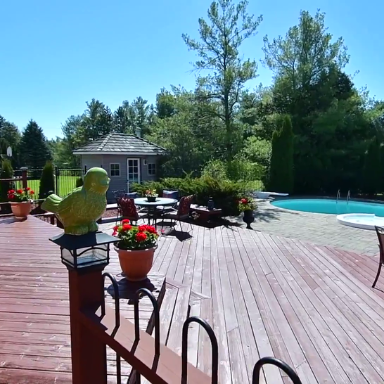} &
\rescellours{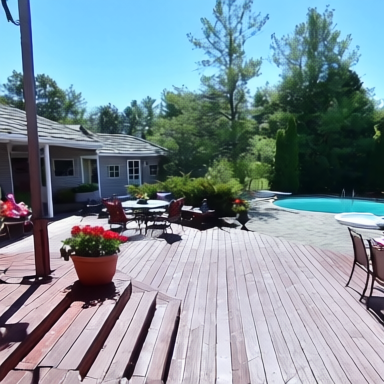} &
\rescellours{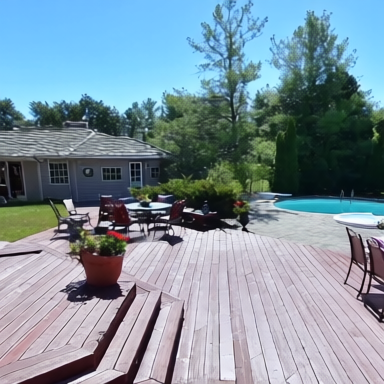} &
% \rescell{\resroot/re10k_0fd536fc3c8fdf19_src-frame_00200_tgt-frame_00250_idx010522/eac_base_0004.png} &
\rescellours{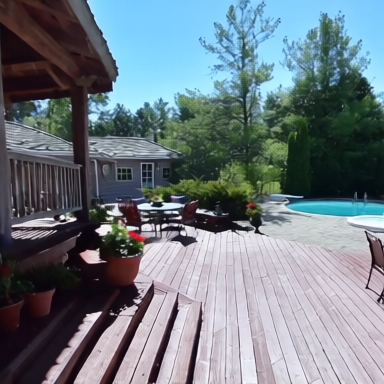} &
% \rescell{\resroot/re10k_0fd536fc3c8fdf19_src-frame_00200_tgt-frame_00250_idx010522/eac_base_0018.png} &
\rescellours{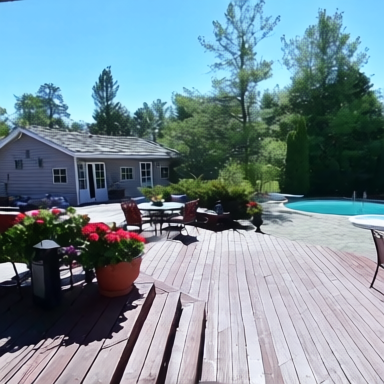} \\
% Scene 2: re10k_1acca83523e5bcf7
\rescellsrc{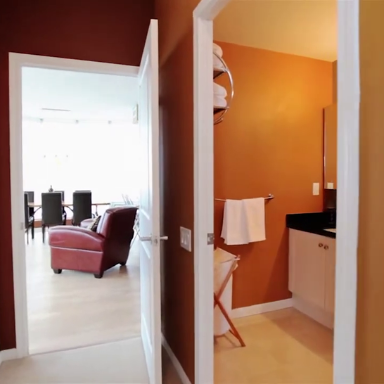} &
\rescellgt{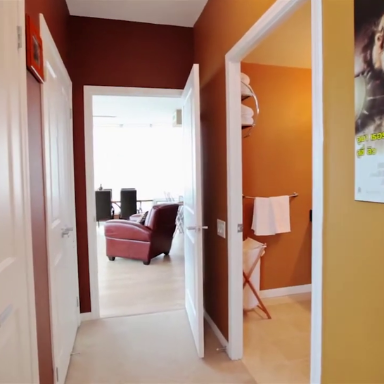} &
\rescellours{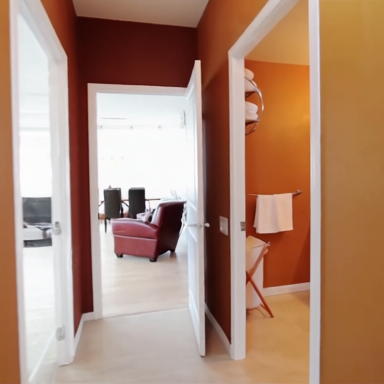} &
\rescellours{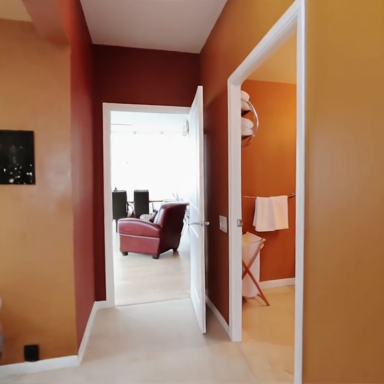} &
% \rescell{\resroot/re10k_1acca83523e5bcf7_src-frame_00070_tgt-frame_00020_idx012940/eac_base_0004.png} &
% \rescell{\resroot/re10k_1acca83523e5bcf7_src-frame_00070_tgt-frame_00020_idx012940/eac_base_0007.png} &
\rescellours{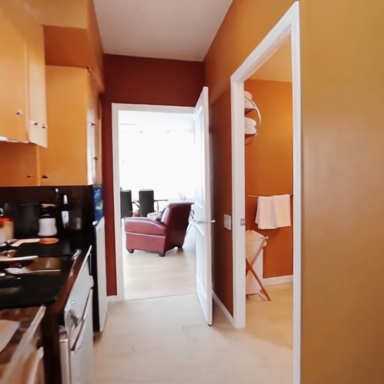} &
\rescellours{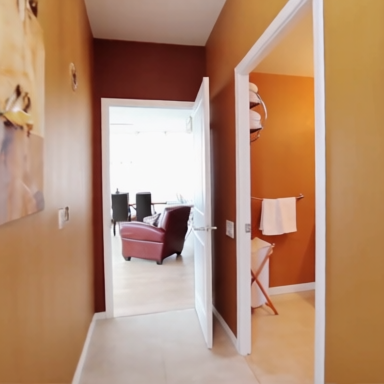} \\
% Scene 3: re10k_2bd7cee1fa9c8996
\rescellsrc{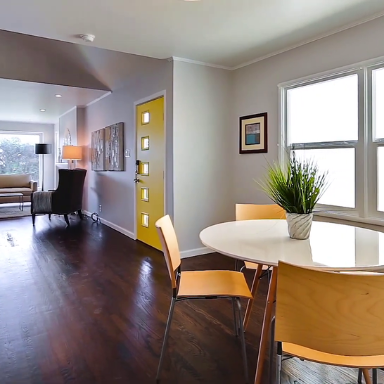} &
\rescellgt{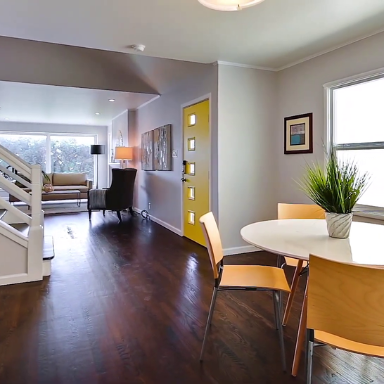} &
\rescellours{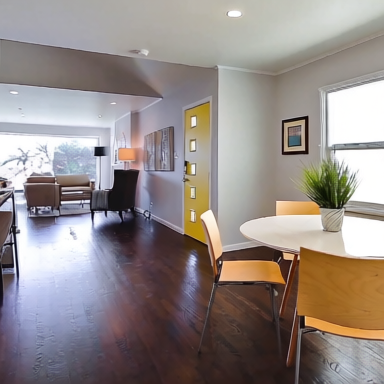} &
\rescellours{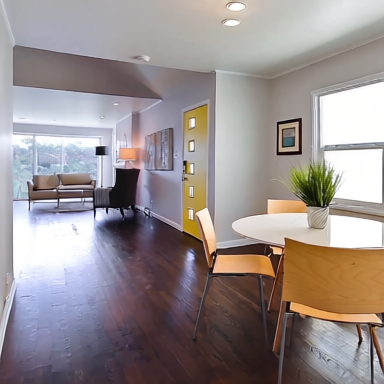} &
\rescellours{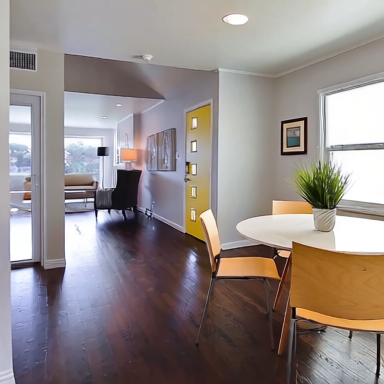} &
% \rescell{\resroot/re10k_2bd7cee1fa9c8996_src-frame_00010_tgt-frame_00060_idx023726/eac_base_0010.png} &
\rescellours{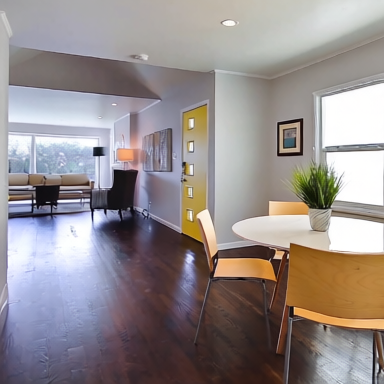} \\
% \rescell{\resroot/re10k_2bd7cee1fa9c8996_src-frame_00010_tgt-frame_00060_idx023726/eac_base_0015.png} \\
% Scene 4: re10k_2c52d9d606a3ece2
\rescellsrc{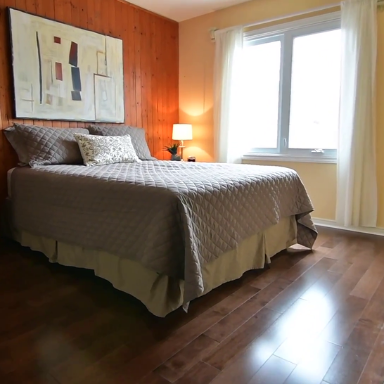} &
\rescellgt{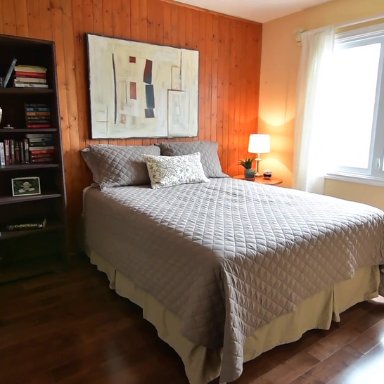} &
\rescellours{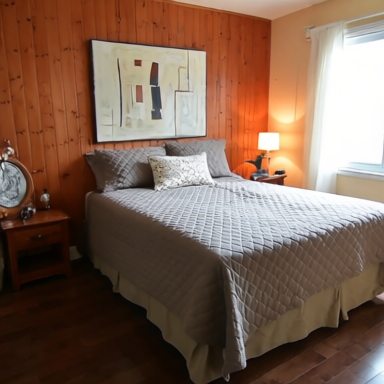} &
\rescellours{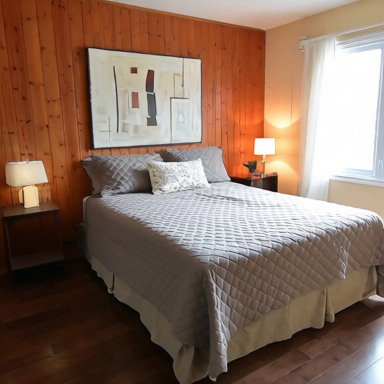} &
% \rescell{\resroot/re10k_2c52d9d606a3ece2_src-frame_00030_tgt-frame_00080_idx025165/eac_base_0011.png} &
% \rescell{\resroot/re10k_2c52d9d606a3ece2_src-frame_00030_tgt-frame_00080_idx025165/eac_base_0013.png} &
\rescellours{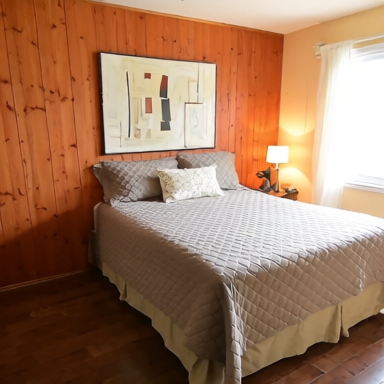} &
\rescellours{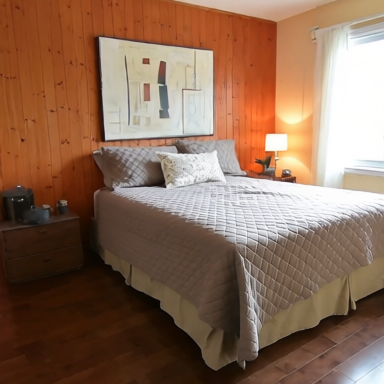} \\
% Scene 5: re10k_3ce8f87fcfc988a8
\rescellsrc{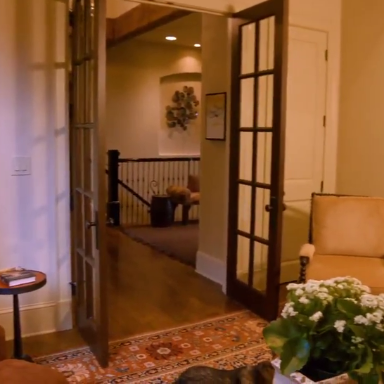} &
\rescellgt{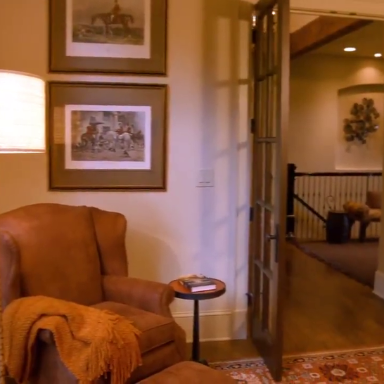} &
\rescellours{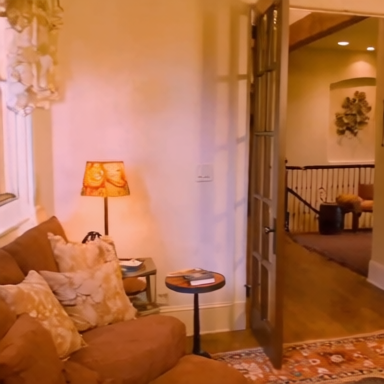} &
% \rescell{\resroot/re10k_3ce8f87fcfc988a8_src-frame_00050_tgt-frame_00000_idx034137/eac_base_0001.png} &
\rescellours{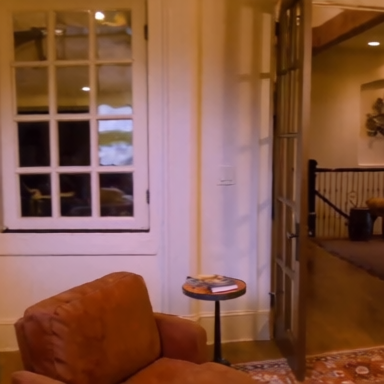} &
\rescellours{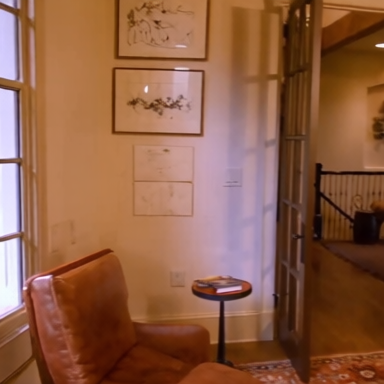} &
\rescellours{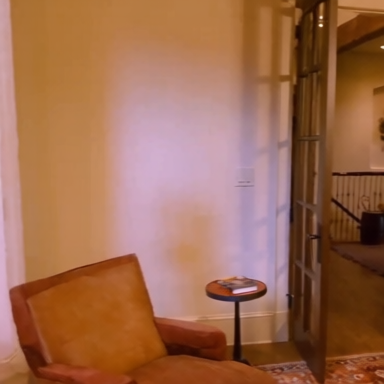} \\
% \rescell{\resroot/re10k_3ce8f87fcfc988a8_src-frame_00050_tgt-frame_00000_idx034137/eac_base_0019.png} \\
% Scene 6: re10k_3d4645318868a4f3
\rescellsrc{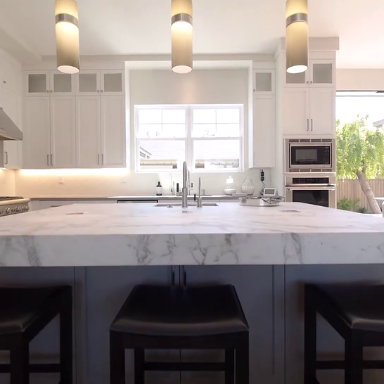} &
\rescellgt{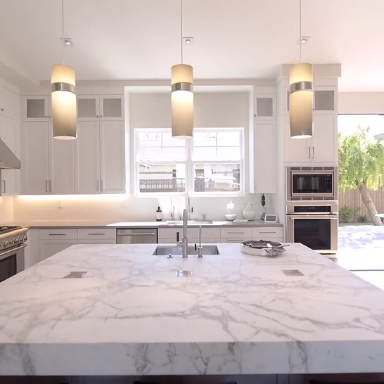} &
\rescellours{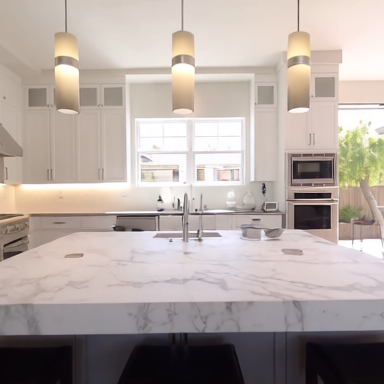} &
\rescellours{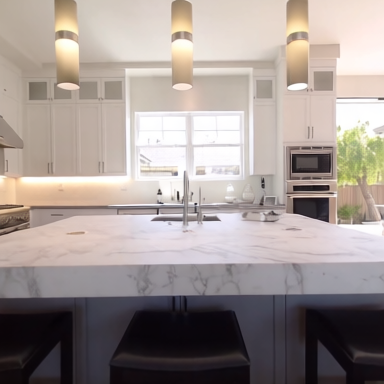} &
% \rescell{\resroot/re10k_3d4645318868a4f3_src-frame_00010_tgt-frame_00060_idx035360/eac_base_0004.png} &
% \rescell{\resroot/re10k_3d4645318868a4f3_src-frame_00010_tgt-frame_00060_idx035360/eac_base_0008.png} &
\rescellours{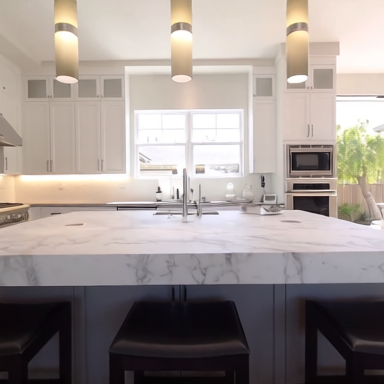} &
\rescellours{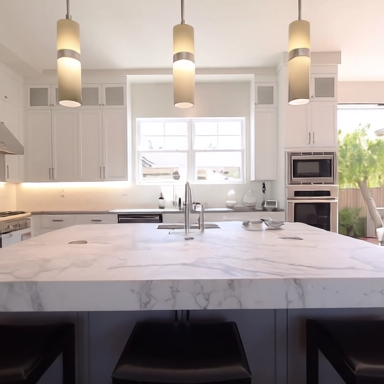} \\
% Scene 7: tnt_truck
\rescellsrc{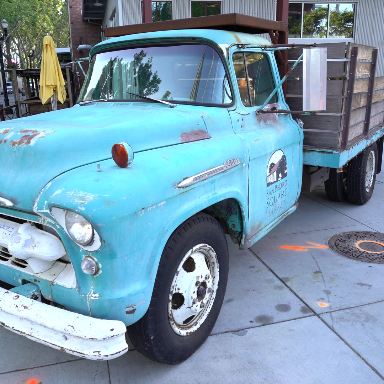} &
\rescellgt{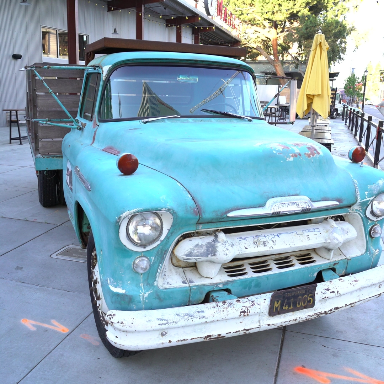} &
\rescellours{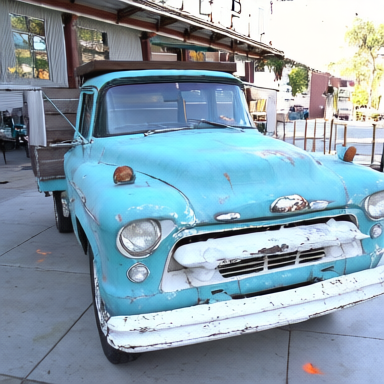} &
\rescellours{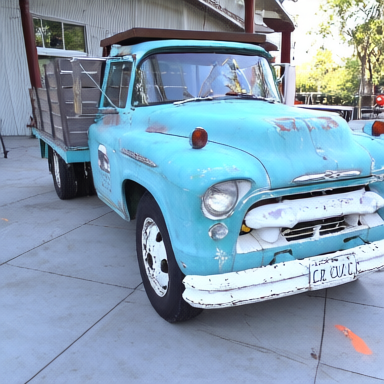} &
\rescellours{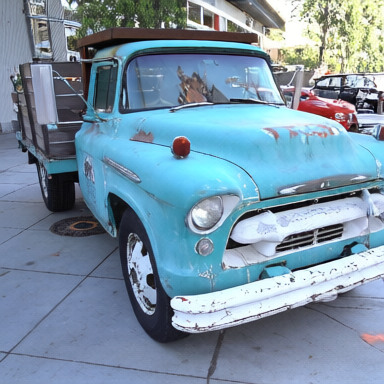} &
% \rescell{\resroot/tnt_truck_src-000056_tgt-000066_idx005317/eac_base_0013.png} &
% \rescell{\resroot/tnt_truck_src-000056_tgt-000066_idx005317/eac_base_0019.png} &
\rescellours{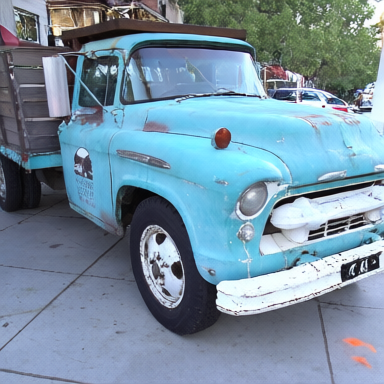} \\
\end{tabularx}
\caption{\textbf{Additional Best-of-N novel-view synthesis.} Each row shows the \textcolor{legendinput}{source view (blue border)}, \textcolor{legendtarget}{ground truth (green border)}, the \textcolor{legendours}{PCC-selected best sample (orange border)}, and three additional \textcolor{legendours}{candidates (orange border)} from the same Best-of-N pool. The first six rows are RE10K scenes; the last row is the \emph{Truck} scene from Tanks \& Temples.}
\label{tab:best_of_n_re10k}
\end{figure*}

\clearpage
\newpage

% ---------------------------------------------------------------------------
% RGB + depth visualization: source / GT / prediction (2 rows x 3 columns).
% Row 1: RGB images; Row 2: corresponding depth maps.
% Borders: blue = source, green = GT, orange = prediction.
% ---------------------------------------------------------------------------
\begin{table*}[p]
\centering
\setlength{\tabcolsep}{1.5pt}
\renewcommand{\arraystretch}{0.5}
% Vertically center each cell on the row baseline so the rotated row
% labels line up with the middle of the images.
\newcommand{\dvcell}[1]{\raisebox{-0.5\height}{#1}}
\newcommand{\dvlabel}[1]{\raisebox{-0.5\height}{\rotatebox{90}{\scriptsize #1}}}
\begin{tabularx}{\textwidth}{@{}c Y Y Y@{}}
& \demohd{\textcolor{legendinput}{Source}} & \demohd{\textcolor{legendtarget}{GT}} & \demohd{\textcolor{legendours}{\textbf{Prediction}}} \\
\hline
% Scene 1
\dvlabel{RGB} &
\dvcell{\rescellsrc{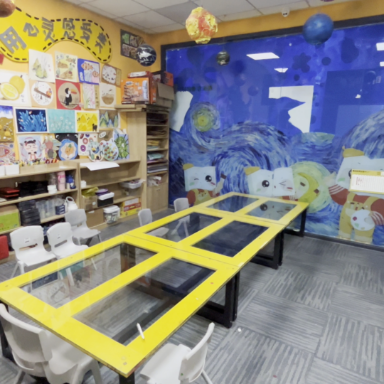}} &
\dvcell{\rescellgt{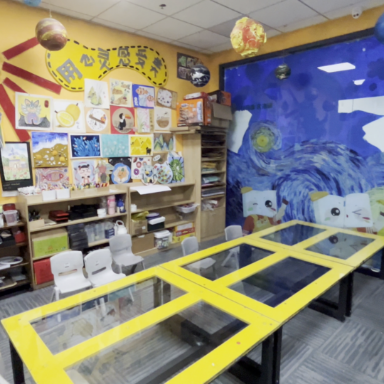}} &
\dvcell{\rescellours{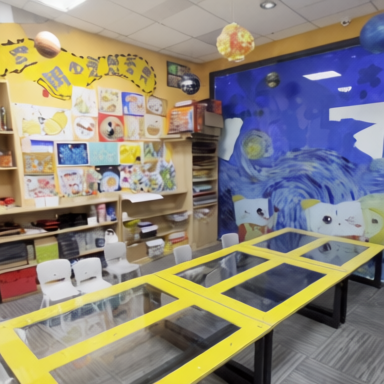}} \\
\dvlabel{Depth} &
\dvcell{\rescellsrc{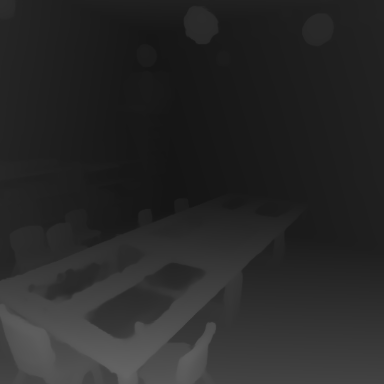}} &
\dvcell{\rescellgt{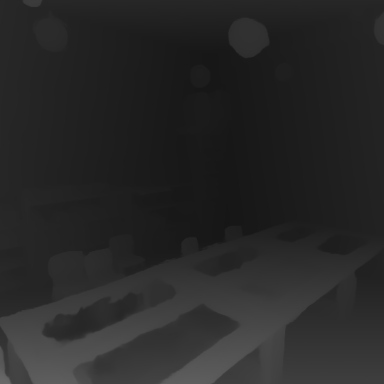}} &
\dvcell{\rescellours{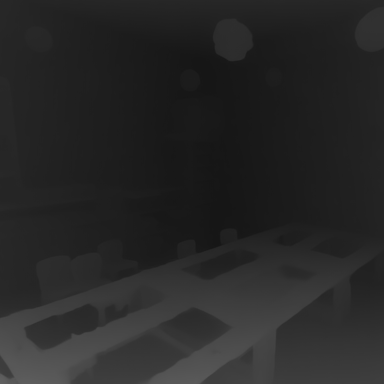}} \\[3pt]
% Scene 2
\dvlabel{RGB} &
\dvcell{\rescellsrc{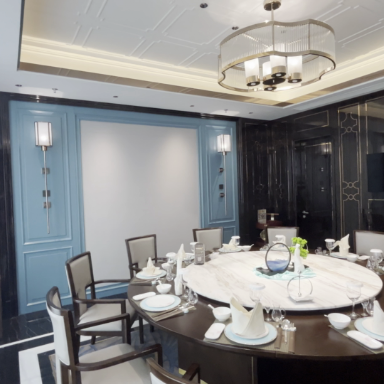}} &
\dvcell{\rescellgt{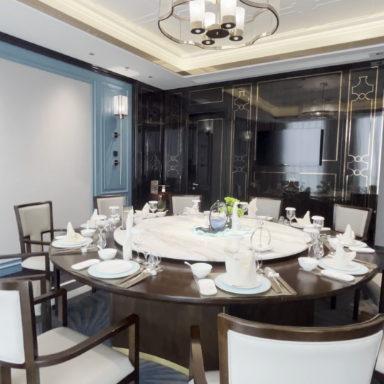}} &
\dvcell{\rescellours{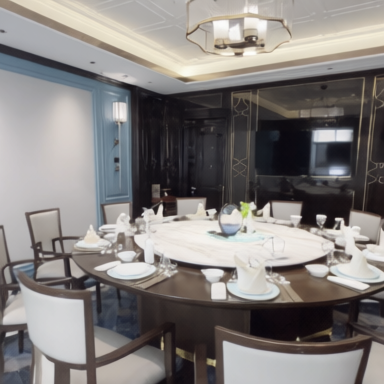}} \\
\dvlabel{Depth} &
\dvcell{\rescellsrc{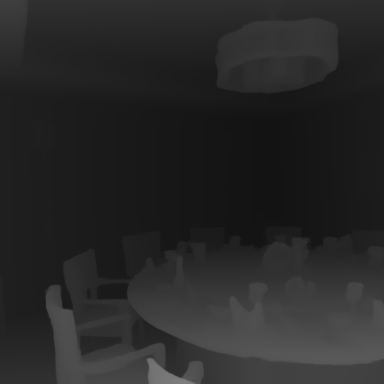}} &
\dvcell{\rescellgt{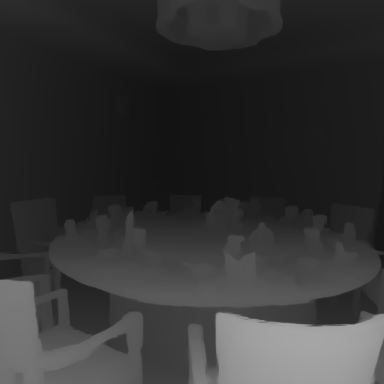}} &
\dvcell{\rescellours{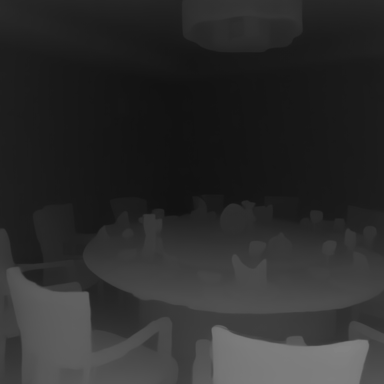}} \\
\end{tabularx}
\caption{\textbf{RGB and depth visualization.} For each of two scenes we show the \textcolor{legendinput}{source view (blue border)}, the \textcolor{legendtarget}{ground-truth target (green border)}, and our \textcolor{legendours}{prediction (orange border)}, as an RGB row and the corresponding depth row. Our predictions closely match the ground-truth targets in both appearance and geometry.}
\label{tab:depth_visual}
\end{table*}

\clearpage
\newpage

%% file: appendix/compute_time.tex
Using the standard estimate $C\!\approx\!6ND$ with $N\!\approx\!2.2\times10^{10}$ active parameters (the frozen Qwen-Image-Edit backbone plus the trainable LoRA and 5-layer ControlNet adapter) and $D\!\approx\!5.2\times10^{10}$ tokens ($177\text{k}$ steps $\times$ a global batch of $128$ $\times$ $2304$ latent tokens per $384{\times}384$ sample), our full training run costs approximately $6.8\times10^{21}$ FLOPs.

%% file: main/ablation/simulator_results.tex
\subsection{How does ESC perform in simulation with ground-truth geometry?}
\label{sub:simulation_results}
\begin{table}[t]
\centering
\small
\begin{tabular}{l cccc c}
\toprule
\textbf{Metric} & \textbf{\modelname-shadow} & \textbf{\modelname-Base} & \textbf{\modelname-PCC-BoN}& \textbf{\modelname-PCC-FT} & \textbf{Oracle} \\
& @130k iters & @100k iters & 100k iters & @100k+30k iters &\\
\midrule

\multicolumn{6}{l}{\textit{Synthesis Quality}} \\
\midrule
PSNR $\uparrow$ & \second{$31.90 \pm 3.38$} & $30.73 \pm 3.69$ & \third{$31.59 \pm 3.29$} & \first{$33.65 \pm 3.87$} & $\cdot$ \\
SSIM $\uparrow$ & \second{$0.9607 \pm 0.0178$} & $0.9532 \pm 0.0259$ & \third{$0.9578 \pm 0.0176$} & \first{$0.9693 \pm 0.0170$} & $\cdot$ \\

\midrule
\multicolumn{6}{l}{\textit{Perceptual Quality}} \\
\midrule
LPIPS $\downarrow$ & \second{$0.0419 \pm 0.0319$} & $0.0584 \pm 0.0566$ & \third{$0.0460 \pm 0.0324$} & \first{$0.0324 \pm 0.0301$} & $\cdot$ \\
DINO-Sim $\uparrow$ & \second{$0.9715 \pm 0.0302$} & $0.9650 \pm 0.0344$ & \third{$0.9652 \pm 0.0326$} & \first{$0.9746 \pm 0.0308$} & $\cdot$ \\

\midrule
\multicolumn{6}{l}{\textit{Geometric Fidelity}} \\
\midrule
PCC $\downarrow$ & \third{$0.1057 \pm 0.1033$} & $0.1220 \pm 0.1059$ & \second{$0.0915 \pm 0.0952$} & \first{$0.0721 \pm 0.0751$} & $0.0576 \pm 0.0530$ \\

\midrule
\multicolumn{6}{l}{\textit{Trajectory \& View Prompt Consistency}} \\
\midrule
AUC@5 $\uparrow$ & \second{$0.5986$} & $0.4403$ & \third{$0.5283$} & \first{$0.6132$} & $0.6250$ \\
AUC@15 $\uparrow$ & \second{$0.8028$} & $0.7067$ & \third{$0.7758$} & \first{$0.8095$} & $0.7500$ \\
AUC@30 $\uparrow$ & \second{$0.8598$} & $0.8046$ & \third{$0.8462$} & \first{$0.8641$} & $0.7948$ \\
RPE-R $\downarrow$ & \second{$10.69 \pm 31.40$} & $12.34 \pm 31.12$ & \third{$11.03 \pm 31.10$} & \first{$10.37 \pm 31.81$} & $15.20 \pm 32.67$ \\
RPE-T $\downarrow$ & \first{$1.58 \pm 1.21$} & $2.69 \pm 3.26$ & \third{$1.82 \pm 1.47$} & \second{$1.71 \pm 1.22$} & $1.15 \pm 0.90$ \\
\bottomrule
\end{tabular}
\small
\caption{\textbf{Ablation of \modelname    variants on the Blender benchmark.} We compare four \modelname configurations and an oracle across four metric groups: synthesis quality (PSNR, SSIM), perceptual quality (LPIPS, DINO similarity), geometric fidelity (PCC), and trajectory \& view-prompt consistency (pose AUC at $5^\circ$/$15^\circ$/$30^\circ$, and relative pose errors RPE-R in degrees and RPE-T). Arrows indicate whether higher ($\uparrow$) or lower ($\downarrow$) is better. \emph{\modelname-Base} uses a single generated sample; \emph{\modelname-PCC-BoN} selects the best of 20 samples per view by geometric consistency (PCC); \emph{\modelname-PCC-FT} additionally fine-tunes with the PCC objective; and \emph{\modelname-shadow} denotes a model variant we trained without depth conditioning and prediction. \emph{Oracle} indicates the ground-truth label compared against the paired DA3 estimate on the source images. The \first{best}, \second{second-best}, and \third{third-best} results among the \modelname variants in each row are highlighted. PCC-aware selection and fine-tuning consistently improve perceptual quality and geometric fidelity, with \modelname-PCC-FT achieving the best results across nearly all metrics. More qualitative results are shown in \Cref{tab:blender_qualitative}.}
\label{tab:blender_transposed_benchmarks_with_std}
\end{table}

\begin{figure}[h]
\centering
\includegraphics[width=\linewidth]{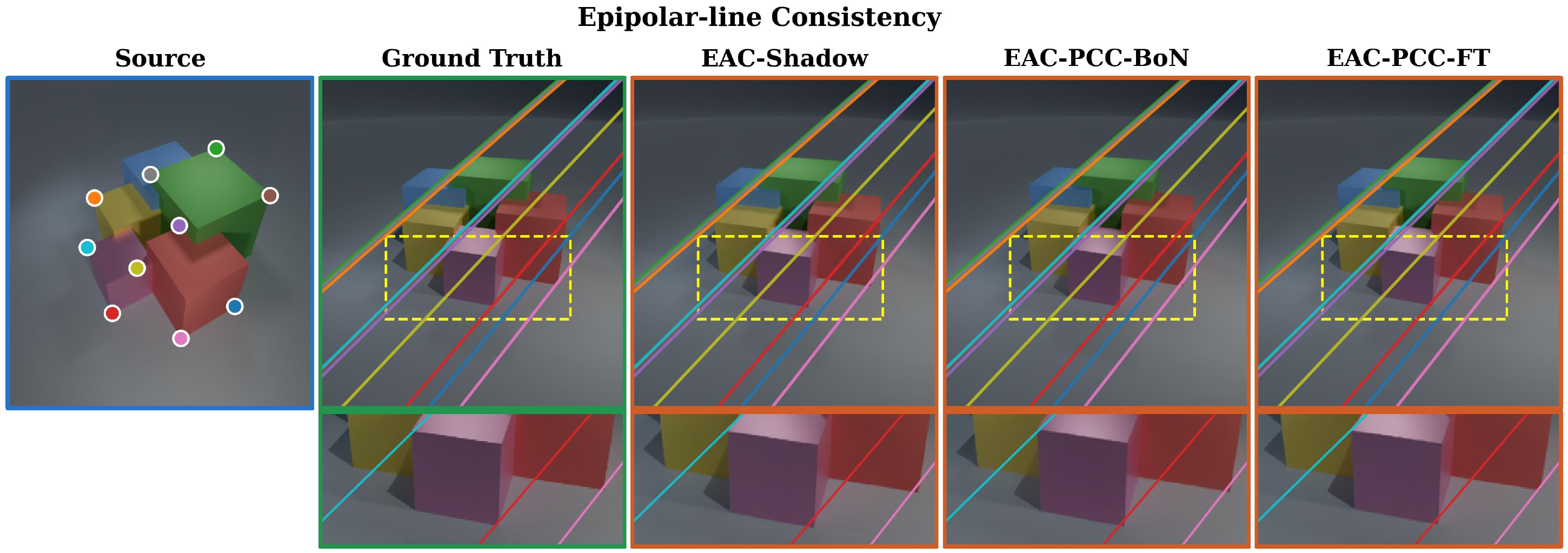}
\caption{\textbf{Qualitative comparison of \modelname variants on a simulation environment.} Panels show the source view (\textcolor{legendinput}{blue border}), ground-truth (\textcolor{legendtarget}{green border}), and three \modelname variant generations (\textcolor{legendours}{orange border}). The bottom row magnifies the yellow dashed region to illustrate 3D geometric consistency using color-matched epipolar lines derived from the source keypoints. While the bottom-right corner of the pink(red) cube nearly touches the red(pink) epipolar line in the ground-truth, it sits nearly flush in \modelname-PCC-FT, shows a minor pixel gap in \modelname-PCC-BoN, and leaves a small but more prominent gap in \modelname-shadow, establishing the order of geometric alignment quality across the variants in this example.}\label{fig:epipolar_alignment_results}
\end{figure}

A primary advantage of utilizing our synthetic Blender dataset is that it inherently provides pristine, exact ground-truth depth maps. Unlike the noisy pseudo-ground-truth depth maps extracted from alternative 3D reconstruction or estimation frameworks, these clean structural priors facilitate exceptionally smooth optimization dynamics and stable convergence across all model variants. Furthermore, this controlled simulation environment grants us the unique flexibility to generate diverse scene variations specifically within the occluded regions of the original viewpoint. This allows us to explicitly evaluate the network's generative completion capabilities under severe disocclusions, with qualitative examples demonstrated in~\Cref{fig:occlusion_comparison}.

In this ablation experiment, we explore using the Projection Consistency Constraint (PCC) directly as a learning objective to fine-tune our model. Initializing from the pre-trained \modelname-Base EMA checkpoint, we incorporate the PCC regularizer (detailed further in \Cref{sec:pcc_analysis}) to explicitly enforce strict geometric alignment across views\footnote{We utilize the model's own predicted depth map for PCC evaluation during this training phase and Best-of-$N$ selection. Conversely, for final benchmarking, we employ the DA3 estimated depth.\label{fn:pcc_depth}}. To avoid disrupting the core flow-matching process at high noise levels, this PCC objective is applied exclusively over a truncated timestep range. This constraint ensures that the geometric regularization effectively shapes only the later, more structurally resolved stages of denoising.

Overall, all trained model configurations produce highly reasonable and geometrically plausible novel-view predictions, with the supervised fine-tuning (\emph{\modelname-PCC-FT}) variant consistently delivering the highest visual and structural fidelity (See \Cref{tab:blender_transposed_benchmarks_with_std} for the quantitative results and \Cref{tab:blender_qualitative} for the qualitative results). To rigorously assess the geometric integrity of these generations, we overlay epipolar lines across the synthesized views in ~\Cref{fig:epipolar_alignment_results}. The precise correspondence of visual features along these epipolar lines demonstrates that our ControlNet adapter effectively interprets Plücker tokens to generate novel views that conform to the underlying camera geometry. This alignment highlights the model's 3D consistency with the original scene.